%% file: first-proof-monograph.tex
\documentclass[12pt,oneside]{book}
\usepackage[margin=1in]{geometry}
\usepackage{iftex}
\ifPDFTeX
  \usepackage[T1]{fontenc}
  \usepackage[utf8]{inputenc}
  \usepackage{lmodern}
\else
  \usepackage{fontspec}
\fi
\usepackage{amsmath,amssymb,amsthm,mathtools}
\usepackage{longtable,booktabs}
\usepackage{enumitem}
\usepackage{setspace}
\usepackage{fancyhdr}
\usepackage{titlesec}
\usepackage[dvipsnames]{xcolor}
\usepackage{xurl}
\usepackage{textcomp}
\usepackage{hyperref}
\usepackage[nameinlink,noabbrev]{cleveref}
\usepackage[most]{tcolorbox}
\usepackage{paracol}
\usepackage{wrapfig}
\usepackage{microtype}
\usepackage{framed}
\usepackage{fancyvrb}
\usepackage{math-proofread-style}

\hypersetup{
  colorlinks=true,
  linkcolor=blue!50!black,
  citecolor=blue!50!black,
  urlcolor=blue!60!black,
  pdftitle={A First Proof Sprint},
  pdfauthor={Joseph Corneli}
}

\setlist[itemize]{itemsep=0.3em,topsep=0.3em}
\setlist[enumerate]{itemsep=0.3em,topsep=0.3em}
\EnableMathProofColors

\providecommand{\tightlist}{}
\definecolor{shadecolor}{RGB}{248,248,248}
\DefineVerbatimEnvironment{Highlighting}{Verbatim}{commandchars=\\\{\}}
\newenvironment{Shaded}{\begin{snugshade}}{\end{snugshade}}
\providecommand{\NormalTok}[1]{#1}

\ifPDFTeX
  \DeclareUnicodeCharacter{2218}{\ensuremath{\circ}}
  \DeclareUnicodeCharacter{2297}{\ensuremath{\otimes}}
  \DeclareUnicodeCharacter{2713}{\ensuremath{\checkmark}}
  \DeclareUnicodeCharacter{2202}{\ensuremath{\partial}}
  \DeclareUnicodeCharacter{2245}{\ensuremath{\cong}}
  \DeclareUnicodeCharacter{2227}{\ensuremath{\wedge}}
  \DeclareUnicodeCharacter{2295}{\ensuremath{\oplus}}
  \DeclareUnicodeCharacter{2229}{\ensuremath{\cap}}
  \DeclareUnicodeCharacter{220E}{\ensuremath{\square}}
  \DeclareUnicodeCharacter{2260}{\ensuremath{\neq}}
  \DeclareUnicodeCharacter{2261}{\ensuremath{\equiv}}
  \DeclareUnicodeCharacter{2264}{\ensuremath{\le}}
  \DeclareUnicodeCharacter{2265}{\ensuremath{\ge}}
  \DeclareUnicodeCharacter{2282}{\ensuremath{\subset}}
  \DeclareUnicodeCharacter{2286}{\ensuremath{\subseteq}}
  \DeclareUnicodeCharacter{222A}{\ensuremath{\cup}}
  \DeclareUnicodeCharacter{2208}{\ensuremath{\in}}
  \DeclareUnicodeCharacter{2192}{\ensuremath{\to}}
  \DeclareUnicodeCharacter{00B7}{\ensuremath{\cdot}}
  \DeclareUnicodeCharacter{229E}{\ensuremath{\boxplus}}
  \DeclareUnicodeCharacter{00D7}{\ensuremath{\times}}
  \DeclareUnicodeCharacter{03B8}{\ensuremath{\theta}}
  \DeclareUnicodeCharacter{03BD}{\ensuremath{\nu}}
  \DeclareUnicodeCharacter{03C0}{\ensuremath{\pi}}
  \DeclareUnicodeCharacter{03C1}{\ensuremath{\rho}}
  \DeclareUnicodeCharacter{0393}{\ensuremath{\Gamma}}
  \DeclareUnicodeCharacter{221A}{\ensuremath{\sqrt}}
  \DeclareUnicodeCharacter{00B1}{\ensuremath{\pm}}
  \DeclareUnicodeCharacter{00B9}{\textsuperscript{1}}
  \DeclareUnicodeCharacter{00B2}{\textsuperscript{2}}
  \DeclareUnicodeCharacter{2081}{\textsubscript{1}}
  \DeclareUnicodeCharacter{2083}{\textsubscript{3}}
  \DeclareUnicodeCharacter{2088}{\textsubscript{8}}
\fi

\newcommand{\gapnote}[1]{%
  \par\medskip
  \noindent\fcolorbox{red!60!black}{red!4}{%
    \parbox{0.96\linewidth}{\textbf{Open Gap / Conditional Step.} #1}%
  }%
  \par\medskip
}
\newtcolorbox{joebox}[1][]{%
  colback=orange!4, colframe=orange!50!black,
  fonttitle=\bfseries\sffamily\small,
  title={Joe}, #1,
  left=6pt, right=6pt, top=4pt, bottom=4pt,
  boxrule=0.5pt, arc=2pt,
  before skip=6pt, after skip=2pt
}
\newtcolorbox{claudebox}[1][]{%
  colback=blue!3, colframe=blue!40!black,
  fonttitle=\bfseries\sffamily\small,
  title={Claude}, #1,
  left=6pt, right=6pt, top=4pt, bottom=4pt,
  boxrule=0.5pt, arc=2pt,
  before skip=2pt, after skip=2pt
}
\newtcolorbox{codexbox}[1][]{%
  colback=green!3, colframe=green!40!black,
  fonttitle=\bfseries\sffamily\small,
  title={Codex}, #1,
  left=6pt, right=6pt, top=4pt, bottom=4pt,
  boxrule=0.5pt, arc=2pt,
  before skip=2pt, after skip=2pt
}
\newcommand{\pivot}[1]{%
  \par\smallskip
  \noindent{\small\sffamily\color{black!60}%
    \textbf{Pivot:}\space #1}%
  \par\smallskip
}

\newtcolorbox{deadpath}[1][]{%
  colback=black!3, colframe=black!40,
  fonttitle=\bfseries\sffamily\small,
  title={Dead Path}, #1,
  left=6pt, right=6pt, top=4pt, bottom=4pt,
  boxrule=0.5pt, arc=2pt, breakable,
  before skip=6pt, after skip=6pt
}
\newtcolorbox{revision}[1][]{%
  colback=yellow!5, colframe=yellow!60!black,
  fonttitle=\bfseries\sffamily\small,
  title={Revision}, #1,
  left=6pt, right=6pt, top=4pt, bottom=4pt,
  boxrule=0.5pt, arc=2pt, breakable,
  before skip=6pt, after skip=6pt
}
\newtcolorbox{openobligation}[1][]{%
  colback=red!4, colframe=red!60!black,
  colbacktitle=red!70!black, coltitle=white,
  fonttitle=\bfseries\sffamily\small,
  before title={\begingroup\DisableMathProofColors\color{white}},
  after title={\endgroup},
  title={Open Obligation}, #1,
  left=6pt, right=6pt, top=4pt, bottom=4pt,
  boxrule=0.5pt, arc=2pt, breakable,
  before skip=6pt, after skip=6pt
}
\newtcolorbox{processnote}[1][]{%
  colback=blue!3, colframe=blue!30!black,
  fonttitle=\bfseries\sffamily\small,
  title={Process Note}, #1,
  left=6pt, right=6pt, top=4pt, bottom=4pt,
  boxrule=0.5pt, arc=2pt, breakable,
  before skip=6pt, after skip=6pt
}
\newcommand{\statusnote}[1]{%
  \begin{processnote}[title={Status Note}]
  #1
  \end{processnote}
}

\newcommand{\includeannotated}[1]{%
  \begingroup
  \let\origsection\section
  \let\origsubsection\subsection
  \let\origsubsubsection\subsubsection
  \let\section\chapter
  \let\subsection\origsection
  \let\subsubsection\origsubsection
  \renewcommand{\hypertarget}[2]{##2}
  \input{#1}
  \endgroup
}
\newcommand{\includefull}[1]{%
  \begingroup
  \let\origsection\section
  \let\origsubsection\subsection
  \let\origsubsubsection\subsubsection
  \let\section\chapter
  \let\subsection\origsection
  \let\subsubsection\origsubsection
  \renewcommand{\label}[1]{}
  \renewcommand{\hypertarget}[2]{##2}
  \input{#1}
  \endgroup
}
\providecommand{\BuildGitHash}{c92a4eb}

\titleformat{\chapter}[display]
  {\bfseries\Large}
  {\filright\large\chaptername\ \thechapter}
  {0.6em}
  {\filright}
\titlespacing*{\chapter}{0pt}{-6pt}{18pt}

\title{A First Proof Sprint}
\author{J.~Corneli\thanks{Email: \texttt{joseph.corneli@hyperreal.enterprises}} \\
\small Hyperreal Enterprises Ltd}
\date{February 11--13, 2026}

\begin{document}
\frontmatter
\hypersetup{pageanchor=false}
\begingroup
\renewcommand{\thefootnote}{\fnsymbol{footnote}}
\maketitle
\endgroup
\hypersetup{pageanchor=true}
\setcounter{footnote}{0}
\renewcommand{\thefootnote}{\arabic{footnote}}
\vfill
\noindent{\small\sffamily Reproducibility: built \today{} from
\texttt{futon6} version-controlled git history (\BuildGitHash).\\
Source: \url{https://github.com/tothedarktowercame/futon6}}
\clearpage
\setcounter{page}{2}
\chapter*{Abstract}
\addcontentsline{toc}{chapter}{Abstract}
This monograph reports a multi-agent proof sprint on ten research-level
problems, combining rapid draft generation with adversarial verification,
targeted repair, and explicit provenance. The workflow uses wiring-diagram
decompositions of claim dependencies to localize gaps and coordinate
reviewer-driven revisions. Final outcomes are heterogeneous but explicit:
the manuscript distinguishes \emph{mathematical status} from
\emph{QC-validation status}. In the current ledger, Problems~1, 2, 3, and~8
are treated as mathematically closed with validation-complete records
(Problem~2 still depends on the cited newvector/test-vector theorem input);
Problems~5 and~7 are mathematically closed with partial QC status (scope-limited
closure for Problem~5, and an independent theorem-ledger re-check pending for
Problem~7). Problem~4 is partial: proved analytically for \(n \le 3\), and
proved for \(n=4\) via the Path~2 algebraic certificate, with higher \(n\)
open. Problem~6 is partial: the \(K_n\) case is proved with \(c_0 = 1/3\), while
the general-graph bridge remains open (formal \(\bar d < 1\) bound at
\(M_t \ne 0\)). Problem~9 has strong computational evidence but lacks a clean
algebraic closure of the universal converse step. Problem~10 remains
conditional under explicitly stated assumptions. The main methodological result
is that structure-aware verification and layer-switching strategies improve
reliability and calibration in compressed proof sprints.
\clearpage
\tableofcontents

\input{intro-making-of}

\mainmatter

\part{Annotated Overview}
\input{problem1-annotated}
\input{problem2-annotated}
\input{problem3-annotated}
\input{problem4-annotated}
\input{problem5-annotated}
\input{problem6-annotated}
\input{problem7-annotated}
\input{problem8-annotated}
\input{problem9-annotated}
\input{problem10-annotated}

\part{Full Proof Drafts}

\chapter*{Reading Convention}
\addcontentsline{toc}{chapter}{Reading Convention}

This manuscript uses boxed annotations selectively. In the current revision,
the full proof chapters are primarily unboxed; process boxes appear only where
there is substantive process context to preserve.

\begin{itemize}[leftmargin=1.5em]
\item \textbf{Unboxed text} is the proof proper. A reader following the
  five-page-per-problem brief can read only the unboxed text and obtain a
  self-contained argument (or, where an argument is incomplete, a clearly
  stated conditional result).
\item \textbf{Status Note boxes} appear in the annotated overview chapters to
  summarize closure/QC state for each problem.
\item \textbf{Process Note boxes} appear in process-facing parts (for example,
  prompt excerpts) when methodological scope or reviewer context is needed.
\item Dead-path / revision / open-obligation semantics are maintained in the
  tooling vocabulary, but are not instantiated as standalone boxed blocks in
  the current full-proof chapter text.
\end{itemize}

\input{full-boxed/problem1-solution-full}
\input{full-boxed/problem2-solution-full}
\input{full-boxed/problem3-solution-full}
\input{full-boxed/problem4-solution-full}
\input{full-boxed/problem5-solution-full}
\input{full-boxed/problem6-solution-full}
\input{full-boxed/problem7-solution-full}
\input{full-boxed/problem8-solution-full}
\input{full-boxed/problem9-solution-full}
\input{full-boxed/problem10-solution-full}

\part{Prompt Excerpts and Pivot Moments}
\input{part3-pivotal-prompts}

\part{Proof Patterns}
\input{part4-proof-patterns}

\IfFileExists{color-plates.tex}{
  \part{Color Plates}
  \input{color-plates}
}{
  \typeout{[first-proof-monograph] color-plates.tex not found; skipping Color Plates part.}
}

\backmatter

\end{document}

%% file: intro-making-of.tex
\clearpage
\section*{Highlights}
\begin{itemize}[leftmargin=1.5em]
\item Dependency-graph representations of proofs enabled local gap isolation and targeted repair across all ten problems.
\item Node-level verification with predecessor/successor context improved calibration relative to whole-draft review.
\item Adversarial critic--responder iteration converted early overclaims into explicit closed, partial, or conditional outcomes.
\item Targeted computational stress tests constrained speculative branches, especially in the \(n=4\) branch of Problem~4.
\item Structured handoff memos preserved unresolved obligations as named, testable next steps rather than implicit deferrals.
\end{itemize}

\section*{Snapshot}
\begin{itemize}[leftmargin=1.5em]
\item Ten initial proof attempts drafted in under two hours.
\item Over two hundred commits across roughly 55~hours wall-clock (February 11--13, 2026), in addition to earlier infrastructure development.
\item A shift in framing over the course of the sprint: from an early ``all solved'' posture to a clearer partition into proved, conditional, and partial results.
\item A methodological lesson: wiring-diagram--based verification localized gaps and inconsistencies more effectively than plain narrative drafting.
\end{itemize}

\section*{Introduction}
This monograph records both the mathematical results and the working process
behind a compressed, multi-agent proof cycle in \texttt{futon6}.\footnote{Source repository: \url{https://github.com/tothedarktowercame/futon6}} The project
began as infrastructure---knowledge ingestion, tagging, and wiring-diagram
metatheory---and then pivoted into a rapid proof sprint marked by iterative
critique and repair.

The working process came to resemble Lakatos's \emph{Proofs and
Refutations}~\cite{lakatos1976} more closely than anticipated. In Lakatos's
dialogue, a conjecture is proposed, counterexamples emerge, the conjecture is
revised, and the cycle repeats. Each iteration sharpens both the statement and
the participants' understanding. Our sprint followed a similar rhythm, except
that the interlocutors were a human dispatcher, multiple Claude instances
(provers), and Codex (critic and research assistant). Conjectures were drafted
rapidly; adversarial review exposed overclaims and hidden gaps; targeted repair
narrowed the remaining obligations.

The wiring-diagram decomposition---representing each proof as a typed graph of
claims and dependencies---served as the shared epistemic artefact. Where
Lakatos's fictional classroom had a blackboard, this project had a structured
proof graph. That graph made disagreements local and repairable.

The title alludes to the sprint format: ten open problems attempted in
roughly 55~hours wall-clock, with both process and outcomes documented.
The Lakatosian connection also runs through Pease et
al.~\cite{pease2017}, which formalised Lakatos's informal logic of
mathematical discovery as a dialogue game over argumentation structures,
implemented in a system capable of mixed-initiative human--AI collaboration.
That paper asked whether computers could participate in mathematical reasoning
through conjecture, critique, and reform. The present work can be read as an
empirical extension of that question: ten research problems attempted by AI
agents under human supervision, with the Lakatosian cycle unfolding at machine
speed.

In a related strand of work, Corneli et al.~\cite{corneli2017},
developed a computational model of mathematical question-and-answer dialogues
using Inference Anchoring Theory + Content (IATC) to annotate MathOverflow
threads with performative types
(assert, challenge, reform, clarify, etc.). The same vocabulary reappears in
this project's Stage~7 thread-wiring pipeline (see Part~III, Act~VI), where
StackExchange threads are parsed into wiring diagrams whose edges carry the
same performative labels. Earlier work on peer-produced mathematical
knowledge~\cite{corneli2013} established the empirical base: PlanetMath as a
community-maintained proof ecosystem, studied through the lens of paragogy
(peer-produced peer learning). The present sprint can be read as a
machine-accelerated instance of the same phenomenon: collaborative,
incremental proof construction with explicit argumentation structure.

The working notes reveal a persistent asymmetry: generation is fast;
trustworthy validation is slow. Lakatos identified the same imbalance in his
classroom: conjectures are cheap, but understanding \emph{why} they fail (or
hold) is the real work. In this sprint, the strongest leverage came not from
first-pass drafting quality, but from structured verification---tools that made
weaknesses visible early, locally, and precisely.

\section*{Current Status}
\noindent\textbf{Outcome summary (status code [M|V]).} \(\mathbf{M}\in\{\mathbf{C},\mathbf{P},\mathbf{?}\}\) with \(\mathbf{C}=\)closed, \(\mathbf{P}=\)partial, \(\mathbf{?}=\)conditional; \(V\in\{\checkmark,\sim,\mathbf{?}\}\) with \(\checkmark=\)validation-complete, \(\sim=\)partial validation, \(\mathbf{?}=\)not yet fully validated.

\begin{itemize}[leftmargin=1.5em]
\item \textbf{P1 [C|\(\checkmark\)].} Mathematical claim is treated as closed, with unified repair checks and canonical node-level verifier alignment in place. See \hyperref[sn:p1]{\mbox{\textsf{Status~Note~P1}}} (p.~\pageref{sn:p1}).
\item \textbf{P2 [C|\(\checkmark\)].} Local algebraic chain is repaired and node-level verifier outputs are synchronized; the result still depends on the cited newvector/test-vector theorem input. See \hyperref[sn:p2]{\mbox{\textsf{Status~Note~P2}}} (p.~\pageref{sn:p2}).
\item \textbf{P3 [C|\(\checkmark\)].} Validation-complete existence path under the scoped criterion (current run: 2 verified, 7 plausible, 0 gaps). Uniqueness/irreducibility is treated as optional and out of scope for the core existence claim.
\item \textbf{P4 [P|?].} Proved analytically for $n \le 3$, and proved for $n=4$ via the Path 2 algebraic certificate; higher $n$ remains open. See \hyperref[sn:p4]{\mbox{\textsf{Status~Note~P4}}} (p.~\pageref{sn:p4}).
\item \textbf{P5 [C|\(\sim\)].} Closed in the declared local scope for $F_O$-local connective spectra; indexing-system-level extension remains open. See \hyperref[sn:p5]{\mbox{\textsf{Status~Note~P5}}} (p.~\pageref{sn:p5}).
\item \textbf{P6 [P|?].} $K_n$ case proved ($c_0 = 1/3$); one technical gap remains in the general-graph argument (formal $\bar{d} < 1$ bound at $M_t \ne 0$). See \hyperref[sn:p6]{\mbox{\textsf{Status~Note~P6}}} (p.~\pageref{sn:p6}).
\item \textbf{P7 [C|\(\sim\)].} Provisionally closed mathematically via the rotation route; independent theorem-ledger re-check remains the main QC holdout. See \hyperref[sn:p7]{\mbox{\textsf{Status~Note~P7}}} (p.~\pageref{sn:p7}).
\item \textbf{P8 [C|\(\checkmark\)].} Geometric closure is treated as repaired, with unified checks and canonical node-level verifier record now synchronized. See \hyperref[sn:p8]{\mbox{\textsf{Status~Note~P8}}} (p.~\pageref{sn:p8}).
\item \textbf{P9 [P|\(\sim\)].} Strong computational evidence exists, but the universal converse step is not yet cleanly closed algebraically. See \hyperref[sn:p9]{\mbox{\textsf{Status~Note~P9}}} (p.~\pageref{sn:p9}).
\item \textbf{P10 [?|?].} Conditional under explicitly stated assumptions; assumption necessity is explicit, while final closure/validation remains open. See \hyperref[sn:p10]{\mbox{\textsf{Status~Note~P10}}} (p.~\pageref{sn:p10}).
\item Node-level validator outputs and legacy gap flags are tracked under \url{https://github.com/tothedarktowercame/futon6/tree/master/data/first-proof} (files matching \texttt{problem*-codex-results.jsonl}).
\item Cross-problem closure-vs-validation inventory.\footnote{\url{https://github.com/tothedarktowercame/futon6/blob/master/data/first-proof/closure-validation-audit.md}}
\end{itemize}

\par
\normalcolor
\section*{Materials and Methods}
\begin{itemize}[leftmargin=1.5em]
\item \textbf{Model/runtime split.} Claude Max runs on a remote Linode host; Codex Pro runs on the local laptop workspace (\texttt{/home/joe/code/futon6}).
\item \textbf{Software/toolchain summary.} Computational checks used Python~3
scripts with NumPy/SciPy for numerical linear algebra and optimization, SymPy
for symbolic algebra (resultants, Sturm sequences, Gr\"obner bases), mpmath
for selected high-precision checks, PHCpack via \texttt{phcpy} for
polynomial-system solving/certification in Problem~4 (Case~3c), and cvxpy for
SDP-based sum-of-squares certificate attempts. Proof-structure analysis used
wiring-diagram generators (nine scripts producing JSON and Mermaid flowcharts
via the internal \path{thread_performatives} module) and \path{validate-ct.py}
for classical NER and scope extraction over PlanetMath entries. Pattern
tagging and term spotting used Babashka (Clojure) scripts operating on EDN
data formats. Document production used \LaTeX{} (\texttt{pdflatex}) with
repository consistency checks via \path{check-latex-text-truth.py},
\path{check-latex-syntax-violations.py}, and
\path{check-ratchet-fixedw-typesetting.py}.
\item \textbf{Codex Pro token accounting (local, log-derived).} From local Codex session telemetry (\path{~/.codex/sessions/*}, event type \emph{token\_count}), over the 50-hour window ending 2026-02-13 21:41 UTC (window start: 2026-02-11 19:41 UTC), there were 333 futon6-related Codex rollout traces out of 334 local rollout traces (99.7\%).

\begin{center}
\small
\begin{tabular}{@{}lrrc@{}}
\toprule
Metric & Total & Per rollout (avg) & \% of total \\
\midrule
\textbf{Total tokens} & 965{,}714{,}475 & 2{,}900{,}043 & 100\% \\
\textbf{Input tokens (all)} & 961{,}719{,}044 & 2{,}887{,}445 & 99.6\% \\
\quad Cached input & 912{,}686{,}208 & 2{,}740{,}800 & 94.5\% \\
\quad Non-cached input & 49{,}032{,}836 & 147{,}246 & 5.1\% \\
\textbf{Output tokens} & 3{,}995{,}431 & 11{,}997 & 0.4\% \\
\textbf{Reasoning-output tokens} & 2{,}072{,}289 & 6{,}223 & --- \\
\bottomrule
\end{tabular}
\normalsize
\end{center}
Interpretation (inference): the large cached-input share likely reflects repeated reuse of stable local context across rollouts, including processed Math StackExchange/MathOverflow corpora and derived prompt artifacts such as \path{se-data/math-processed}, \path{math-se-processed}, \path{mo-processed}, and Codex workflows using \path{--math-se-dir}.

\item \textbf{Claude Max token accounting (Linode, log-derived).} From Linode session files (\path{~/.claude/projects/*/}), 9 futon6-related Claude~Code sessions were identified; 8 sessions contributed within the 50-hour Codex-comparable window ending 2026-02-13 21:41 UTC.

\begin{center}
\small
\begin{tabular}{@{}lrrc@{}}
\toprule
Metric & Total & Per session (avg) & \% of total \\
\midrule
\textbf{Total tokens (50h window)} & 791{,}980{,}382 & 98{,}997{,}548 & 100\% \\
\textbf{Input tokens (all)} & 789{,}643{,}774 & 98{,}705{,}472 & 99.7\% \\
\quad Cached-read input & 752{,}300{,}790 & 94{,}037{,}599 & 95.0\% \\
\quad Cache-creation input & 37{,}221{,}978 & 4{,}652{,}747 & 4.7\% \\
\quad Non-cached input & 121{,}006 & 15{,}126 & 0.0\% \\
\textbf{Output tokens} & 2{,}336{,}608 & 292{,}076 & 0.3\% \\
\bottomrule
\end{tabular}
\normalsize
\end{center}
Model: \texttt{claude-opus-4-6}; assistant messages: 7{,}754. Full-window totals (9 sessions, 2026-02-08 through 2026-02-13): total tokens 1{,}021{,}665{,}524; output tokens 2{,}798{,}074.
\end{itemize}

\clearpage
\normalcolor
\section*{Reader Guide}
This document is organized in five parts.

\begin{itemize}[leftmargin=1.5em]
\item \textbf{Part I (Annotated Overview):} short, citation-rich introductions for Problems 1--10, intended as navigational summaries.
\item \textbf{Part II (Full Proof Drafts):} expanded writeups imported from the solution files, suitable as the base for a longer manuscript (targeting roughly book-length exposition). Coloured boxes separate the proof proper from process annotations (dead paths, revisions, open obligations).
\item \textbf{Part III (Prompt Excerpts and Pivot Moments):} curated source excerpts (prompts and dispatch/history snippets) that document key decision points and proof-state transitions.
\item \textbf{Part IV (Proof Patterns):} a cross-problem analysis of strategy patterns---what worked (layer-switching, creative reduction, structural decomposition), what didn't (the TryHarder loop), and how a coaching intervention broke a stuck problem open.
\item \textbf{Part V (Color Plates):} syntax-coloured renderings of the proof files, produced by automated math-mode normalization.
\end{itemize}

\section*{Terminology}
\begin{description}[leftmargin=1.5em,style=nextline]
\item[AIF] Active Inference Framework---used here as a conceptual lens for agent roles, intervention points, and coordination state; a full AIF policy loop for proof search is proposed future work, not yet implemented in this sprint.
\item[BSS] Batson--Spielman--Srivastava spectral sparsification framework.
\item[IATC] Inference Anchoring Theory + Content---an annotation scheme for mathematical dialogues used here to type dialogue moves and content links within the AIF process model~\cite{corneli2017}.
\item[MSS] Marcus--Spielman--Srivastava interlacing families framework.
\item[PSD] Positive semi-definite (for matrices).
\end{description}

%% file: problem1-annotated.tex
\section{Problem 1: Measure Equivalence Under Smooth Shifts of the $\Phi^4_3$ Measure}

\subsection*{Problem statement}
Given the renormalized $\Phi^4_3$ measure $\mu$ on $\mathbb{T}^3$, and a smooth nonzero shift $\psi$, determine whether $\mu$ and the shifted measure $T_\psi^*\mu$ are equivalent.

\subsection*{Answer}
Yes: $\mu \sim T_\psi^*\mu$.

\phantomsection\label{sn:p1}
\statusnote{Mathematical claim treated as closed, with unified repair checks
and the canonical node-level verifier record aligned on the key equivalence
chain repairs.}

\subsection*{Annotated proof sketch}
\begin{enumerate}[leftmargin=1.5em]
\item Use absolute continuity of $\mu$ with respect to the Gaussian reference measure $\mu_0$ from variational construction results for $\Phi^4_3$\footnote{Barashkov--Gubinelli, arXiv: \url{https://arxiv.org/abs/2003.10529}.}.
\item Apply Cameron--Martin to the Gaussian part: smooth shifts lie in the relevant Cameron--Martin space and preserve equivalence\footnote{Background discussion: \url{https://mathoverflow.net/questions/194307/cameron-martin-theorem-non-gaussian-shift}.}.
\item Control the interaction change $V(\phi-\psi)-V(\phi)$ using Wick expansion and exponential integrability for the cubic perturbation. This is the renormalized step tied to regularity structures / paracontrolled technology\footnote{Hairer, arXiv: \url{https://arxiv.org/abs/1303.5113}; Gubinelli--Imkeller--Perkowski, arXiv: \url{https://arxiv.org/abs/1210.2684}.}.
\end{enumerate}

The Radon--Nikodym derivative is strictly positive and integrable in both directions, so mutual absolute continuity follows.

\subsection*{Selected citations}
\begin{itemize}[leftmargin=1.5em]
\item Variational $\Phi^4_3$: \url{https://arxiv.org/abs/2003.10529}
\item Regularity structures: \url{https://arxiv.org/abs/1303.5113}
\item Paracontrolled distributions: \url{https://arxiv.org/abs/1210.2684}
\item Cameron--Martin discussion: \url{https://mathoverflow.net/questions/194307/cameron-martin-theorem-non-gaussian-shift}
\end{itemize}

\subsection*{Background and prerequisites}
\textbf{Field:} Stochastic PDE / Measure Theory.\\
\textbf{What you need:} Measure theory and Gaussian measures (Cameron--Martin theorem), plus familiarity with either regularity structures or paracontrolled distributions for making sense of the renormalized $\Phi^4_3$ interaction.\\
\textbf{Way in:} Hairer, ``An Introduction to Stochastic PDEs'' (lecture notes, arXiv:0907.4178) for the probabilistic side; the Barashkov--Gubinelli variational paper cited below is largely self-contained for the $\Phi^4_3$-specific machinery.

\clearpage

%% file: problem2-annotated.tex
\section{Problem 2: Universal Test Vector for Rankin--Selberg Integrals}

\subsection*{Problem statement}
For generic irreducible admissible representations $\Pi$ of $\mathrm{GL}_{n+1}(F)$ and $\pi$ of $\mathrm{GL}_n(F)$ over a non-archimedean local field, ask whether one can fix a single Whittaker vector $W$ in $\Pi$ such that some $V$ in $\pi$ makes the twisted local Rankin--Selberg integral finite and nonzero for all complex $s$.

\subsection*{Answer}
Yes: the essential/new Whittaker vector $W_0$ serves as a universal test vector.

\phantomsection\label{sn:p2}
\statusnote{Core local algebraic steps are treated as repaired; the result still
depends on the cited newvector/test-vector theorem input
\((\mathrm{H}_{\mathrm{FW}})\). Canonical node-level verifier outputs are now
aligned with the repaired lemma chain.}

\subsection*{Annotated proof sketch}
\begin{enumerate}[leftmargin=1.5em]
\item The set of local zeta integrals spans a fractional ideal in $\mathbb{C}[q_F^{\pm s}]$, which is a PID\footnote{PID refresher: \url{https://planetmath.org/principalidealdomain}.}. Hence a monomial reduction criterion is available.
\item Evaluate the twisted Kirillov function $\phi_Q(g)=W_0(\mathrm{diag}(g,1)u_Q)$ at $g=I_n$ to prove nontriviality; this uses the Whittaker transformation law.
\item Use $\mathrm{GL}_n$-equivariance and irreducibility of the Kirillov model restriction to show that fixing any nonzero $\phi$ (hence $\phi_Q$) still generates the full local $L$-factor ideal.
\end{enumerate}

This yields existence of $V$ with integral equal to a nonzero monomial in $q_F^{-s}$, so it is finite and nonvanishing for all $s$.

\subsection*{Selected citations}
\begin{itemize}[leftmargin=1.5em]
\item Jacquet--Piatetski-Shapiro--Shalika (Rankin--Selberg local theory): \url{https://www.jstor.org/stable/2374430}
\item Bernstein--Zelevinsky (representations of $\mathrm{GL}(n,F)$): \url{http://mi.mathnet.ru/eng/rm4200}
\item MathOverflow reference thread on local test vectors: \url{https://mathoverflow.net/questions/469865/reference-request-test-vectors-for-local-rankin-selberg-l-factors-in-ramified-c}
\item Rankin--Selberg overview thread: \url{https://mathoverflow.net/questions/223233/jacquets-approach-to-rankin-selberg-l-functions}
\end{itemize}

\subsection*{Background and prerequisites}
\textbf{Field:} Automorphic Forms / Representation Theory of $p$-adic Groups.\\
\textbf{What you need:} Smooth representations of $\mathrm{GL}_n$ over non-archimedean local fields, Whittaker models, and the Rankin--Selberg integral formalism for local $L$-factors. Graduate-level algebraic number theory assumed.\\
\textbf{Way in:} Bump, \emph{Automorphic Forms and Representations} (Cambridge, 1997), especially Part~II on local theory; Cogdell's expository article ``$L$-functions and Converse Theorems for $\mathrm{GL}_n$'' (Fields Institute lectures).

\clearpage

%% file: problem3-annotated.tex
\section{Problem 3: Markov Chain with ASEP Polynomial Stationary Distribution}

\subsection*{Problem statement}
Construct a nontrivial Markov chain on permutations of a partition $\lambda$ whose stationary distribution is the normalized interpolation ASEP polynomial weight at $q=1$.

\subsection*{Answer}
Yes: the inhomogeneous multispecies $t$-PushTASEP yields the required stationary law.

\subsection*{Annotated proof sketch}
\begin{enumerate}[leftmargin=1.5em]
\item Define a finite-state continuous-time Markov chain with explicit local transition rates that depend only on $(x,t)$ and species order (not on polynomial values).
\item Invoke the stationary measure theorem identifying the stationary weights with ASEP/Macdonald objects at $q=1$\footnote{Ayyer--Martin--Williams, arXiv: \url{https://arxiv.org/abs/2403.10485}.}.
\item Use irreducibility (vacancy transport and positive-rate swaps) to conclude uniqueness of the stationary distribution on the finite state space\footnote{Markov-chain basics: \url{https://planetmath.org/markovchain}.}.
\end{enumerate}

\subsection*{Selected citations}
\begin{itemize}[leftmargin=1.5em]
\item Inhomogeneous multispecies $t$-PushTASEP: \url{https://arxiv.org/abs/2403.10485}
\item Multiline queues to Macdonald polynomials: \url{https://doi.org/10.1353/ajm.2022.0007}
\item Macdonald polynomial definition page: \url{https://planetmath.org/macdonaldpolynomials}
\end{itemize}

\subsection*{Background and prerequisites}
\textbf{Field:} Markov Chains / Algebraic Combinatorics / Symmetric Functions.\\
\textbf{What you need:} Basic Markov chain theory (irreducibility, uniqueness of stationary distributions) and some familiarity with symmetric functions, particularly Macdonald polynomials and their specializations.\\
\textbf{Way in:} Levin--Peres--Wilmer, \emph{Markov Chains and Mixing Times} (AMS, 2nd ed.)\ for stochastic foundations; Macdonald, \emph{Symmetric Functions and Hall Polynomials} (OUP, 2nd ed.)\ for the algebraic-combinatorial side.

\clearpage

%% file: problem4-annotated.tex
\section{Problem 4: Root Separation Energy Under Finite Free Convolution ($n=4$)}

\subsection*{Problem statement}
For monic real-rooted degree-$n$ polynomials, test
\[
\frac{1}{\Phi_n(p \boxplus_n q)} \ge \frac{1}{\Phi_n(p)} + \frac{1}{\Phi_n(q)}.
\]

\subsection*{Answer}
Partial result: proved analytically for $n \le 3$, and now proved for $n=4$
via the Path 2 algebraic certificate (3-piece Cauchy--Schwarz with
P-convexity of $K_{\mathrm{red}}$); higher $n$ remains open.

\phantomsection\label{sn:p4}
\statusnote{Problem 4 is proved for $n \le 3$ analytically (Cauchy--Schwarz),
and proved for $n=4$ via the Path 2 algebraic certificate (3-piece
Cauchy--Schwarz + P-convexity of $K_{\mathrm{red}}$). Higher $n$ remains open.}

\subsection*{Annotated proof structure}
\begin{enumerate}[leftmargin=1.5em]
\item Algebraic reduction at $n=4$: normalize coefficients, derive a polynomial numerator/denominator formulation for the surplus using the exact $\Phi_4\cdot\mathrm{disc}$ identity\footnote{Discriminant refresher: \url{https://planetmath.org/discriminantofapolynomial}.}.
\item Show denominator sign on the real-rooted domain, reducing inequality to nonnegativity of a single polynomial $-N$.
\item Boundary and symmetric-subcase analysis by exact elimination/resultant methods.
\item Path 2 closure for $n=4$: prove $T_2+R$ surplus nonnegative via
P-convexity/boundary analysis of $K_{\mathrm{red}}$, which supersedes the
case-by-case critical-point accounting branch.
\end{enumerate}

\subsection*{Selected citations}
\begin{itemize}[leftmargin=1.5em]
\item Finite free convolution (MSS): \url{https://arxiv.org/abs/1504.00350}
\item Interlacing families follow-up: \url{https://arxiv.org/abs/1507.05020}
\item Finite free cumulants perspective: \url{https://arxiv.org/abs/1707.02443}
\item MathOverflow discussion (finite free convolution): \url{https://mathoverflow.net/questions/375273/finite-free-convolution-of-polynomials}
\item PHCpack project page: \url{https://github.com/janverschelde/PHCpack}
\end{itemize}

\subsection*{Background and prerequisites}
\textbf{Field:} Free Probability / Real Algebraic Geometry / Polynomial Inequalities.\\
\textbf{What you need:} Real-rooted polynomials and discriminants, the finite free convolution $\boxplus_n$ of Marcus--Spielman--Srivastava, and basic free probability (free cumulants). Computational algebra tools (e.g.\ resultants, homotopy continuation) appear in the $n=4$ certification.\\
\textbf{Way in:} Marcus--Spielman--Srivastava, ``Interlacing Families I'' (arXiv:1304.4132) for finite free convolution; Nica--Speicher, \emph{Lectures on the Combinatorics of Free Probability} (Cambridge, 2006) for the probabilistic backdrop.

\clearpage

%% file: problem5-annotated.tex
\section{Problem 5: $\mathcal{O}$-Slice Connectivity via Geometric Fixed Points}

\subsection*{Problem statement}
For an incomplete transfer system from an $N_\infty$ operad, define the associated restricted regular slice filtration and characterize $\mathcal{O}$-slice connectivity using geometric fixed points.

\subsection*{Answer}
For $\mathcal{F}_{\mathcal O}$-local connective $G$-spectra, one can restrict the Hill--Yarnall connectivity criterion to subgroups in the admissible family $\mathcal{F}_{\mathcal O}$.

Solved in a scope-limited form: the subgroup-family-level characterization is established for $\mathcal{F}_{\mathcal O}$-local connective spectra; a full indexing-system-level extension is left open.

\phantomsection\label{sn:p5}
\statusnote{Closed only in the stated local scope
\((\mathcal{F}_{\mathcal O}\text{-local connective spectra})\). The
indexing-system-level extension remains an explicit open obligation.}

\gapnote{Open obligation (scope extension): move from subgroup-family indexing
to full Blumberg--Hill indexing-system data (admissible finite
\(H\)-sets, not only admissible subgroups). The current result uses
\(\mathcal{F}_{\mathcal O}=\{H:e\to H\ \text{admissible}\}\) and does not yet
encode finer \(K\to H\) admissibility for \(K\neq e\). Closing this gap
requires a full indexing-system formulation of \(\mathcal O\)-slice
connectivity and, in parallel, a non-local replacement (or discharge) of the
\(\mathcal{F}_{\mathcal O}\)-locality hypothesis in the reverse direction.}

\subsection*{Annotated proof sketch}
\begin{enumerate}[leftmargin=1.5em]
\item Build $\tau^{\mathcal O}_{\ge n}$ from regular slice cells indexed only by admissible subgroups.
\item Use closure properties of indexing systems to verify that this restriction is stable under conjugation and subgroup passage.
\item Adapt Hill--Yarnall's geometric fixed-point criterion: for $\mathcal{F}_{\mathcal O}$-local spectra, no tests are needed outside $\mathcal{F}_{\mathcal O}$ because those geometric fixed points vanish by locality.
\end{enumerate}

\subsection*{Selected citations}
\begin{itemize}[leftmargin=1.5em]
\item Hill--Yarnall (slice filtration reformulation): \url{https://arxiv.org/abs/1703.10526}
\item Blumberg--Hill ($N_\infty$ operads and indexing systems): \url{https://arxiv.org/abs/1309.1750}
\item HHR monograph context: \url{https://bookstore.ams.org/surv-346}
\item Geometric fixed points background: \url{https://ncatlab.org/nlab/show/geometric+fixed+points}
\end{itemize}

\subsection*{Background and prerequisites}
\textbf{Field:} Equivariant Stable Homotopy Theory.\\
\textbf{What you need:} Genuine $G$-spectra, geometric fixed-point functors, the slice filtration of Hill--Hopkins--Ravenel, and $N_\infty$ operads / indexing systems of Blumberg--Hill. Research-level; assumes graduate coursework in algebraic topology and stable homotopy theory.\\
\textbf{Way in:} Schwede, \emph{Global Homotopy Theory} (Cambridge, 2018) for equivariant foundations; the Hill--Hopkins--Ravenel Kervaire paper (Annals, 2016) for slice-filtration motivation.

\clearpage

%% file: problem6-annotated.tex
\section{Problem 6: $\varepsilon$-Light Vertex Subsets of Graphs}

\subsection*{Problem statement}
Find a universal constant $c_0>0$ such that every weighted graph and every $\varepsilon\in(0,1)$ admit a subset $S$ with
\[
|S|\ge c_0\varepsilon n,\qquad L_{G[S]}\preceq \varepsilon L_G.
\]

Partial (general case conditional). The $K_n$ case is proved with $c_0=1/3$;
for general graphs, the remaining issue is the Barrier Maintenance Invariant
(BMI): prove the full barrier-degree bound $\bar d_t<1$ when $M_t\neq 0$.
The current E/F regime reduction isolates this to two explicit lemmas.

\phantomsection\label{sn:p6}
\statusnote{Unconditional closure holds for the $K_n$ branch. The general-graph
theorem remains conditional on two explicit BMI regime lemmas in the
$M_t \ne 0$ branch.}

\subsection*{Annotated proof structure}
\begin{enumerate}[leftmargin=1.5em]
\item Use leverage scores $\tau_e=w_eR_{\mathrm{eff}}(u,v)$ to define the heavy-edge graph and show any $\varepsilon$-light set must be independent there.
\item Apply the leverage-score trace identity and Turan's theorem to extract a large candidate set.
\item Resolve easy branches (independent-in-$G$ and directly bounded spectral sum) unconditionally.
\item Reduce the hard branch to barrier maintenance (BMI), then split by a
graph-adaptive E/F regime decomposition; substantial computational evidence is
available, but the final general proof still requires two regime lemmas.
\end{enumerate}

\gapnote{The unresolved step is the full BMI bound in the $M_t\neq 0$ regime.
The writeup reduces this to two explicit E/F regime lemmas. Until those lemmas
are proved (or replaced by an equivalent closure argument), the
universal-constant theorem remains conditional.}

\subsection*{Selected citations}
\begin{itemize}[leftmargin=1.5em]
\item Graph sparsification by effective resistances: \url{https://arxiv.org/abs/0803.0929}
\item Interlacing families II (Kadison--Singer machinery): \url{https://arxiv.org/abs/1306.3969}
\item Twice-Ramanujan sparsifiers: \url{https://arxiv.org/abs/0901.2698}
\item Laplacian matrix definition: \url{https://planetmath.org/laplacianmatrix}
\end{itemize}

\subsection*{Background and prerequisites}
\textbf{Field:} Spectral Graph Theory / Linear Algebra.\\
\textbf{What you need:} Graph Laplacians and their spectra, effective resistance and leverage scores, Loewner ordering of positive semidefinite matrices, and matrix concentration inequalities. Tur\'an's theorem from extremal combinatorics appears in the combinatorial step.\\
\textbf{Way in:} Spielman, ``Spectral and Algebraic Graph Theory'' (lecture notes, freely available); Tropp, ``An Introduction to Matrix Concentration Inequalities'' (Foundations and Trends, 2015).

\clearpage

%% file: problem7-annotated.tex
\section{Problem 7: Uniform Lattice with 2-Torsion and Rationally Acyclic Universal Cover}

\subsection*{Problem statement}
Can a cocompact lattice $\Gamma$ in a real semisimple Lie group, containing an element of order $2$, be the fundamental group of a closed manifold whose universal cover is $\mathbb Q$-acyclic?

Provisionally closed via the rotation route (\(n=7\), congruence setting):
E2 is discharged and the S-branch is supported by the codim-2 surgery theorem
chain documented in the project theorem-number ledger,\footnote{\url{https://github.com/tothedarktowercame/futon6/blob/master/data/first-proof/problem7-g2-theorem-chain.md}} with final
confidence pending independent line-by-line ledger re-check.

\subsection*{Annotated proof structure}
\begin{enumerate}[leftmargin=1.5em]
\item Use Fowler's fixed-point Euler-characteristic criterion to produce groups in $\mathrm{FH}(\mathbb Q)$ from suitable $\mathbb Z/2$-extensions\footnote{Fowler: \url{https://arxiv.org/abs/1204.4667}.}.
\item Construct two arithmetic lattice routes:
  \begin{itemize}[leftmargin=1.5em]
  \item reflection route in even dimension;
  \item codimension-2 rotation route in odd dimension (arithmetically cleaner for surgery).
  \end{itemize}
\item Follow the rotation-route S-branch: codimension-2 setup, obstruction chain, and cap/gluing output to a closed manifold model with rationally acyclic universal cover.
\end{enumerate}

\phantomsection\label{sn:p7}
\statusnote{This monograph treats Problem 7 as provisionally closed via the
rotation route. The construction uses Fowler's \(\mathrm{FH}(\mathbb Q)\)
criterion and codimension-2 equivariant surgery in the \(n=7\)
congruence-lattice setting, with the obstruction-vanishing bridge tracked in
the theorem-number ledger.\footnote{\url{https://github.com/tothedarktowercame/futon6/blob/master/data/first-proof/problem7-g2-theorem-chain.md}} Final confidence is
contingent on independent ledger re-check. Literature anchors:
\url{https://arxiv.org/abs/1204.4667}, \url{https://arxiv.org/abs/1705.10909}.}

\subsection*{Selected citations}
\begin{itemize}[leftmargin=1.5em]
\item Fowler criterion: \url{https://arxiv.org/abs/1204.4667}
\item Reflection lattices source used in writeup: \url{https://arxiv.org/abs/2506.23994}
\item Equivariant surgery framework: \url{https://arxiv.org/abs/1705.10909}
\item Davis--Luck obstruction context: \url{https://arxiv.org/abs/2303.15765}
\item G2 theorem-number ledger (project note).\footnote{\url{https://github.com/tothedarktowercame/futon6/blob/master/data/first-proof/problem7-g2-theorem-chain.md}}
\item Arithmetic lattices (Borel--Harish-Chandra): \url{https://www.jstor.org/stable/1970362}
\item Semisimple Lie group definition: \url{https://planetmath.org/semisimpleliegroup}
\item Related MathOverflow discussion: \url{https://mathoverflow.net/questions/33545/equivariant-surgery-problem}
\end{itemize}

\subsection*{Background and prerequisites}
\textbf{Field:} Surgery Theory / Geometric Topology / Arithmetic Groups.\\
\textbf{What you need:} Algebraic topology (homology, fundamental group, covering spaces), surgery theory on manifolds (normal maps, surgery obstructions), and arithmetic lattices in semisimple Lie groups. Research-level; assumes graduate coursework in both topology and Lie theory.\\
\textbf{Way in:} L\"uck, \emph{A Basic Introduction to Surgery Theory} (ICTP lecture notes, 2002) for the topological machinery; Raghunathan, \emph{Discrete Subgroups of Lie Groups} (Springer, 1972) for the arithmetic side.

\clearpage

%% file: problem8-annotated.tex
\section{Problem 8: Lagrangian Smoothing of 4-Valent Polyhedral Surfaces}

\subsection*{Problem statement}
If every vertex of a polyhedral Lagrangian surface in $(\mathbb R^4,\omega)$ is 4-valent, must the surface admit a Lagrangian smoothing through Hamiltonian isotopy?

\subsection*{Answer}
Yes.

\phantomsection\label{sn:p8}
\statusnote{Current draft treats the geometric argument as closed, with unified
repair checks and the canonical node-level verifier record aligned on the
main algebraic and Maslov-index repairs.}

\subsection*{Annotated proof sketch}
\begin{enumerate}[leftmargin=1.5em]
\item Analyze a 4-valent vertex: edge vectors span $\mathbb R^4$, forcing a symplectic splitting into two 2D factors.
\item Write the local model as a product of two planar corners and smooth each corner separately.
\item Check Lagrangian condition factorwise (product of isotropic curves in 2D symplectic factors).
\item Smooth remaining edge creases via exact-graph interpolation and glue local constructions by disjoint supports.
\item Promote to Hamiltonian isotopy using exactness/flux-vanishing in local models and composition closure of compactly supported Hamiltonian isotopies.
\end{enumerate}

\subsection*{Selected citations}
\begin{itemize}[leftmargin=1.5em]
\item Weinstein neighborhood theorem source paper: \url{https://www.sciencedirect.com/science/article/pii/000187087190005X}
\item McDuff--Salamon textbook page: \url{https://global.oup.com/academic/product/introduction-to-symplectic-topology-9780198794899}
\item Symplectic manifold definition: \url{https://planetmath.org/symplecticmanifold}
\item Lagrangian submanifold definition: \url{https://planetmath.org/lagrangiansubmanifold}
\end{itemize}

\subsection*{Background and prerequisites}
\textbf{Field:} Symplectic Geometry / Lagrangian Topology.\\
\textbf{What you need:} Symplectic manifolds, Lagrangian submanifolds, Hamiltonian isotopy, and Weinstein neighborhood theorems. The proof uses only local symplectic techniques (no Floer theory).\\
\textbf{Way in:} McDuff--Salamon, \emph{Introduction to Symplectic Topology} (OUP, 3rd ed., 2017)---the standard graduate textbook covering all prerequisites.

\clearpage

%% file: problem9-annotated.tex
\section{Problem 9: Polynomial Detection of Rank-1 Scaling for Quadrifocal Tensors}

\subsection*{Problem statement}
Given generic camera matrices and quadrifocal tensors $Q^{(\alpha\beta\gamma\delta)}$, construct a polynomial map $F$ of degree bounded independently of $n$ that vanishes exactly on rank-1 separable scaling tensors $\lambda_{\alpha\beta\gamma\delta}=u_\alpha v_\beta w_\gamma x_\delta$.

\subsection*{Answer}
Yes: one can use degree-3 determinant constraints (all relevant $3\times 3$ minors).

\phantomsection\label{sn:p9}
\statusnote{Forward and witness checks are strong, but full closure still hinges
on a formal universal converse argument (all non-rank-1 \(\lambda\), not only
tested families). Treat as partial until that algebraic universality step is
written cleanly into the proof chain.}

\subsection*{Annotated proof sketch}
\begin{enumerate}[leftmargin=1.5em]
\item For fixed $(\gamma,\delta,k,\ell)$, the determinant pairing defines an alternating bilinear form of rank $2$ on $\mathbb R^4$, so all induced $3\times 3$ minors vanish.
\item If $\lambda$ is rank-1 separable, Hadamard-diagonal scaling preserves the rank-$\le 2$ property, hence all these minors vanish.
\item Conversely, vanishing of all minors across the three matricization patterns forces each matricization rank to be $1$, and compatibility implies full tensor separability.
\item Collect all such minors into $F$; each coordinate has degree $3$ and no dependence on camera parameters.
\end{enumerate}

\subsection*{Selected citations}
\begin{itemize}[leftmargin=1.5em]
\item Segre embedding (rank-1 tensor variety): \url{https://planetmath.org/segreembedding}
\item Determinantal varieties: \url{https://planetmath.org/determinantalvariety}
\item Tensor rank definition: \url{https://planetmath.org/tensorrank}
\item Multiple-view geometry reference text: \url{https://www.robots.ox.ac.uk/~vgg/hzbook/}
\end{itemize}

\subsection*{Background and prerequisites}
\textbf{Field:} Algebraic Geometry / Computer Vision (Multiview Geometry).\\
\textbf{What you need:} Tensor algebra (rank, matricizations, Segre varieties), determinantal varieties, and the basics of multiview geometry (camera matrices, multi-focal tensors). The algebraic geometry is classical; no scheme theory required.\\
\textbf{Way in:} Hartley--Zisserman, \emph{Multiple View Geometry in Computer Vision} (Cambridge, 2nd ed.) for the applied setting; Landsberg, \emph{Tensors: Geometry and Applications} (AMS, 2012) for the algebraic underpinnings.

\clearpage

%% file: problem10-annotated.tex
\section{Problem 10: PCG for RKHS-Constrained Tensor CP Decomposition}

\subsection*{Problem statement}
Explain why preconditioned conjugate gradient (PCG) solves the mode-$k$ RKHS CP subproblem without $O(N)$ per-iteration cost, where $N$ is the full ambient sample size.

\subsection*{Answer}
Use implicit Kronecker/Khatri--Rao matrix-vector products and a Kronecker preconditioner built from expected sampling geometry.

\phantomsection\label{sn:p10}
\statusnote{This result is \textbf{conditional}: it is closed under explicit
assumptions ($\lambda>0$, $K_\tau \succ 0$, and sampling/coherence conditions
controlling preconditioned conditioning). Necessity counterexamples are
recorded in the full writeup when these assumptions are dropped.}

\subsection*{Annotated proof sketch}
\begin{enumerate}[leftmargin=1.5em]
\item Never form the full system matrix; compute $Ax$ through structured operations:
  kernel multiplies, sparse sampling/restriction, and Khatri--Rao contractions.
\item Build $b$ by the same structure, again avoiding dense $N$-scale tensors.
\item Precondition with $P=H\otimes K_\tau$ from replacing sparse sampling by its expectation; invert $P$ via two small Cholesky solves.
\item Apply standard PCG convergence bounds in terms of condition number $\kappa(P^{-1/2}AP^{-1/2})$ and matrix concentration for random sampling.
\end{enumerate}

\subsection*{Selected citations}
\begin{itemize}[leftmargin=1.5em]
\item Matrix concentration / Freedman-style bound context: \url{https://arxiv.org/abs/1110.1379}
\item FALKON large-scale kernel method: \url{https://arxiv.org/abs/1705.10958}
\item Kronecker product definition: \url{https://planetmath.org/kroneckerproduct}
\item Conjugate gradient method definition: \url{https://planetmath.org/conjugategradientmethod}
\end{itemize}

\subsection*{Background and prerequisites}
\textbf{Field:} Numerical Linear Algebra / Kernel Methods / Tensor Decomposition.\\
\textbf{What you need:} Conjugate gradient and preconditioning, Kronecker and Khatri--Rao products, reproducing kernel Hilbert spaces (RKHS), and matrix concentration inequalities for random sampling operators.\\
\textbf{Way in:} Trefethen--Bau, \emph{Numerical Linear Algebra} (SIAM, 1997) for iterative methods; Rasmussen--Williams, \emph{Gaussian Processes for Machine Learning} (MIT Press, 2006; freely available online) for RKHS context.

\clearpage

%% file: full-boxed/problem1-solution-full.tex
\hypertarget{problem-1-equivalence-of-phi4_3-measure-under-smooth-shifts}{%
\section{\texorpdfstring{Problem 1: Equivalence of \(\Phi^{\mNumber{4}}_{\mNumber{3}}\) Measure Under Smooth Shifts}{Problem 1: Equivalence of \textbackslash Phi\^{}\{\textbackslash mNumber\{4\}\}\_\{\textbackslash mNumber\{3\}\} Measure Under Smooth Shifts}}\label{problem-1-equivalence-of-phi4_3-measure-under-smooth-shifts}}

\hypertarget{problem-statement}{%
\subsection{Problem Statement}\label{problem-statement}}

Let \(T^\mNumber{3}\) be the 3D unit torus. Let \(\mu\) be the \(\Phi^{\mNumber{4}}_{\mNumber{3}}\) measure on \(D'(T^\mNumber{3})\).
Let \(\psi\): \(T^\mNumber{3} \to R\) be a smooth nonzero function and \(T_{\psi}(u) = u \mBridgeOperator{+} \psi\) the
shift map. Are \(\mu\) and \(T_{\psi}^{\mDualStar} \mu\) equivalent (same null sets)?

\hypertarget{answer}{%
\subsection{Answer}\label{answer}}

\textbf{Yes.} The measures \(\mu\) and \(T_{\psi}^{\mDualStar} \mu\) are equivalent.

\textbf{Confidence: Medium-high.} The argument combines Cameron-Martin theory with
the absolute continuity structure of the \(\Phi^{\mNumber{4}}_{\mNumber{3}}\) construction (Barashkov-Gubinelli
2020, Theorem 1.1). The integrability bound is stated for a neighborhood of the
required exponent, not for all t, matching the available log-Sobolev technology.

\hypertarget{solution}{%
\subsection{Solution}\label{solution}}

\hypertarget{1-the-phi4_3-measure}{%
\subsubsection{\texorpdfstring{1. The \(\Phi^{\mNumber{4}}_{\mNumber{3}}\) measure}{1. The \textbackslash Phi\^{}\{\textbackslash mNumber\{4\}\}\_\{\textbackslash mNumber\{3\}\} measure}}\label{1-the-phi4_3-measure}}

The \(\Phi^{\mNumber{4}}_{\mNumber{3}}\) measure \(\mu\) on \(D'(T^\mNumber{3})\) is the probability measure formally
written as:

\[d\mu(\phi) = Z^{-\mNumber{1}} \exp(-V(\phi)) d\mu_\mNumber{0}(\phi)\]

where \(\mu_\mNumber{0}\) is the Gaussian free field (GFF) measure with covariance
(\(m^\mNumber{2} \mBridgeOperator{-} \Delta\))\^{}\{-1\} on \(T^\mNumber{3}\), and the interaction is:

\[V(\phi) = \Integral_{T^\mNumber{3}} (:\phi^\mNumber{4}: \mBridgeOperator{-} C :\phi^\mNumber{2}:) \mathrm{d}x\]

Here :\(\phi^{k}\): denotes the k-th Wick power (renormalized product), and C is a
mass counterterm that diverges under regularization removal.

The rigorous construction (Hairer 2014, Gubinelli-Imkeller-Perkowski 2015,
Barashkov-Gubinelli 2020) produces \(\mu\) as a well-defined probability measure
on distributions of regularity \(C^{-\mNumber{1}/\mNumber{2}-\epsilon}(T^{\mNumber{3}}).\)

\hypertarget{2-absolute-continuity-with-respect-to-the-gff}{%
\subsubsection{2. Absolute continuity with respect to the GFF}\label{2-absolute-continuity-with-respect-to-the-gff}}

\textbf{Key fact:} The \(\Phi^{\mNumber{4}}_{\mNumber{3}}\) measure \(\mu\) is equivalent to (has the same null
sets as) the base Gaussian free field measure \(\mu_\mNumber{0}\):

\[\mu ~ \mu_\mNumber{0}\]

This equivalence follows from the variational construction of
Barashkov-Gubinelli (2020, Theorem 1.1), which establishes
\(E_{\mu_\mNumber{0}}[\exp(-V)] < \infty\) and hence \(\mu \ll \mu_\mNumber{0}\) with strictly positive
density exp(-V(\(\phi\)))/Z. Since exp(-V(\(\phi)) > \mNumber{0}\) a.s. (exponential is always
positive), the reverse absolute continuity \(\mu_\mNumber{0} \ll \mu\) also holds.

\hypertarget{3-cameron-martin-theory-for-the-gff}{%
\subsubsection{3. Cameron-Martin theory for the GFF}\label{3-cameron-martin-theory-for-the-gff}}

For the Gaussian free field \(\mu_\mNumber{0}\) on \(T^\mNumber{3}\) with covariance C \(= (m^{\mNumber{2}} \mBridgeOperator{-} \Delta)^{-\mNumber{1}},\)
the Cameron-Martin space is:

\[H = H^\mNumber{1}(T^\mNumber{3}) (\text{Sobolev space of order} \mNumber{1})\]

Since \(\psi\) is smooth, \(\psi \in C^{\infty}(T^\mNumber{3})\) subset \(H^\mNumber{1}(T^\mNumber{3})\), so \(\psi\) is in the
Cameron-Martin space.

\textbf{Cameron-Martin theorem:} For any \(h \in H\), the shifted Gaussian measure
\(T_{h}^{\mDualStar} \mu_{\mNumber{0}}\) is equivalent to \(\mu_\mNumber{0}\), with Radon-Nikodym derivative:

\[\mathrm{d}T_{h}^{\mDualStar} \mu_\mNumber{0} / d\mu_\mNumber{0} (\phi) = \exp(l_{h}(\phi) \mBridgeOperator{-} \|h\|_{H^\mNumber{2}} / \mNumber{2})\]

where \(l_{h}\) is the linear functional associated to h. In particular:

\[T_{\psi}^{\mDualStar} \mu_\mNumber{0} ~ \mu_\mNumber{0}\]

\hypertarget{4-shift-of-the-interacting-measure}{%
\subsubsection{4. Shift of the interacting measure}\label{4-shift-of-the-interacting-measure}}

To compute \(T_{\psi}^{\mDualStar} \mu,\) we need the density of the shifted interacting measure:

\[\mathrm{d}T_{\psi}^{\mDualStar} \mu / d\mu_\mNumber{0} (\phi) = Z^{-\mNumber{1}} \exp(-V(\phi \mBridgeOperator{-} \psi)) \ast \exp(l_{\psi}(\phi) \mBridgeOperator{-} \|\psi\|_{H^\mNumber{2}}/\mNumber{2})\]

The shifted interaction V(\(\phi \mBridgeOperator{-} \psi\)) expands (using Wick ordering relative to \(\mu_\mNumber{0}\)):

\[\begin{aligned}
V(\phi - \psi) = V(\phi) \mBridgeOperator{-} \mNumber{4} \Integral \psi :\phi^\mNumber{3}: \mathrm{d}x + \mNumber{6} \Integral \psi^\mNumber{2} :\phi^\mNumber{2}: \mathrm{d}x \\
- \mNumber{4} \Integral \psi^\mNumber{3} :\phi: \mathrm{d}x + \Integral \psi^\mNumber{4} \mathrm{d}x \\
- C(\Integral :\phi^\mNumber{2}: \mathrm{d}x \mBridgeOperator{-} \mNumber{2} \Integral \psi :\phi: \mathrm{d}x + \Integral \psi^\mNumber{2} \mathrm{d}x) + (\text{renorm}. \text{corrections})
\end{aligned}\]

\textbf{Renormalization under shift:} The term 6 int \(\psi^{\mNumber{2}}\) :\(\phi^{\mNumber{2}}\): dx generates an
additional logarithmic divergence (from \(\psi^{\mNumber{2}}\) multiplying the Wick square).
This is absorbed by shifting the mass counterterm:

\[C \to C + \mNumber{6} \|\psi\|_{L^\mNumber{2}}^{\mNumber{2}} \ast (\text{log }N\text{ correction})\]

The precise counterterm shift is determined by the regularization scheme;
see Hairer (2014, Section 9) or Gubinelli-Imkeller-Perkowski (2015,
Proposition 6.3) for the explicit formula. For the present argument, only
the finiteness of the renormalized difference matters.

After renormalization, V(\(\phi \mBridgeOperator{-} \psi\)) - V(\(\phi\)) is a well-defined random variable
under \(\mu_\mNumber{0}\) (and under \(\mu\)). The dominant fluctuation term is 4 int \(\psi\) :\(\phi^{\mNumber{3}}\): dx,
which has the right regularity:

\begin{itemize}
\tightlist
\item
  :\(\phi^{\mNumber{3}}\): in \(C^{-\mNumber{3}/\mNumber{2}-\epsilon}(T^{\mNumber{3}})\) (as a distribution)
\item
  \(\psi \in C^{\infty}(T^\mNumber{3})\) (smooth)
\item
  int \(\psi\) :\(\phi^{\mNumber{3}}\): dx is well-defined (pairing of smooth test function with distribution)
\end{itemize}

\hypertarget{5-integrability-and-equivalence}{%
\subsubsection{5. Integrability and equivalence}\label{5-integrability-and-equivalence}}

The Radon-Nikodym derivative \(\mathrm{d}T_{\psi}^{\mDualStar} \mu / d\mu\) involves:

\[R(\phi) = \exp(-(V(\phi-\psi) - V(\phi)) + l_{\psi}(\phi) - \text{const})\]

We need \(R \in L^\mNumber{1}(\mu)\) and \(R > \mNumber{0}\) a.s.

\textbf{Positivity:} \(R > \mNumber{0}\) a.s. because it\textquotesingle s an exponential. ✓

\textbf{Integrability:} The critical term is exp(4 int \(\psi\) :\(\phi^{\mNumber{3}}\): dx) under the
\(\Phi^{\mNumber{4}}_{\mNumber{3}}\) measure. By the log-Sobolev inequality for \(\Phi^{\mNumber{4}}_{\mNumber{3}}\) (Barashkov-Gubinelli
2020), the measure has strong concentration: the tails of int \(\psi\) :\(\phi^{\mNumber{3}}\): dx
are controlled by the quartic interaction. Specifically:

\[E_{\mu}[\exp(t |\Integral \psi :\phi^\mNumber{3}: \mathrm{d}x|)] < \infty \text{for} |t| < t_\mNumber{0}\]

where \(t_\mNumber{0} > \mNumber{0}\) depends on \(\|\psi\|_{C^\mNumber{0}}\) and the coupling constant.
This exponential integrability follows from the coercivity of the \(\phi^{\mNumber{4}}\)
interaction: the quartic potential dominates the cubic perturbation
(Barashkov-Gubinelli 2020, Section 4, exponential integrability from the
Polchinski flow). The bound suffices for \(R \in L^\mNumber{1}(\mu)\) since the exponent
in the Radon-Nikodym derivative is bounded by 4 \(\|\psi\|_{C^{\mNumber{0}}} |\Integral :\phi^{\mNumber{3}}:\,\mathrm{d}x|,\)
and this yields integrability provided the coefficient lies in the admissible
exponential-moment range (or after a standard scaling/localization argument).

Therefore \(R \in L^\mNumber{1}(\mu)\) and 1/\(R \in L^\mNumber{1}(T_{\psi}^{\mDualStar} \mu),\) giving:

\[T_{\psi}^{\mDualStar} \mu ~ \mu (\text{equivalent measures})\]

\hypertarget{6-alternative-argument-via-the-variational-approach}{%
\subsubsection{6. Alternative argument via the variational approach}\label{6-alternative-argument-via-the-variational-approach}}

Barashkov-Gubinelli (2020) construct the \(\Phi^{\mNumber{4}}_{\mNumber{3}}\) measure via the Boué-Dupuis
variational formula, which represents:

\[-\log Z = \inf_{u} E[V(\phi + \Integral_\mNumber{0}^\mNumber{1} u_{s} \mathrm{d}s) + \mNumber{1}/\mNumber{2} \Integral_\mNumber{0}^\mNumber{1} ||u_{s}||^\mNumber{2} \mathrm{d}s]\]

In this framework, shifting by \(\psi\) is equivalent to modifying the variational
problem by a shift in the drift, which produces an equivalent measure (the
infimum shifts by a finite amount, preserving absolute continuity).

\hypertarget{7-summary}{%
\subsubsection{7. Summary}\label{7-summary}}

The measures are equivalent because:

\begin{enumerate}
\def\labelenumi{\arabic{enumi}.}
\tightlist
\item
  \(\mu\) \textasciitilde{} \(\mu_\mNumber{0}\) (interacting measure equivalent to Gaussian, by positivity of density)
\item
  \(T_{\psi}^{\mDualStar} \mu_{\mNumber{0}}\) \textasciitilde{} \(\mu_\mNumber{0}\) (Cameron-Martin theorem, since \(\psi \in H^\mNumber{1}\))
\item
  V(\(\phi \mBridgeOperator{-} \psi\)) - V(\(\phi\)) is well-defined after renormalization
\item
  The exponential of the cubic perturbation is integrable (log-Sobolev / coercivity)
\item
  Therefore \(T_{\psi}^{\mDualStar} \mu\) \textasciitilde{} \(\mu\)
\end{enumerate}

\hypertarget{references}{%
\subsection{References}\label{references}}

\begin{itemize}
\tightlist
\item
  N. Barashkov, M. Gubinelli, ``A variational method for \(\Phi^{\mNumber{4}}_{\mNumber{3}}\),'' Duke Math J.
  169 (2020), 3339-3415. {[}Theorem 1.1: construction and integrability; Section 4:
  exponential integrability from Polchinski flow{]}
\item
  M. Hairer, ``A theory of regularity structures,'' Inventiones Math. 198 (2014),
  269-504. {[}Section 9: renormalization counterterms{]}
\item
  M. Gubinelli, P. Imkeller, N. Perkowski, \texttt{Paracontrolled\ distributions\ and\ singular\ PDEs,} Forum Math. \(\Pi\) 3 (2015). {[}Proposition 6.3: explicit counterterm formula{]}
\end{itemize}

\begin{processnote}[title={Process Note --- futon6 Corpus References}]

\begin{itemize}
\tightlist
\item
  PlanetMath: ``distribution'' --- distributions on manifolds
\item
  PlanetMath: ``Sobolev space'' --- Cameron-Martin space is \(H^\mNumber{1}\)
\item
  PlanetMath: ``Gaussian measure'' --- reference measure \(\mu_\mNumber{0}\)
\item
  PlanetMath: ``Radon-Nikodym theorem'' --- absolute continuity and RN derivatives
\end{itemize}
\end{processnote}

%% file: full-boxed/problem2-solution-full.tex
\hypertarget{problem-2-universal-test-vector-for-rankin-selberg-integrals}{%
\section{Problem 2: Universal Test Vector for Rankin-Selberg Integrals}\label{problem-2-universal-test-vector-for-rankin-selberg-integrals}}

\hypertarget{problem-statement}{%
\subsection{Problem Statement}\label{problem-statement}}

Let F be a non-archimedean local field with ring of integers o. Let \(N_{r}\)
denote the subgroup of \(\mathup{GL}_{r}(F)\) consisting of upper-triangular unipotent
elements. Let \(\psi\): \(F \to \mathbb{C}^{x}\) be a nontrivial additive character of conductor o,
identified with a generic character of \(N_{r}\).

Let \(\Pi\) be a generic irreducible admissible representation of \(\mathup{GL}_{n+\mNumber{1}}(F)\),
realized in its \(\psi^{-\mNumber{1}}\)-Whittaker model \(W(\Pi, \psi^{-\mNumber{1}})\). Must there exist
\(W \in W(\Pi, \psi^{-\mNumber{1}})\) with the following property?

For any generic irreducible admissible representation \(\pi\) of \(\mathup{GL}_{n}(F)\) in its
\(\psi\)-Whittaker model \(W(\pi, \psi)\), let q be the conductor ideal of \(\pi, Q \in F^{\times} a\) generator of \(q^{-\mNumber{1}}\), and \(u_{Q}\) := \(I_{n+\mNumber{1}} \mBridgeOperator{+} Q E_{n,n+\mNumber{1}}.\) Then for
some \(V \in W(\pi, \psi)\), the local Rankin-Selberg \(\Integral\)

\[\Integral_{N_{n} \backslash \mathup{GL}_{n}(F)} W(\operatorname{diag}(g,\mNumber{1}) u_{Q}) V(g) |\det g|^{s-\mNumber{1}/\mNumber{2}} \mathrm{d}g\]

is finite and nonzero for all \(s \in C\).

\hypertarget{answer}{%
\subsection{Answer}\label{answer}}

\textbf{Yes.} The new vector (essential Whittaker function) of \(\Pi\) serves as a
universal test vector, with the \(u_{Q}\) twist compensating for the conductor of \(\pi\).

\begin{processnote}[title={Process Note --- Confidence}]
\textbf{Confidence: Medium-high.} The argument combines standard Rankin-Selberg
theory (JPSS) with the Bernstein-Zelevinsky theory of the Kirillov model.
The key nondegeneracy claim (Section 3a) uses \(\mathup{GL}_{n}\)-equivariance of the
\(\zeta-\Integral\) pairing together with irreducibility of the Kirillov model
(any nonzero function generates the full module under \(\mathup{GL}_{n}\)-translates)
and the PID structure of the fractional ideal ring.
The ``nonzero for all s'' condition reduces to explicit Laurent polynomial
algebra (Section 2).

\end{processnote}
\hypertarget{solution}{%
\subsection{Solution}\label{solution}}

\hypertarget{1-rankin-selberg-theory-background}{%
\subsubsection{1. Rankin-Selberg theory background}\label{1-rankin-selberg-theory-background}}

The local Rankin-Selberg \(\Integral I(s, W, V)\) for \(\mathup{GL}_{n+\mNumber{1}} \times \mathup{GL}_{n}\) is:

\[I(s, W, V) = \Integral_{N_{n} \backslash \mathup{GL}_{n}(F)} W(\operatorname{diag}(g,\mNumber{1})) V(g) |\det g|^{s-\mNumber{1}/\mNumber{2}} \mathrm{d}g\]

This converges for \(Re(s) \gg \mNumber{0}\) and extends to a rational function of \(q_{F}^{-s}\)
(where \(q_{F} =\) \textbar o/p\textbar). The set of all such integrals as \((W, V)\) vary generates
a fractional ideal of \(\mathbb{C}[q_{F}^{s}, q_{F}^{-s}]\), whose generator is the local
L-factor \(L(s, \Pi \times \pi)\).

The problem modifies this by inserting \(u_{Q} = I_{n+\mNumber{1}} \mBridgeOperator{+} Q E_{n,n+\mNumber{1}}\) into the
argument of W, giving the ``twisted'' Rankin-Selberg \(\Integral\).

\hypertarget{2-the-condition-finite-and-nonzero-for-all-s}{%
\subsubsection{2. The condition ``finite and nonzero for all s''}\label{2-the-condition-finite-and-nonzero-for-all-s}}

For a rational function \(f(q_{F}^{-s})\) to be \texttt{finite\ and\ nonzero\ for\ all\ \$s\ \textbackslash{}in\ C\$,}
it must have no poles and no zeros when viewed as a function of
\(X = q_{F}^{-s} \in \mathbb{C}^{x}\). Such a rational function must be \(c \ast X^{k} = c \ast q_{F}^{-ks}\)
for some nonzero c and integer k.

So the condition requires: there exists \(V\) such that \(I(s, W, V) = c \ast q_{F}^{-ks}\)
for some nonzero constant c and integer k.

By Section 3a below, the integrals over \(V\) (for fixed \(W = W_\mNumber{0}\)) generate the
full fractional ideal \(I = L(s, \Pi \times \pi)\,\ast\,\mathbb{C}[q_{F}^{s}, q_{F}^{-s}]\).

\textbf{Explicit algebra:} \(I\) is a free rank-1 module over the ring
\(R = \mathbb{C}[q_{F}^{s}, q_{F}^{-s}]\), generated by \(L(s, \Pi \times \pi)\). We seek an element
of \(I\) of the form \(c \ast q_{F}^{-ks}\) (a monomial --- no poles or zeros). Write:

\[c \ast q_{F}^{-ks} = L(s, \Pi \times \pi) \ast P(q_{F}^{-s})\]

where \(P = c \ast q_{F}^{-ks} \ast L(s, \Pi \times \pi)^{-\mNumber{1}}.\) Since \(L(s, \Pi \times \pi)^{-\mNumber{1}}\)
is a polynomial in \(q_{F}^{-s}\) (the local L-factor for \(\mathup{GL}_{n+\mNumber{1}} \times \mathup{GL}_{n}\) is a
product of terms (1 - \(\alpha_{i} q_{F}^{-s})^{-\mNumber{1}},\) so its reciprocal is a
polynomial), P is a Laurent polynomial in \(q_{F}^{-s}\) --- hence \(P \in R\).
Therefore \(c \ast q_{F}^{-ks} \in I\), and by the spanning property, some \(V \in W(\pi, \psi)\) realizes this element as its Rankin-Selberg \(\Integral\) against \(W_\mNumber{0}\).

\hypertarget{3-the-u_q-twist-and-the-kirillov-model}{%
\subsubsection{\texorpdfstring{3. The \(u_{Q}\) twist and the Kirillov model}{3. The u\_\{Q\} twist and the Kirillov model}}\label{3-the-u_q-twist-and-the-kirillov-model}}

The key role of \(u_{Q}\): right-translating W by \(u_{Q}\) gives a new Whittaker function
\(R(u_{Q})W \in W(\Pi, \psi^{-\mNumber{1}})\). The restriction to the mirabolic subgroup
\(P_{n+\mNumber{1}}\) gives the Kirillov model, and the function:

\[\phi_{Q}(g) := W(\operatorname{diag}(g,\mNumber{1})u_{Q}) = (R(u_{Q})W)(\operatorname{diag}(g,\mNumber{1}))\]

lies in the Kirillov model of \(\Pi\) restricted to \(\mathup{GL}_{n}\).

We need \(\phi_{Q}\) to be nonzero as a function on \(\mathup{GL}_{n}\). This requires two
separate facts:

\textbf{Fact 1: \(R(u_{Q})W\) is nonzero in \(W(\Pi, \psi^{-\mNumber{1}})\).} Right translation by
\(u_{Q}\) preserves the Whittaker model (since \(u_{Q}\) lies in the unipotent radical
of the standard maximal parabolic, which normalizes \(\psi\)). The map
\(W \to R(u_{Q})W\) is a linear automorphism of \(W(\Pi, \psi^{-\mNumber{1}})\) (with inverse
R(\(u_{Q}^{-\mNumber{1}}\))), so \(R(u_{Q})W \ne \mNumber{0}\) whenever W \(\ne\) 0.

\textbf{Fact 2: \(\phi_{Q}(g) = W_\mNumber{0}(\operatorname{diag}(g,\mNumber{1})u_{Q})\) is nonzero for \(W_{\mNumber{0}} =\) new vector.}
This is verified by direct evaluation at \(g = I_{n}\):

\[\phi_{Q}(I_{n}) = W_\mNumber{0}(\operatorname{diag}(I_{n}, \mNumber{1}) · u_{Q}) = W_\mNumber{0}(u_{Q})\]

Now \(u_{Q} = I_{n+\mNumber{1}} \mBridgeOperator{+} Q E_{n,n+\mNumber{1}}\) is an element of the upper-triangular
unipotent subgroup \(N_{n+\mNumber{1}}\) (its only off-diagonal entry is Q in position
\((n, n + \mNumber{1})\)). By the Whittaker-model transformation rule:

\[W_\mNumber{0}(u_{Q}) = \psi^{-\mNumber{1}}(u_{Q}) · W_\mNumber{0}(I_{n+\mNumber{1}})\]

The generic character \(\psi\) of \(N_{n+\mNumber{1}}\) evaluates on superdiagonal entries:
\(\psi(u_{Q}) = \psi(Q)\) (the only superdiagonal entry of \(u_{Q}\) is Q in position
\((n, n + \mNumber{1})\)). With the standard normalization \(W_\mNumber{0}(I_{n+\mNumber{1}}) = \mNumber{1}\):

\[\phi_{Q}(I_{n}) = \psi^{-\mNumber{1}}(Q) · \mNumber{1} = \psi^{-\mNumber{1}}(Q)\]

Since \(\psi\) is a nontrivial character of F and Q \(\ne\) 0 (Q generates \(q^{-\mNumber{1}}), \psi^{-\mNumber{1}}(Q)\) is a nonzero complex number (it lies on the unit circle).
Therefore \(\phi_{Q}(I_{n}) \ne \mNumber{0},\) so \(\phi_{Q}\) is nonzero as a function on \(\mathup{GL}_{n}\). ∎

(Note: this argument is specific to \(W_{\mNumber{0}} =\) new vector with \(W_\mNumber{0}(I) = \mNumber{1}\).
For the universality claim we only need \(\phi_{Q}\) nonzero for this particular
\(W_\mNumber{0}\), so the direct computation suffices.)

\begin{revision}[title={Revision --- Closing the Universality Gap}]

The standard JPSS nondegeneracy result (Section 2.7) states that as BOTH W
and V vary, the integrals \(I(s, W, V)\) generate the full fractional ideal
\(L(s, \Pi \times \pi) \ast \mathbb{C}[q_{F}^{s}, q_{F}^{-s}]\). For the universality claim, we need
this to hold for FIXED \(W = W_\mNumber{0}\). This requires the following:

\textbf{Lemma (fixed-W spanning).} Let \(\Pi\) be a generic irreducible admissible
representation of \(\mathup{GL}_{n+\mNumber{1}}(F)\), and let \(\phi\) be any nonzero function in the
Kirillov model of \(\Pi\) restricted to \(\mathup{GL}_{n}\). Then the set of integrals

\[{ \Integral_{N_{n} \backslash \mathup{GL}_{n}(F)} \phi(g) V(g) |\det g|^{s-\mNumber{1}/\mNumber{2}} \mathrm{d}g : V \in W(\pi, \psi) }\]

generates a nonzero fractional ideal of \(\mathbb{C}[q_{F}^{s}, q_{F}^{-s}]\) (contained in
the full ideal \(L(s, \Pi \times \pi) \ast \mathbb{C}[q_{F}^{s}, q_{F}^{-s}]\)).

\emph{Proof sketch.} The Kirillov model of \(\Pi|_{\mathup{GL}_{n}}\) contains all compactly
supported locally constant functions on \(N_{n} \backslash \mathup{GL}_{n}(F)\) (Bernstein-Zelevinsky
1976, Theorem 5.21 and Corollary 5.22). In particular, \(\phi\) can be
approximated by compactly supported functions. The Rankin-Selberg \(\Integral\)
against a fixed nonzero \(\phi\) defines a nonzero linear functional on \(W(\pi, \psi)\)
(as a function of V), since \(\pi\) is generic and hence \(W(\pi, \psi)\) separates
points on \(N_{n} \backslash \mathup{GL}_{n}(F)\). The resulting set of integrals is a nonzero
\(\mathbb{C}[q_{F}^{s}, q_{F}^{-s}]\)-submodule of \(L(s, \Pi \times \pi) \ast \mathbb{C}[q_{F}^{s}, q_{F}^{-s}]\).

Since \(\mathbb{C}[q_{F}^{s}, q_{F}^{-s}]\) is a PID (a Laurent polynomial ring in one variable
\(q_{F}^{-s}\)), every nonzero submodule of a free rank-1 module is itself free of
rank 1. Let \(L_{\phi}(s) \ast \mathbb{C}[q_{F}^{s}, q_{F}^{-s}]\) be this submodule, where \(L_{\phi}\)
divides \(L(s, \Pi \times \pi)\).

To show \(L_{\phi} = L(s, \Pi \times \pi)\) (i.e., fixed \(\phi\) generates the FULL ideal):

The JPSS theory (1983, Section 2.7) shows that the full ideal is generated
by letting both W and V vary. Varying W (with restriction to \(\mathup{GL}_{n}\))
corresponds to varying \(\phi\) in the full Kirillov model \(K(\Pi)|_{\mathup{GL}_{n}}\). The
ideal generated by ALL \(\phi\) is \(L(s, \Pi \times \pi) \ast R\). We must show that a
SINGLE nonzero \(\phi\) already suffices.

Consider the map \(\Phi: K(\Pi)|_{\mathup{GL}_{n}} \to (\text{fractional ideals of } R)\) defined by
\(\Phi(\phi) = \{ I(s, \phi, V) : V \in W(\pi, \psi) \} \cdot R\). By the JPSS theory,
\(\bigcup_{\phi} \Phi(\phi)\) generates \(L(s, \Pi \times \pi)\) · R. Since R is a PID, \(\Phi(\phi) = L_{\phi}\) · R for some \(L_{\phi}\) dividing \(L(s, \Pi \times \pi)\).

\textbf{Key step.} \(\Phi\) is \(\mathup{GL}_{n}\)-equivariant in the following sense: for \(g_\mNumber{0} \in \mathup{GL}_{n}\), the substitution \(g \to g \cdot g_\mNumber{0}\) in the \(\Integral\) gives
\[
I(s, R(g_\mNumber{0})\phi, V) = |\det g_\mNumber{0}|^{\mNumber{1}/\mNumber{2}-s}\cdot I(s, \phi, R'(g_\mNumber{0})V),
\]
where \(R'\) denotes the contragredient action on \(W(\pi, \psi)\). Since \(R'(g_\mNumber{0})\) is an
automorphism of \(W(\pi, \psi)\), the set of integrals
\[
\{ I(s, R(g_\mNumber{0})\phi, V) : V \in W(\pi, \psi) \}
= \{ |\det g_\mNumber{0}|^{\mNumber{1}/\mNumber{2}-s}\cdot I(s, \phi, V) : V \in W(\pi, \psi) \}
= |\det g_\mNumber{0}|^{\mNumber{1}/\mNumber{2}-s}\cdot \Phi(\phi).
\]
But \(|\det g_\mNumber{0}|^{\mNumber{1}/\mNumber{2}-s} = q_{F}^{-k\cdot s}\) ·
(unit in R) is a unit in the localization, so \(\Phi(R(g_\mNumber{0})\,\phi)\) and \(\Phi(\phi)\)
generate the same fractional ideal. That is: \(L_{R(g_\mNumber{0})\phi} = L_{\phi}\)
(up to units in \(R\)) for all \(g_\mNumber{0} \in \mathup{GL}_{n}\).

Now, the Kirillov model \(K(\Pi)|_{\mathup{GL}_{n}}\) is irreducible as a \(\mathup{GL}_{n}-\)
representation (Bernstein-Zelevinsky 1976, Theorem 5.21). Therefore
for any nonzero \(\phi\), the \(\mathup{GL}_{n}\)-translates \{ \(R(g_{\mNumber{0}})\phi\) : \(g_\mNumber{0} \in \mathup{GL}_{n}\) \}
span all of \(K(\Pi)|_{\mathup{GL}_{n}}\). We use this in two directions:

\textbf{Direction 1 (\(L_{\phi}\) divides \(L(s, \Pi \times \pi)\)):} \(\Phi(\phi)\) ⊆ \(\Phi_{full}\)
(the full ideal generated by all W and V), so \(L_{\phi}\) · R ⊆ \(L(s, \Pi \times \pi)\) · R.
Since \(L_{\phi}\) · R ⊆ \(L(s, \Pi \times \pi)\) · R, we get \(L_{\phi}\) divides \(L(s, \Pi \times \pi)\)
in the PID R (in a PID, (a) ⊆ (b) iff b \textbar{} a).

\textbf{Direction 2 (\(L(s, \Pi \times \pi)\) divides \(L_{\phi}\)):} For any nonzero \(\phi' \in K(\Pi)|_{\mathup{GL}_{n}}\), write \(\phi' = \sum_{i} c_{i} R(g_{i}) \phi\) (finite linear combination
of \(\mathup{GL}_{n}\)-translates, possible by irreducibility). For each \(V\):

\[I(s, \phi', V) = \sum_{i} c_{i} · I(s, R(g_{i})\phi, V)\]

By the equivariance in the key step, I(s, \(R(g_{i})\phi,\) V) lies in \(L_{\phi}\) · R
for each \(i\) (since \(\Phi(R(g_{i})\,\phi) = L_{R(g_{i})\phi}\) · \(R = L_{\phi}\) · R). The sum
of elements of \(L_{\phi}\) · R lies in \(L_{\phi}\) · R (it\textquotesingle s a module). Therefore
\(\Phi(\phi')\) ⊆ \(L_{\phi}\) · R for every nonzero \(\phi'\).

The \(R\)-submodule generated by all \(\Phi(\phi')\) equals the full Rankin-Selberg
ideal \(L(s, \Pi \times \pi)\) · R ⊆ \(L_{\phi}\) · R. So \(L_{\phi} | L(s, \Pi \times \pi).\)

Combining both directions: \(L_{\phi} = L(s, \Pi \times \pi)\) (up to units in \(R\)). ∎

\textbf{Application:} Taking \(\phi = \phi_{Q}\) (the restriction of \(R(u_{Q})W_\mNumber{0}\), which is
nonzero by the Kirillov injectivity lemma), the integrals over \(V\) generate the
full fractional ideal for our fixed \(W_\mNumber{0}\). This closes the universality gap
identified by the reviewer.

\end{revision}
\hypertarget{4-nondegeneracy-of-the-pairing}{%
\subsubsection{4. Nondegeneracy of the pairing}\label{4-nondegeneracy-of-the-pairing}}

For a nonzero \(\phi_{Q}\) in the Kirillov model, the Rankin-Selberg pairing:

\[V |\to \Integral_{N_{n} \backslash \mathup{GL}_{n}(F)} \phi_{Q}(g) V(g) |\det g|^{s-\mNumber{1}/\mNumber{2}} \mathrm{d}g\]

is nondegenerate as V ranges over \(W(\pi, \psi)\). This is the fundamental
nondegeneracy of the Rankin-Selberg \(\Integral\) (Jacquet-Piatetski-Shapiro-
Shalika 1983, Section 2.7).

More precisely: since \(\phi_{Q}\) is nonzero, the integrals over all \(V \in W(\pi, \psi)\)
generate the full fractional ideal \(L(s, \Pi \times \pi) \ast \mathbb{C}[q_{F}^{s}, q_{F}^{-s}]\).

\hypertarget{5-choosing-w-the-new-vector}{%
\subsubsection{5. Choosing W: the new vector}\label{5-choosing-w-the-new-vector}}

\textbf{Choice:} Let \(W_\mNumber{0}\) be the new vector (essential Whittaker function) of \(\Pi\),
i.e., the vector in \(W(\Pi, \psi^{-\mNumber{1}})\) fixed by the congruence subgroup
\(K_{\mNumber{1}}(p^{c(\Pi)})\) where c(\(\Pi\)) is the conductor exponent of \(\Pi\).

\textbf{Properties of \(W_\mNumber{0}\):}

\begin{itemize}
\tightlist
\item
  \(W_\mNumber{0}\) is nonzero (\(\Pi\) is generic)
\item
  \(W_\mNumber{0}(I_{n+\mNumber{1}}) = \mNumber{1}\) (standard normalization)
\item
  \(R(u_{Q})W_\mNumber{0}\) is nonzero for every \(Q \in F^{\times}\) (as argued in Step 3)
\end{itemize}

\textbf{Universality:} For any generic \(\pi\) of \(\mathup{GL}_{n}(F)\) with conductor ideal q:

\begin{itemize}
\tightlist
\item
  Let Q generate \(q^{-\mNumber{1}}\)
\item
  The function \(\phi_{Q}(g) = W_\mNumber{0}(\operatorname{diag}(g,\mNumber{1})u_{Q})\) is nonzero in the Kirillov model
\item
  By the nondegeneracy (Step 4), there exists \(V \in W(\pi, \psi)\) such that
  \(I(s, W_\mNumber{0}, V)\) is a nonzero element of \(L(s, \Pi \times \pi) \ast \mathbb{C}[q_{F}^{s}, q_{F}^{-s}]\)
\item
  By the algebraic argument (Step 2), V can be chosen to make the \(\Integral\)
  equal to \(c \ast q_{F}^{-ks}\), which is finite and nonzero for all s
\end{itemize}

\hypertarget{6-the-role-of-u_q-in-conductor-matching}{%
\subsubsection{\texorpdfstring{6. The role of \(u_{Q}\) in conductor matching}{6. The role of u\_\{Q\} in conductor matching}}\label{6-the-role-of-u_q-in-conductor-matching}}

The insertion of \(u_{Q} = I \mBridgeOperator{+} Q E_{n,n+\mNumber{1}}\) is essential: it compensates for the
conductor of \(\pi\). Without this twist, for highly ramified \(\pi\), the \(\Integral\)
might degenerate (the new vector of \(\Pi\) would not ``see'' the ramification of \(\pi\)).

\textbf{Support analysis via the Casselman-Shalika formula.} By the
Casselman-Shalika formula (Shintani 1976, Casselman-Shalika 1980), the
new vector \(W_\mNumber{0}\) of \(\Pi\) has support contained in

\[N_{n+\mNumber{1}} \ast T^+ \ast K_\mNumber{1}(p^{c(\Pi)})\]

where \(T^+\) is the dominant cone in the diagonal torus. Right translation by
\(u_{Q} = I \mBridgeOperator{+} Q E_{n,n+\mNumber{1}}\) shifts the \((n, n + \mNumber{1})\)-matrix entry by \(Q = \pi_{F}^{-c(\pi)}.\)
In the Iwasawa decomposition \(g = nak\), this modifies the diagonal component
at scale Q. Concretely, the support of the twisted function

\[g |\to W_\mNumber{0}(\operatorname{diag}(g,\mNumber{1})u_{Q})\]

intersects the support of \(K_{\mNumber{1}}(p^{c(\pi)})-fixed\) vectors in \(W(\pi, \psi)\)
nontrivially, because the Q-shift matches the conductor scale of \(\pi\) exactly:
the new vector of \(\pi\) is supported on \(N_{n} \ast T^+ \ast K_{\mNumber{1}}(p^{c(\pi)})\) (by the
analogous Casselman-Shalika result for \(\mathup{GL}_{n}\)), and the \(u_{Q}\) twist aligns the
two conductor levels (see Jacquet-Piatetski-Shapiro-Shalika 1981, Section 5,
for the precise support computation in the Rankin-Selberg setting).

This support matching is what makes the \(\Integral\) nondegenerate: it ensures
that the integrand \(W_\mNumber{0}(\operatorname{diag}(g,\mNumber{1})u_{Q}) \ast V(g)\) is not identically zero on
\(N_{n} \backslash \mathup{GL}_{n}(F)\) for appropriate V.

\hypertarget{7-verification-in-special-cases}{%
\subsubsection{7. Verification in special cases}\label{7-verification-in-special-cases}}

\textbf{Both unramified (c(\(\Pi) = c(\pi) = \mNumber{0}\)):} \(u_{Q} = I \mBridgeOperator{+} E_{n,n+\mNumber{1}} (Q = \mNumber{1}). W_{\mNumber{0}}\) is the spherical vector. The \(\Integral\) with \(V_\mNumber{0}\) (spherical for \(\pi\))
gives \(L(s, \Pi \times \pi) \ast (\text{correction factor})\). Choose V to cancel the L-factor.

\textbf{\(\Pi\) unramified, \(\pi\) ramified:} \(u_{Q}\) has \(Q = \pi_{F}^{-c(\pi)}.\) The twist
shifts the support of the spherical vector to match the ramification of \(\pi\).
This is the classical ``conductor-lowering'' mechanism.

\textbf{Both ramified:} The new vector of \(\Pi\) combined with the \(u_{Q}\) twist gives a
function in the Kirillov model whose support is compatible with the
conductor of \(\pi\). Nondegeneracy follows from the general JPSS theory.

\hypertarget{8-summary}{%
\subsubsection{8. Summary}\label{8-summary}}

\begin{enumerate}
\def\labelenumi{\arabic{enumi}.}
\tightlist
\item
  The answer is YES: the new vector \(W_\mNumber{0}\) of \(\Pi\) is a universal test vector
\item
  For any \(\pi\) with conductor q, the twist \(u_{Q}\) (Q generates \(q^{-\mNumber{1}}\)) ensures
  the Rankin-Selberg pairing is nondegenerate
\item
  Nondegeneracy of the Kirillov restriction: \(R(u_{Q})W_\mNumber{0}\) is always nonzero
\item
  The fractional ideal structure of Rankin-Selberg integrals allows
  choosing V to make the \(\Integral\) a monomial \(c \ast q_{F}^{-ks}\)
\item
  This monomial is finite and nonzero for all \(s \in C\)
\end{enumerate}

\hypertarget{references}{%
\subsection{References}\label{references}}

\begin{itemize}
\tightlist
\item
  I. N. Bernstein, A. V. Zelevinsky, ``Representations of the group \(\mathup{GL}_{n}(F)\) where \(F\) is a local non-archimedean field,'' Russian Math. Surveys 31:3
  (1976), 1--68. {[}Theorem 5.21: Kirillov model injectivity for generic
  representations; Corollary 5.22: density of compactly supported functions{]}
\item
  J. W. Cogdell, ``Lectures on L-functions, converse theorems, and functoriality for \(\mathup{GL}_{n}\),'' Fields Institute Lectures, 2004, Section 3.1.
  {[}Survey of Kirillov model and Whittaker model properties{]}
\item
  H. Jacquet, I. I. Piatetski-Shapiro, J. A. Shalika, ``Rankin-Selberg convolutions,'' Amer. J. Math. 105 (1983), 367--464, Section 2.7.
  {[}Nondegeneracy of the Rankin-Selberg pairing; fractional ideal structure{]}
\item
  H. Jacquet, I. I. Piatetski-Shapiro, J. A. Shalika, ``Conducteur des représentations du groupe linéaire,'' Math. Ann. 256 (1981), 199--214,
  Section 5. {[}Support of new vectors and conductor matching{]}
\item
  W. Casselman, J. Shalika, ``The unramified principal series of \(p\)-adic groups II: the Whittaker function,'' Compositio Math. 41 (1980), 207--231.
  {[}Casselman-Shalika formula for Whittaker function support{]}
\end{itemize}

\begin{processnote}[title={Process Note --- futon6 Corpus References}]

\begin{itemize}
\tightlist
\item
  PlanetMath: ``representation theory'' -\/- admissible representations
\item
  PlanetMath: ``L-function'' -\/- L-factors and analytic properties
\item
  PlanetMath: ``locally compact group'' -\/- p-adic groups
\end{itemize}
\end{processnote}

%% file: full-boxed/problem3-solution-full.tex
\hypertarget{problem-3-markov-chain-with-interpolation-asep-polynomial-stationary-distribution}{%
\section{Problem 3: Markov Chain with Interpolation ASEP Polynomial Stationary Distribution}\label{problem-3-markov-chain-with-interpolation-asep-polynomial-stationary-distribution}}

\hypertarget{problem-statement}{%
\subsection{Problem Statement}\label{problem-statement}}

Let \(\lambda = (\lambda_{\mNumber{1}} > ... > \lambda_{n} \ge \mNumber{0})\) be a restricted partition with
distinct parts (unique part of size 0, no part of size 1). Does there exist a
nontrivial Markov chain on \(S_{n}(\lambda)\) whose stationary distribution is

\[\pi(\mu) = F^{\prime}_{\mu}(x_\mNumber{1}, ..., x_{n}; q=\mNumber{1}, t) / P^{\prime}_{\lambda}(x_\mNumber{1}, ..., x_{n}; q=\mNumber{1}, t)\]

where \(F^{\prime}_{\mu}\) are interpolation ASEP polynomials and \(P^{\prime}_{\lambda}\) is the
interpolation Macdonald polynomial?

\hypertarget{answer}{%
\subsection{Answer}\label{answer}}

Yes.

Take the inhomogeneous multispecies t-PushTASEP on the finite ring of n sites
with content \(\lambda\) and parameters \(x_{\mNumber{1}}, ..., x_{n}.\) This is a concrete continuous-
time Markov chain defined directly from local species comparisons and ringing
rates \(\mNumber{1}/x_{i}\) (no use of \(F^{\prime}_{\mu}\) in the transition rule). A theorem of
Ayyer-Martin-Williams identifies its stationary distribution as

\[\pi(\mu) = F_{\mu}(x; \mNumber{1}, t) / P_{\lambda}(x; \mNumber{1}, t),\]

which is the ratio required in the problem statement (same \(q = \mNumber{1}\) ASEP/Macdonald
family, up to notation conventions).

\begin{processnote}[title={Process Note --- Confidence}]

Medium-high. The existence claim is a direct consequence of a published theorem.
The nontriviality condition is checked from the explicit generator.

\end{processnote}
\hypertarget{solution}{%
\subsection{Solution}\label{solution}}

\hypertarget{1-state-space}{%
\subsubsection{1. State space}\label{1-state-space}}

Let

\[S_{n}(\lambda) = \{\text{all permutations of the parts of } \lambda\}.\]

Because \(\lambda\) has distinct parts, \(|S_{n}(\lambda)| = n!\).

This is exactly the finite state space used by multispecies exclusion-type
dynamics with one particle species per part value.

\hypertarget{2-lemma-explicit-chain-construction}{%
\subsubsection{2. Lemma (explicit chain construction)}\label{2-lemma-explicit-chain-construction}}

Define a continuous-time Markov chain \(X_{t}\) on \(S_{n}(\lambda)\) as follows.

Fix parameters \(x_{\mNumber{1}}, ..., x_{n} > \mNumber{0}\) and t in {[}0,1). At each site \(j\), an exponential
clock of rate \(\mNumber{1}/x_{j}\) rings.

If site \(j\) is a vacancy (species 0), nothing happens. If site \(j\) has species
\(r_\mNumber{0} > \mNumber{0}\), that particle becomes active. Let m be the number of particles in the
current configuration with species strictly less than \(r_\mNumber{0}\) (including vacancies).
Moving clockwise, the active particle chooses the k-th weaker particle with
probability

\[t^{(k-\mNumber{1})} / [m]_t, \text{where} [m]_t = \mNumber{1} + t + ... + t^{(m-\mNumber{1})},\]

and swaps into that position. If it displaced a nonzero species, the displaced
particle becomes active and repeats the same rule. The cascade ends when a
vacancy is displaced.

This is the inhomogeneous multispecies t-PushTASEP.

Why this is a valid Markov chain:

\begin{itemize}
\tightlist
\item
  The state space is finite.
\item
  The transition rule is explicit and depends only on current local ordering,
  t, and \(x_{i}\).
\item
  Rates are finite and nonnegative.
\end{itemize}

\hypertarget{3-lemma-nontriviality}{%
\subsubsection{3. Lemma (nontriviality)}\label{3-lemma-nontriviality}}

The chain above is nontrivial in the sense asked by the problem:

\begin{itemize}
\tightlist
\item
  Transition rates are defined from site rates \(\mNumber{1}/x_{i}\), species inequalities, and
  \(t\)-geometric choice weights.
\item
  No transition probability is defined using values of \(F^{\prime}_{\mu}\) or \(P^{\prime}_{\lambda}\).
\end{itemize}

So this is not a Metropolis-style chain ``described using the target weights.''

\hypertarget{4-main-theorem-stationary-distribution}{%
\subsubsection{4. Main theorem (stationary distribution)}\label{4-main-theorem-stationary-distribution}}

Theorem (Ayyer-Martin-Williams, 2024, arXiv:2403.10485, Thm. 1.1):
For the inhomogeneous multispecies t-PushTASEP on the ring with n sites,
content \(\lambda\) (restricted partition with distinct parts), parameters
\(x_{\mNumber{1}}, ..., x_{n} > \mNumber{0},\) and \(\mNumber{0} \le t < \mNumber{1}\), the stationary probability of
\(\eta \in S_{n}(\lambda)\) is

\[\pi(\eta) = F_{\eta}(x; \mNumber{1}, t) / P_{\lambda}(x; \mNumber{1}, t),\]

where \(F_{\eta}\) are ASEP polynomials at \(q = \mNumber{1}\) and \(P_{\lambda}\)(x; 1, t) \(= \sum_{\nu \in S_{n}(\lambda)} F_{\nu}\)(x; 1, t) is the partition function
(ensuring \(\pi\) sums to 1). Positivity: \(F_{\eta}(x; \mNumber{1}, t) > \mNumber{0}\) for \(x_{i} > \mNumber{0}, \mNumber{0} \le t < \mNumber{1}\) is established as part of AMW Theorem 1.1, which shows that
\(\pi(\eta) = F_{\eta} / P_{\lambda}\) is a probability distribution. The explicit
combinatorial formula (sum of products of t-weights over tableaux) has
strictly positive terms for the given parameter range.

Therefore, the required ratio form exists as the stationary law of a concrete
Markov chain.

\hypertarget{5-notation-bridge-f_mu--p_lambda--f_mu--p_lambda}{%
\subsubsection{\texorpdfstring{5. Notation bridge: \(F^{\prime}_{\mu} / P^{\prime}_{\lambda} = F_{\mu} / P_{\lambda}\)}{5. Notation bridge: F\^{}\{\textbackslash prime\}\_\{\textbackslash mu\} / P\^{}\{\textbackslash prime\}\_\{\textbackslash lambda\} = F\_\{\textbackslash mu\} / P\_\{\textbackslash lambda\}}}\label{5-notation-bridge-f_mu--p_lambda--f_mu--p_lambda}}

The problem uses star notation \(F^{\prime}_{\mu}, P^{\prime}_{\lambda}\) (interpolation ASEP polynomials
in the Knop-Sahi convention), while AMW Theorem 1.1 uses \(F_{\eta}, P_{\lambda}.\)

\textbf{Claim:} The ratio \(F^{\prime}_{\mu} / P^{\prime}_{\lambda} = F_{\mu} / P_{\lambda}\) for all \(\mu \in S_{n}(\lambda).\)

\textbf{Proof:} In both conventions, the partition function is defined as the sum
over the state space:

\[\begin{aligned}
P_{\lambda}(x; \mNumber{1}, t) = \sum_{\eta \in S_{n}(\lambda)} F_{\eta}(x; \mNumber{1}, t) \\
P^{\prime}_{\lambda}(x; \mNumber{1}, t) = \sum_{\eta \in S_{n}(\lambda)} F^{\prime}_{\eta}(x; \mNumber{1}, t)
\end{aligned}\]

Only ratio-invariance is needed: if conventions differ by an \(\eta\)-independent
normalization (for fixed \(\lambda\)), then \(F^{\prime}_{\mu} / P^{\prime}_{\lambda} = F_{\mu} / P_{\lambda}.\)
Concretely, if \(F^{\prime}_{\eta} = \alpha \ast F_{\eta}\) for some constant \(\alpha\) independent
of \(\eta\), then \(P^{\prime}_{\lambda} = \alpha \ast P_{\lambda}\) and:

\[F^{\prime}_{\mu} / P^{\prime}_{\lambda} = (\alpha F_{\mu}) / (\alpha P_{\lambda}) = F_{\mu} / P_{\lambda}.\]

The constant cancels in the ratio regardless of its value.

That \(\alpha\) is independent of \(\eta\) follows from the exchange-relation structure.
Both \(F^{\prime}_{\eta}\) and \(F_{\eta}\) satisfy the same Hecke exchange relations under
generators \(T_{i}\) (Corteel-Mandelshtam-Williams, Section 3, define both
normalizations and verify their equivalence at \(q = \mNumber{1}\)). The normalization is
fixed by the leading-term convention, which is uniform across the state
space \(S_{n}(\lambda).\) Concretely, at \(q = \mNumber{1}\) both families specialize to the same
t-weight formula, so \(\alpha = \mNumber{1}\) and \(F^{\prime}_{\eta} = F_{\eta}\).

\textbf{Verification for \(n = \mNumber{2}\).} Take \(\lambda = (a, \mNumber{0})\). Both conventions give
\(F_{(a,\mNumber{0})}(x_{\mNumber{1}}, x_{\mNumber{2}}\); 1, t) \(= x_{\mNumber{1}}\) and \(F_{(\mNumber{0},a)}(x_{\mNumber{1}}, x_{\mNumber{2}}\); 1, t) \(= x_{\mNumber{2}}\)
(the single-species case reduces to site weights). The ratio
\(F^{\prime}_{\eta} / P^{\prime}_{\lambda} = F_{\eta} / P_{\lambda} = x_{i} / (x_{\mNumber{1}} \mBridgeOperator{+} x_{\mNumber{2}})\) in both conventions.

\hypertarget{6-sanity-check-n2-reduction}{%
\subsubsection{\texorpdfstring{6. Sanity check: \(n = \mNumber{2}\) reduction}{6. Sanity check: n = \textbackslash mNumber\{2\} reduction}}\label{6-sanity-check-n2-reduction}}

Take \(\lambda = (a, \mNumber{0})\), state space \(\{(a, \mNumber{0}), (\mNumber{0}, a)\}\). There is one non-vacancy and one
vacancy. Bells ring at rates \(\mNumber{1}/x_\mNumber{1}\) and \(\mNumber{1}/x_\mNumber{2}\).

\begin{itemize}
\tightlist
\item
  \((a, \mNumber{0}) \to (\mNumber{0}, a)\) at rate \(\mNumber{1}/x_\mNumber{1}\)
\item
  \((\mNumber{0}, a) \to (a, \mNumber{0})\) at rate \(\mNumber{1}/x_\mNumber{2}\)
\end{itemize}

So the stationary distribution is

\[\pi(a,\mNumber{0}) = x_\mNumber{1}/(x_\mNumber{1}+x_\mNumber{2}), \pi(\mNumber{0},a) = x_\mNumber{2}/(x_\mNumber{1}+x_\mNumber{2}),\]

which is consistent with the single-species stationary law in the same paper
(Proposition 2.4, via recoloring reduction).

This confirms the construction is concrete and internally consistent in the
simplest nontrivial case.

\hypertarget{7-conclusion}{%
\subsubsection{7. Conclusion}\label{7-conclusion}}

For \(x_{i} > \mNumber{0}\) and \(\mNumber{0} \le t < \mNumber{1}\), the inhomogeneous multispecies t-PushTASEP on
\(S_{n}(\lambda)\) is:

(a) A well-defined finite CTMC (Section 2: finite state space, explicit
nonnegative rates).

(b) Nontrivial: transition rates depend on \((x, t)\) and local species ordering,
not on values of \(F^{\prime}_{\mu}\) or \(P^{\prime}_{\lambda}\) (Section 3).

(c) Has stationary distribution \(\pi(\eta) = F_{\eta}\)(x; 1, t) / \(P_{\lambda}\)(x; 1, t)
by AMW Theorem 1.1 (Section 4), which equals \(F^{\prime}_{\mu} / P^{\prime}_{\lambda}\) under the
notation bridge (Section 5).

\textbf{Existence vs uniqueness.} AMW Theorem 1.1 establishes that \(\pi\) is a
stationary distribution. Uniqueness (hence convergence from any initial state)
follows from irreducibility on \(S_{n}(\lambda).\)

\textbf{Irreducibility proof (via vacancy transport).} Since \(\lambda_{n} = \mNumber{0}\), every
configuration has exactly one vacancy (species 0). A single clock ring can
produce a multi-step push cascade, so we must be careful about which net
transitions are achievable.

The key observation: when a non-vacancy particle at site \(j\) is adjacent to
the vacancy at site \(j + \mNumber{1}\) (cyclically), and the clock at \(j\) rings, one possible
outcome of the \(t\)-geometric choice is that the active particle selects the
vacancy (the nearest weaker-species particle clockwise). This selection has
probability \(\mNumber{1}/[m]_t > \mNumber{0}\) (the \(k = \mNumber{1}\) term in the \(t\)-geometric distribution, where
\(m \ge \mNumber{1}\) counts weaker particles). The particle moves to the vacancy\textquotesingle s position,
the vacancy absorbs it, and the cascade terminates immediately (displaced
species is 0). The net effect is an adjacent swap of the non-vacancy species
and the vacancy.

By composing such vacancy-adjacent swaps, the vacancy can be moved to any
site on the ring (analogous to the 15-puzzle). Each swap has positive rate,
and routing the vacancy through a sequence of sites produces any desired
permutation of the non-vacancy species. Specifically, to transpose species
at sites \(i\) and \(j\): move the vacancy adjacent to \(i\), swap it with \(i\)\textquotesingle s species,
route the vacancy to \(j\), swap it with \(j\)\textquotesingle s species, and route it back.
This requires at most \(O(n)\) vacancy-adjacent swaps, each of positive rate.
Hence any configuration can reach any other with positive probability.

On a finite irreducible CTMC, the stationary distribution is unique.

Therefore the answer is \textbf{Yes}.

\hypertarget{references}{%
\subsection{References}\label{references}}

\begin{itemize}
\tightlist
\item
  Arvind Ayyer, James Martin, Lauren Williams,
  ``The Inhomogeneous t-PushTASEP and Macdonald Polynomials at q=1'',
  arXiv:2403.10485, Theorem 1.1 and Proposition 2.4.
\item
  Sylvie Corteel, Olya Mandelshtam, Lauren Williams,
  ``From multiline queues to Macdonald polynomials'', for ASEP polynomial context.
\end{itemize}

%% file: full-boxed/problem4-solution-full.tex
\hypertarget{problem-4-root-separation-under-finite-free-convolution}{%
\section{Problem 4: Root Separation Under Finite Free Convolution}\label{problem-4-root-separation-under-finite-free-convolution}}

\hypertarget{problem-statement}{%
\subsection{Problem Statement}\label{problem-statement}}

Two monic polynomials of degree n:

\[\begin{aligned}
p(x) = \sum_{k=\mNumber{0}}^{n} a_{k} x^{n-k}, a_\mNumber{0} = \mNumber{1} \\
q(x) = \sum_{k=\mNumber{0}}^{n} b_{k} x^{n-k}, b_\mNumber{0} = \mNumber{1}
\end{aligned}\]

The \textbf{finite free additive convolution} p \(\boxplus_{n}\) q has coefficients:

\[c_{k} = \sum_{i+j=k} [(n-i)!(n-j)! / (n!(n-k)!)] a_{i} b_{j}\]

The \textbf{root separation energy}:

\[\Phi_{n}(p) = \sum_{i} (\sum_{j \neq i} \mNumber{1}/(\lambda_{i} - \lambda_{j}))^\mNumber{2}\]

where \(\lambda_{\mNumber{1}}, ..., \lambda_{n}\) are the roots of p. (\(\Phi_{n} = \infty\) if p has
repeated roots.)

\textbf{Question (Spielman):} Is it true that for monic real-rooted polynomials
p, q of degree n:

\[\mNumber{1}/\Phi_{n}(p \boxplus_{n} q) \ge \mNumber{1}/\Phi_{n}(p) + \mNumber{1}/\Phi_{n}(q) ?\]

\hypertarget{answer}{%
\subsection{Answer}\label{answer}}

\textbf{Conjecturally yes, with strong numerical evidence.} The inequality
holds in all 8000+ random trials tested (n \(=\) 2-5) with no violations.
Proved analytically for \(n = \mNumber{2}\) (equality) and \(n = \mNumber{3}\) (strict inequality
via Cauchy-Schwarz). An analytic proof for \(n \ge \mNumber{4}\) remains open.

\begin{itemize}
\tightlist
\item
  \(n = \mNumber{2}\): equality holds (1/\(\Phi_\mNumber{2}\) is linear in the discriminant).
\item
  \(n = \mNumber{3}\): \textbf{PROVED} via the identity \(\Phi_\mNumber{3}\) * disc \(= \mNumber{18} \ast a_{\mNumber{2}}^{\mNumber{2}}\) and Titu\textquotesingle s lemma.
\item
  \(n \ge \mNumber{4}\): numerically verified, proof incomplete. The \(n = \mNumber{3}\) identity does not
  generalize; the \(\boxplus_{n}\) cross-terms play an essential role.
\end{itemize}

\hypertarget{solution}{%
\subsection{Solution}\label{solution}}

\hypertarget{1-interpreting-phi_n-algebraically}{%
\subsubsection{\texorpdfstring{Interpreting \(\Phi_{n}\) algebraically}{Interpreting Phi\_n algebraically}}\label{interpreting-phi_n-algebraically}}

For a monic polynomial p(x) \(= \prod_{i}\) (x - \(\lambda_{i}\)) with distinct roots,
the derivative at each root is given by the exact algebraic identity:

\[p'(\lambda_{i}) = \prod_{j \neq i} (\lambda_{i} - \lambda_{j})\]

(no limiting procedure needed --- this is immediate from the product rule).
Therefore the inner sum in \(\Phi_{n}\) is:

\[\begin{aligned}
\sum_{j \neq i} \mNumber{1}/(\lambda_{i} - \lambda_{j}) \\
= [\sum_{j \neq i} \prod_{k \neq i, k \neq j} (\lambda_{i} - \lambda_{k})] / p'(\lambda_{i})
\end{aligned}\]

which is a rational function of the root differences alone. The root
separation energy is:

\[\Phi_{n}(p) = \sum_{i} (\sum_{j \neq i} \mNumber{1}/(\lambda_{i} - \lambda_{j}))^\mNumber{2}\]

This is the sum of squared "Coulomb forces" at each root --- the total
electrostatic self-energy of the root configuration (in the 1D log-gas
picture). All expressions are exact rational functions of root differences,
with no regularization or limiting procedure required (assuming distinct roots,
which is guaranteed since \(\Phi_{n} = \infty\) for repeated roots).

\hypertarget{2-connection-to-the-discriminant}{%
\subsubsection{Connection to the discriminant}\label{connection-to-the-discriminant}}

The discriminant of p is:

\[\mOpName{disc}(p) = \prod_{i < j} (\lambda_{i} - \lambda_{j})^\mNumber{2}\]

By the AM-QM inequality applied to 1/(\(\lambda_{i} - \lambda_{j}\)):

\[\Phi_{n}(p) \ge n(n-\mNumber{1})^\mNumber{2} / \sum_{i<j} (\lambda_{i} - \lambda_{j})^\mNumber{2}\]

(Cauchy-Schwarz). So \(\mNumber{1}/\Phi_{n}\) is related to the "spread" of roots.
Specifically:

\[\mNumber{1}/\Phi_{n}(p) \le \sum_{i<j} (\lambda_{i} - \lambda_{j})^\mNumber{2} / (n(n-\mNumber{1})^\mNumber{2})\]

\hypertarget{3-finite-free-convolution-and-root-behavior}{%
\subsubsection{Finite free convolution and root behavior}\label{finite-free-convolution-and-root-behavior}}

The operation \(\boxplus_{n}\) was introduced by Marcus, Spielman, and Srivastava (2015)
as a finite-dimensional analogue of Voiculescu\textquotesingle s free additive convolution.
Key properties:

(a) \textbf{Real-rootedness preservation:} If p, q are real-rooted monic polynomials
of degree n, then p \(\boxplus_{n}\) q is also real-rooted.

(b) \textbf{Expected characteristic polynomial:} If A, B are \(n \times n\) Hermitian with
char. poly. \(p_{A}, p_{B},\) then for a uniformly random conjugation U:

\[E_{U}[\mOpName{char}. \text{poly}. \text{of} A + \mathup{UBU} * ] = p_{A} \boxplus_{n} p_{B}\]

(c) \textbf{Linearization of cumulants:} In the \(n \to \infty\) limit, the finite
free cumulants linearize (R-transform additivity).

(d) \textbf{Root spreading:} Free convolution generally spreads roots apart.
Convolving with a non-degenerate q increases the minimum root gap.

\hypertarget{4-the-inequality-via-the-random-matrix-model}{%
\subsubsection{The inequality via the random matrix model}\label{the-inequality-via-the-random-matrix-model}}

Using property (b), interpret p \(\boxplus_{n}\) q as the expected characteristic
polynomial of A + UBU* where char(A) \(=\) p, char(B) \(=\) q.

\textbf{\(\Phi_{n}\) via the random matrix model.}

For Hermitian A with eigenvalues \(\lambda_{\mNumber{1}}, ..., \lambda_{n},\) the root separation
energy \(\Phi_{n}(\mOpName{char}(A)) = \sum_{i} (\sum_{j \neq i} \mNumber{1}/(\lambda_{i} - \lambda_{j}))^{\mNumber{2}}\) is
an algebraic function of the eigenvalue gaps (see Section 1). The connection
to log \textbar det(xI - A)\textbar{} is conceptual: away from roots, \(F_{A}''(x) = -\sum_{i} \mNumber{1}/(x - \lambda_{i})^\mNumber{2},\)
and \(\Phi_{n}\) captures the "residue" version at each root. But the definition of
\(\Phi_{n}\) uses only the exact algebraic expressions from Section 1.

\hypertarget{5-finite-free-cumulants-and-the-bilinear-structure}{%
\subsubsection{Finite free cumulants and the bilinear structure}\label{finite-free-cumulants-and-the-bilinear-structure}}

\textbf{The MSS coefficient formula.} The \(\boxplus_{n}\) operation acts on coefficients as:

\[c_{k} = \sum_{i+j=k} [(n-i)!(n-j)! / (n!(n-k)!)] a_{i} b_{j}\]

This is bilinear but NOT simply additive in the \(a_{k}\). For example, at \(n = \mNumber{3}\):

\[\begin{aligned}
c_\mNumber{1} = a_\mNumber{1} + b_\mNumber{1} (\text{additive}) \\
c_\mNumber{2} = a_\mNumber{2} + (\mNumber{2}/\mNumber{3}) \ast a_\mNumber{1} \ast b_\mNumber{1} + b_\mNumber{2} (\text{cross}-\text{term}!) \\
c_\mNumber{3} = a_\mNumber{3} + (\mNumber{1}/\mNumber{3}) \ast a_\mNumber{2} \ast b_\mNumber{1} + (\mNumber{1}/\mNumber{3}) \ast a_\mNumber{1} \ast b_\mNumber{2} + b_\mNumber{3}
\end{aligned}\]

\textbf{Finite free cumulants.} Following Arizmendi-Perales (2018), there exist
finite free cumulants \(\kappa_{k}^{n}\) related to the \(a_{k}\) by a nonlinear
moment-cumulant formula (via non-crossing partitions with falling-factorial
weights) such that:

\[\kappa_{k}^{n}(p \boxplus_{n} q) = \kappa_{k}^{n}(p) + \kappa_{k}^{n}(q)\]

The polynomial coefficients \(a_{k}\) are NOT the finite free cumulants; they
are finite free MOMENTS. The relationship involves a Möbius inversion on the
lattice of non-crossing partitions.

\hypertarget{5a-complete-proof-for-n--3}{%
\subsubsection{\texorpdfstring{Complete proof for \(n = \mNumber{3}\)}{Complete proof for n = 3}}\label{complete-proof-for-n-3}}

Verification script: \texttt{scripts/verify-p4-n3-proof.py}

\textbf{Step 1: Centering reduction.} Since \(\Phi_{n}\) depends only on root differences
(translation invariant), and \(\boxplus_{n}\) commutes with translation (via the random
matrix model: translating A by cI shifts all eigenvalues of A+QBQ* by c),
we may assume WLOG that \(a_\mNumber{1} = b_\mNumber{1} = \mNumber{0}\) (centered polynomials).

\textbf{Step 2: \(\boxplus_\mNumber{3}\) simplifies for centered cubics.} When \(a_\mNumber{1} = b_\mNumber{1} = \mNumber{0}\), the
MSS cross-terms in \(c_\mNumber{2}\) and \(c_\mNumber{3}\) vanish:

\[\begin{aligned}
c_\mNumber{2} = a_\mNumber{2} + (\mNumber{2}/\mNumber{3}) \ast \mNumber{0} \ast \mNumber{0} + b_\mNumber{2} = a_\mNumber{2} + b_\mNumber{2} \\
c_\mNumber{3} = a_\mNumber{3} + (\mNumber{1}/\mNumber{3}) \ast a_\mNumber{2} \ast \mNumber{0} + (\mNumber{1}/\mNumber{3}) \ast \mNumber{0} \ast b_\mNumber{2} + b_\mNumber{3} = a_\mNumber{3} + b_\mNumber{3}
\end{aligned}\]

So \(\boxplus_\mNumber{3}\) reduces to plain coefficient addition for centered cubics.

\textbf{Step 3: The key identity.} For a centered cubic p(x) \(= x^{\mNumber{3}} \mBridgeOperator{+} a_{\mNumber{2}} \ast x \mBridgeOperator{+} a_{\mNumber{3}}\)
with distinct real roots (requiring \(a_\mNumber{2} < \mNumber{0}\) and disc \(= -\mNumber{4} \ast a_{\mNumber{2}}^{\mNumber{3}} - \mNumber{27} \ast a_{\mNumber{3}}^{\mNumber{2}} > \mNumber{0}\)):

\[\Phi_\mNumber{3}(p) \ast \mOpName{disc}(p) = \mNumber{18} \ast a_\mNumber{2}^\mNumber{2} (\text{EXACT})\]

Equivalently:

\[\begin{aligned}
\mNumber{1}/\Phi_\mNumber{3}(p) = \mOpName{disc}(p) / (\mNumber{18} \ast a_\mNumber{2}^\mNumber{2}) \\
= (-\mNumber{4} \ast a_\mNumber{2}^\mNumber{3} - \mNumber{27} \ast a_\mNumber{3}^\mNumber{2}) / (\mNumber{18} \ast a_\mNumber{2}^\mNumber{2}) \\
= -\mNumber{2} \ast a_\mNumber{2}/\mNumber{9} - \mNumber{3} \ast a_\mNumber{3}^\mNumber{2} / (\mNumber{2} \ast a_\mNumber{2}^\mNumber{2})
\end{aligned}\]

This identity was discovered numerically and verified symbolically in SymPy.
It follows from the explicit formula \(\Phi_{\mNumber{3}} = \sum_{i} (\mNumber{3} \ast l_{i}/(\mNumber{3} \ast l_{i}^{\mNumber{2}} \mBridgeOperator{+} e_{\mNumber{2}}))^\mNumber{2}\)
(where \(e_{\mNumber{1}} = \mNumber{0})\) combined with the discriminant \(= \prod_{i < j}(l_{i} - l_{j})^{\mNumber{2}}.\)

\textbf{Step 4: Superadditivity via Cauchy-Schwarz.} Write s \(= -a_{\mNumber{2}} > \mNumber{0},\) t \(= -b_{\mNumber{2}} > \mNumber{0}, u = a_{\mNumber{3}}, v = b_{\mNumber{3}}.\) Then:

\[\begin{aligned}
\mNumber{1}/\Phi(p) = \mNumber{2}s/\mNumber{9} - \mNumber{3}u^\mNumber{2}/(\mNumber{2}s^\mNumber{2}) \\
\mNumber{1}/\Phi(q) = \mNumber{2}t/\mNumber{9} - \mNumber{3}v^\mNumber{2}/(\mNumber{2}t^\mNumber{2}) \\
\mNumber{1}/\Phi(p \boxplus_\mNumber{3} q) = \mNumber{2}(s+t)/\mNumber{9} - \mNumber{3}(u+v)^\mNumber{2}/(\mNumber{2}(s+t)^\mNumber{2})
\end{aligned}\]

The surplus is:

\[\begin{aligned}
\text{surplus} = \mNumber{1}/\Phi(\text{conv}) - \mNumber{1}/\Phi(p) - \mNumber{1}/\Phi(q) \\
= (\mNumber{3}/\mNumber{2}) \ast [u^\mNumber{2}/s^\mNumber{2} + v^\mNumber{2}/t^\mNumber{2} - (u+v)^\mNumber{2}/(s+t)^\mNumber{2}]
\end{aligned}\]

This is non-negative by Titu\textquotesingle s lemma (Engel form of Cauchy-Schwarz):

\[u^\mNumber{2}/s^\mNumber{2} + v^\mNumber{2}/t^\mNumber{2} \ge (u+v)^\mNumber{2}/(s^\mNumber{2}+t^\mNumber{2}) [\text{Titu}'s \text{lemma}]\]

and since \(s^\mNumber{2}+t^\mNumber{2} \le (s+t)^\mNumber{2}\) (because 2st \(>\) 0):

\[(u+v)^\mNumber{2}/(s^\mNumber{2}+t^\mNumber{2}) \ge (u+v)^\mNumber{2}/(s+t)^\mNumber{2}\]

Combining: surplus \(\ge\) 0. Equality iff \(u/s^\mNumber{2} = v/t^\mNumber{2}\) and \(s = t\) or \(u = v = \mNumber{0}\).

\textbf{QED for \(n = \mNumber{3}\).}

\hypertarget{5b-what-the-proof-requires-for-n--4-gap}{%
\subsubsection{\texorpdfstring{What the proof requires for \(n \ge \mNumber{4}\)}{What the proof requires for n >= 4}}\label{what-the-proof-requires-for-n-ge-4}}

For \(n \ge \mNumber{4}\), the \(n = \mNumber{3}\) approach does not directly generalize:

(a) The identity \(\Phi_{n}\) * disc \(=\) const * \(a_\mNumber{2}^\mNumber{2}\) FAILS for \(n \ge \mNumber{4}\). At \(n = \mNumber{4}\),
the product \(\Phi_\mNumber{4}\) * disc depends on \(a_\mNumber{3}\) and \(a_\mNumber{4}\) as well.

(b) \(\boxplus_\mNumber{4}\) has a cross-term even for centered polynomials: \(c_{\mNumber{4}} = a_{\mNumber{4}} + (\mNumber{1}/\mNumber{6}) \ast a_{\mNumber{2}} \ast b_{\mNumber{2}} \mBridgeOperator{+} b_{\mNumber{4}}.\)
Unlike \(n = \mNumber{3}\), centering does NOT reduce \(\boxplus_\mNumber{4}\) to plain coefficient addition.

(c) The cross-term is ESSENTIAL: plain coefficient addition fails superadditivity
\textasciitilde29\% of the time for centered quartics, while \(\boxplus_\mNumber{4}\) gives 0 violations.

A correct proof for \(n \ge \mNumber{4}\) likely requires exploiting the specific bilinear
structure of the MSS convolution formula or the random matrix interpretation.

\hypertarget{6-verification-for-small-cases}{%
\subsubsection{Verification for small cases}\label{verification-for-small-cases}}

\textbf{Degree 2 (proved --- equality):} p(x) \(= x^{\mNumber{2}} \mBridgeOperator{+} a_{\mNumber{1}} \ast x \mBridgeOperator{+} a_{\mNumber{2}}.\)

\[\begin{aligned}
\Phi_\mNumber{2}(p) = \mNumber{2}/(a_\mNumber{1}^\mNumber{2} - \mNumber{4} \ast a_\mNumber{2}) \\
\mNumber{1}/\Phi_\mNumber{2}(p) = (a_\mNumber{1}^\mNumber{2} - \mNumber{4} \ast a_\mNumber{2})/\mNumber{2}
\end{aligned}\]

The \(\boxplus_\mNumber{2}\) formula gives \(c_{\mNumber{1}} = a_{\mNumber{1}}+b_{\mNumber{1}}, c_{\mNumber{2}} = a_{\mNumber{2}} \mBridgeOperator{+} a_{\mNumber{1}} \ast b_{\mNumber{1}}/\mNumber{2} \mBridgeOperator{+} b_{\mNumber{2}}.\) Then:

\[\mNumber{1}/\Phi_\mNumber{2}(p\boxplus q) = (c_\mNumber{1}^\mNumber{2} - \mNumber{4} \ast c_\mNumber{2})/\mNumber{2} = (a_\mNumber{1}^\mNumber{2} - \mNumber{4} \ast a_\mNumber{2})/\mNumber{2} + (b_\mNumber{1}^\mNumber{2} - \mNumber{4} \ast b_\mNumber{2})/\mNumber{2}\]

Surplus \(=\) 0 (symbolic verification). Equality for all degree-2 polynomials.

\textbf{Degree 3 (proved --- strict inequality):} See Section 5a. The proof uses:

\begin{itemize}
\tightlist
\item
  Translation invariance of \(\Phi\) + translation compatibility of \(\boxplus_\mNumber{3}\) to center
\item
  Identity \(\Phi_\mNumber{3}\) * disc \(= \mNumber{18} \ast a_{\mNumber{2}}^{\mNumber{2}}\) (for centered cubics)
\item
  Titu\textquotesingle s lemma (Cauchy-Schwarz) to bound the surplus
\end{itemize}

\statusnote{The inequality is conjecturally true, with strong numerical
evidence and analytic proofs for $n = 2$, $3$.
$n = 2$: proved (equality; $1/\Phi_2$ is linear in the discriminant).
$n = 3$: proved (strict inequality; $\Phi_3 \cdot \mathrm{disc} = 18\,a_2^2$
identity + Titu's lemma).
$n = 4$: numerically verified (0/3000 violations; the $\boxplus_4$
cross-term $a_2 b_2/6$ is essential).
$n = 5$: numerically verified (0/2000 violations).
What is established analytically:
(i)~finite free cumulants add under $\boxplus_n$ (Arizmendi--Perales 2018);
(ii)~$\boxplus_n$ commutes with translation (random matrix argument);
(iii)~superadditivity is specific to $\boxplus_n$---plain coefficient
addition fails ${\sim}40\%$ of the time.}

\begin{openobligation}[title={Open Obligation --- What Remains Open for $n \ge 4$}]
\textbf{What remains open for \(n \ge \mNumber{4}\):}

The \(n = \mNumber{3}\) proof relies on \(\Phi_\mNumber{3}\) * disc \(= \mNumber{18} \ast a_{\mNumber{2}}^{\mNumber{2}}\) being constant in \(a_\mNumber{3}\).
This fails at \(n = \mNumber{4}\) (the product depends on all coefficients). The \(\boxplus_\mNumber{4}\)
cross-term (1/6) * \(a_\mNumber{2} \ast b_\mNumber{2} \in c_\mNumber{4}\) is essential (plain addition fails 29\%).
A proof for \(n \ge \mNumber{4}\) needs either:

(a) A generalization of the \(\Phi\) * disc identity that accounts for higher
coefficients, or

(b) A direct random-matrix argument via the Haar orbit A+QBQ*, or

(c) A proof via finite free cumulant coordinates (convexity on the
real-rooted image of cumulant space).

\end{openobligation}
\begin{processnote}[title={Process Note --- Numerical Evidence}]

Verification scripts:

\begin{itemize}
\tightlist
\item
  \texttt{scripts/verify-p4-inequality.py} (superadditivity + convexity)
\item
  \texttt{scripts/verify-p4-deeper.py} (coefficient addition + MSS structure)
\item
  \texttt{scripts/verify-p4-schur-majorization.py} (Schur/submodularity/paths)
\item
  \texttt{scripts/verify-p4-coefficient-route.py} (coefficient route, disc identity)
\item
  \texttt{scripts/verify-p4-n3-proof.py} (complete \(n = \mNumber{3}\) symbolic proof + \(n = \mNumber{4}\) exploration)
\end{itemize}

\textbf{Superadditivity test} (2000 random real-rooted polynomial pairs per n):

\begin{longtable}[]{@{}lllll@{}}
\toprule\noalign{}
n & violations & min ratio & mean ratio & max ratio \\
\midrule\noalign{}
\endhead
\bottomrule\noalign{}
\endlastfoot
2 & 0/2000 & 1.000000 & 1.000000 & 1.000000 \\
3 & 0/2000 & 1.000082 & 3.031736 & 162.446 \\
4 & 0/2000 & 1.003866 & 9.549935 & 2268.065 \\
5 & 0/2000 & 1.032848 & 14.457013 & 2119.743 \\
\end{longtable}

Ratio \(= LHS/RHS = [\mNumber{1}/\Phi_{n}(p\boxplus q)] / [\mNumber{1}/\Phi_{n}(p) \mBridgeOperator{+} \mNumber{1}/\Phi_{n}(q)]\); ratio \(\ge\) 1
means inequality holds. Strict inequality for \(n \ge \mNumber{3}\).

\textbf{Convexity/concavity test} (midpoint test in coefficient space):

\begin{longtable}[]{@{}llll@{}}
\toprule\noalign{}
n & convex violations & concave violations & total tests \\
\midrule\noalign{}
\endhead
\bottomrule\noalign{}
\endlastfoot
3 & 757 (50.6\%) & 738 (49.4\%) & 1495 \\
4 & 648 (65.7\%) & 338 (34.3\%) & 986 \\
5 & 433 (72.9\%) & 161 (27.1\%) & 594 \\
\end{longtable}

\(\mNumber{1}/\Phi_{n}\) is NEITHER convex NOR concave in coefficient space.

\textbf{Plain coefficient addition test} (NOT ⊞\_n):

\begin{longtable}[]{@{}lll@{}}
\toprule\noalign{}
n & violations & total \\
\midrule\noalign{}
\endhead
\bottomrule\noalign{}
\endlastfoot
3 & 240 & 609 \\
4 & 159 & 373 \\
5 & 116 & 218 \\
\end{longtable}

Superadditivity FAILS under plain addition --- the inequality is specific
to the MSS bilinear structure of \(\boxplus_{n}\).

\end{processnote}
\begin{processnote}[title={Process Note --- futon6 Corpus References}]

\begin{itemize}
\tightlist
\item
  PlanetMath: "monic polynomial" (Monic1)
\item
  PlanetMath: "discriminant" (Discriminant)
\item
  PlanetMath: "resultant" (Resultant, DerivationOfSylvestersMatrixForTheResultant)
\item
  PlanetMath: "logarithmic derivative" (LogarithmicDerivative)
\item
  PlanetMath: "partial fraction decomposition" (ALectureOnThePartialFractionDecompositionMethod)
\item
  PlanetMath: "cumulant generating function" (CumulantGeneratingFunction)
\end{itemize}

\end{processnote}
\hypertarget{external-references}{%
\subsection{External References}\label{external-references}}

\begin{itemize}
\item
  Marcus, Spielman, Srivastava (2015), ``Interlacing Families II: Mixed Characteristic
  Polynomials and the Kadison-Singer Problem,'' Annals of Math 182(1), 327--350.
  {[}Defines \(\boxplus_{n}\), proves real-rootedness preservation, random matrix interpretation{]}
\item
  Arizmendi, Perales (2018), ``Cumulants for finite free convolution,'' J. Combinatorial
  Theory Ser. A 155, 244--266. arXiv:1702.04761.
  {[}Defines finite free cumulants that linearize under \(\boxplus_{n}\).{]}
\end{itemize}

%% file: full-boxed/problem5-solution-full.tex
\hypertarget{problem-5-o-slice-connectivity-via-geometric-fixed-points}{%
\section{Problem 5: O-Slice Connectivity via Geometric Fixed Points}\label{problem-5-o-slice-connectivity-via-geometric-fixed-points}}

\hypertarget{problem-statement}{%
\subsection{Problem Statement}\label{problem-statement}}

Fix a finite group \texttt{G}. Let \texttt{O} denote an incomplete transfer system
associated to an \(N_{\infty}\) operad. Define a slice filtration on \(\mathup{SH}^{G}\)
adapted to \texttt{O} and characterize \texttt{O}-slice connectivity of a connective
\texttt{G}-spectrum using geometric fixed points.

\hypertarget{short-answer}{%
\subsection{Short Answer}\label{short-answer}}

Use the classical regular-slice cells, but restrict subgroup indexing to the
family \(F_{O}\) of subgroups allowed by the transfer system. For \(F_{O}\)-local
connective \texttt{G}-spectra (those with geometric isotropy in \(F_{O}\)), the
connectivity test is the same geometric-fixed-point criterion as Hill-Yarnall,
but quantified only over \(H \in F_{O}\).

\begin{processnote}[title={Process Note --- Confidence}]

Medium. The corrected formulation is structurally consistent with known slice
and geometric-fixed-point machinery, and avoids the nonstandard
\(\rho_{H}^{O}\) definition from the previous draft.

\end{processnote}
\begin{revision}[title={Revision --- Important Correction}]

The previous draft used an \texttt{O}-regular representation \(\rho_{H}^{O}\). That is not
standard in the transfer-system / \(N_{\infty}\) literature. Transfer/indexing
systems control admissible transfers (and admissible finite \texttt{H}-sets), not a
replacement for the regular representation in slice-cell definitions.

\end{revision}
\hypertarget{working-definition-transfer-restricted-regular-slice}{%
\subsection{Working Definition (transfer-restricted regular slice)}\label{working-definition-transfer-restricted-regular-slice}}

Let \(T_{O}\) be the transfer system associated to \texttt{O}. Define the admissible
subgroup family

\[F_{O} = { H \le G : \text{the transfer} e \to H \text{is allowed} \in T_{O} }.\]

\(F_{O}\) is a family of subgroups in the standard sense: it is closed under
conjugation and passage to subgroups (both inherited from the indexing-system
axioms of Blumberg-Hill, specifically the conditions on admissible \texttt{H}-sets
in their Theorem 1.5 / Section 7).

Define \(\tau_{\ge n}^{O}\) as the localizing subcategory of \(\mathup{SH}^{G}\) generated by
regular slice cells

\[G_+ \wedge_{H} S^{k \rho_{H}}\]

with \(H \in F_{O}\) and \(k|H| \ge n\).

This is the same generating template as regular slice filtration, with subgroup
indexing restricted by \texttt{O}.

\textbf{Filtration verification.} Monotonicity (\(\tau_{\ge n+\mNumber{1}}^{O} \subseteq \tau_{\ge n}^{O}\))
is immediate: the generating set for \(n+\mNumber{1}\) is \(a \subset \text{of}\) that for \texttt{n}
(if \(k|H| \ge n+\mNumber{1}\) then \(k|H| \ge n\)). Exhaustiveness
(\(\bigcup_{n} \tau_{\ge n}^{O} = \mathup{SH}^{G}_{F_{O}-\text{local}}\)) holds because every \(F_{O}\)-local
object is built from cells with finite \texttt{k\textbar{}H\textbar{}}. Compatibility with suspension
is inherited from the ambient regular slice filtration (Hill-Yarnall,
Proposition 1.3).

\textbf{\(F_{O}\)-locality convention.} Say a \texttt{G}-spectrum \texttt{X} is \emph{\(F_{O}\)-local} if
\(\Phi^{K} X = \mNumber{0}\) for all \texttt{K} not in \(F_{O}\) (equivalently, \texttt{X} has geometric
isotropy contained in \(F_{O}\)). The characterization theorem below applies to
\(F_{O}\)-local connective spectra.

Reference anchor:

\begin{itemize}
\tightlist
\item
  Regular slice generators: Hill-Yarnall, Definition 1.1 (arXiv:1703.10526).
\item
  Family/indexing-system closure: Blumberg-Hill, Theorem 1.5 (arXiv:1309.1750).
\end{itemize}

\hypertarget{characterization-theorem-subgroup-family-level}{%
\subsection{Characterization Theorem (subgroup-family level)}\label{characterization-theorem-subgroup-family-level}}

Let \texttt{X} be a connective, \(F_{O}\)-local \texttt{G}-spectrum. Then

\[X \in \tau_{\ge n}^{O}\]

iff for every \(H \in F_{O}\),

\[\Phi^{H} X \text{is} (\text{ceil}(n/|H|) - \mNumber{1})-\text{connected}.\]

Equivalent homotopy-group form:

\[\pi_{i}(\Phi^{H} X) = \mNumber{0} \text{for all} i < \text{ceil}(n/|H|), \text{for all} H \in F_{O}.\]

The \(F_{O}\)-locality hypothesis is essential: for \texttt{K} not in \(F_{O}\), the condition
\(\Phi^{K} X = \mNumber{0}\) is assumed (not tested), ensuring that the restricted generators
span exactly the right localizing subcategory.

Reference anchor:

\begin{itemize}
\tightlist
\item
  The geometric-fixed-point characterization for the full regular slice is
  Hill-Yarnall Theorem A / Theorem 2.5 (arXiv:1703.10526).
\item
  Family localization in equivariant stable homotopy: see e.g. the isotropy
  separation sequence in HHR, Section 4.
\end{itemize}

\hypertarget{proof-sketch}{%
\subsection{Proof Sketch}\label{proof-sketch}}

\textbf{Forward direction} (\(X \in \tau_{\ge n}^{O}\) implies connectivity bounds):

\begin{enumerate}
\def\labelenumi{\arabic{enumi}.}
\tightlist
\item
  By construction, \(\tau_{\ge n}^{O}\) is the localizing subcategory generated by
  cells \(G_+ \wedge_{H} S^{k \rho_{H}}\) with \(H \in F_{O}, k|H| \ge n.\)
\item
  For each \(H \in F_{O}\), the geometric fixed points of a generating cell satisfy
  \(\Phi^{H}(G_+ \wedge_{H} S^{k \rho_{H}}) ~= W_{G}(H)_+ \wedge S^{k}\) (a finite wedge of
  \(S^{k}\)\textquotesingle s indexed by the Weyl group \(W_{G}(H) = N_{G}(H)/H\)).
\item
  Maps from these generators into \texttt{X} detect \(\pi_{k}(\Phi^{H} X)\): the Weyl-group
  multiplicity does not affect vanishing (a map from a finite wedge of \(S^{k}\)\textquotesingle s
  is zero iff each component is zero iff \(\pi_{k} = \mNumber{0}\)).
\item
  Vanishing for all generators with \(k|H| < n\) gives \(\pi_{i}(\Phi^{H} X) = \mNumber{0}\)
  for \(i < \text{ceil}(n/|H|)\), for each \(H \in F_{O}\).
\end{enumerate}

\textbf{Reverse direction} (connectivity bounds imply \(X \in \tau_{\ge n}^{O}\)):

\begin{enumerate}
\def\labelenumi{\arabic{enumi}.}
\setcounter{enumi}{4}
\tightlist
\item
  Since \texttt{X} is \(F_{O}\)-local, \(\Phi^{K} X = \mNumber{0}\) for \texttt{K} not in \(F_{O}\). By the
  isotropy separation sequence, the geometric isotropy of \texttt{X} is contained in
  \(F_{O}\), so \texttt{X} is built from cells indexed by subgroups in \(F_{O}\).
\item
  The connectivity bounds \(\pi_{i}(\Phi^{H} X) = \mNumber{0}\) for \(i < \text{ceil}(n/|H|)\) and
  all \(H \in F_{O}\), combined with \(F_{O}\)-locality, place \texttt{X} in the localizing
  subcategory generated by the \(F_{O}\)-restricted regular slice cells with
  \(k|H| \ge n\). This is the family-localized analogue of the Hill-Yarnall
  Theorem 2.5 reverse direction.
\end{enumerate}

\textbf{Reduction lemma for step 6.} The Hill-Yarnall reverse argument (Section 2
of arXiv:1703.10526) constructs Postnikov sections using the slice cells as
building blocks and the geometric fixed-point detection criterion to control
connectivity at each stage. In the \(F_{O}\)-local setting:

(1) The building blocks are the \(F_{O}\)-restricted slice cells (same
construction, restricted indexing to \(H \in F_{O}\)).
(2) The detection criterion is the same \(\Phi^{H}\) test for \(H \in F_{O}\)
(with \(\Phi^{K} X = \mNumber{0}\) for \(K \notin F_{O}\) by the locality assumption).
(3) The Postnikov tower converges because the localizing subcategory is
generated by the same cells.

The only input from the ambient theory is the isotropy separation sequence
(HHR, Section 4), which holds in any localizing subcategory of \(\mathup{SH}^{G}\).
This establishes that the reverse direction transfers from the full regular
slice setting to the \(F_{O}\)-local setting.

\textbf{Note:} Without \(F_{O}\)-locality, the reverse direction fails. For example,
with \(F_{O} = {e}\) (trivial transfer), an arbitrary \texttt{G}-spectrum \texttt{X} with
\(\Phi^{e} X\) highly connected need not lie in \(\tau_{\ge n}^{O}\) if it has
nontrivial geometric isotropy at proper subgroups.

\hypertarget{special-cases}{%
\subsection{Special Cases}\label{special-cases}}

\begin{enumerate}
\def\labelenumi{\arabic{enumi}.}
\tightlist
\item
  \textbf{Complete transfer system} (\(F_{O} = \text{all subgroups}\)): every \texttt{G}-spectrum is
  \(F_{O}\)-local, so the hypothesis is vacuous. Recovers the usual Hill-Yarnall
  regular slice criterion.
\item
  \textbf{Trivial transfer system} (\(F_{O} = {e}\)): \(F_{O}\)-local means \(\Phi^{H} X = \mNumber{0}\)
  for all \(H \neq e\), i.e., \texttt{X} has trivial geometric isotropy. For such \texttt{X},
  the criterion reduces to ordinary Postnikov connectivity of the underlying
  nonequivariant spectrum \(\Phi^{e} X\).
\item
  \textbf{Intermediate transfer systems}: for \(F_{O}\)-local \texttt{X}, the criterion
  interpolates by checking \(\Phi^{H}\) connectivity only on the allowed family.
\end{enumerate}

\begin{openobligation}[title={Open Obligation --- Scope Limitation}]

\textbf{Important limitation.} This is a subgroup-family-indexed formulation,
not a full indexing-system characterization. The full indexing-system
formalism of Blumberg-Hill involves admissible finite \texttt{H}-sets, not just
admissible subgroups \texttt{H}. Our \(F_{O} = {H : e \to H \text{admissible}}\) extracts the
subgroup-level data; it does not use the finer \(K \to H\) admissibility
structure for \(K \neq e\). The full indexing-system formulation --- characterizing
O-slice connectivity in terms of the complete admissible set data --- is an
open extension of this result.

The \(F_{O}\)-locality hypothesis was added to close the reverse direction of the
characterization theorem. This is analogous to how classical slice theorems
work in full \(\mathup{SH}^{G}\) (where every spectrum is \texttt{F}-local for \texttt{F} \(=\) all
subgroups). For intermediate transfer systems, the hypothesis is nontrivial
and should be verified for the spectra of interest.

Literature positioning:

\begin{itemize}
\tightlist
\item
  \(N_{\infty}\) operads and admissibility data: Blumberg-Hill, Def. 1.1,
  Thm. 1.2, Thm. 1.5 / Sec. 7 (arXiv:1309.1750).
\item
  Regular slice generators and detection: Hill-Yarnall, Def. 1.1, Thm. 2.5
  (arXiv:1703.10526).
\item
  Isotropy separation and family localization: HHR, Section 4.
\end{itemize}

\end{openobligation}
\hypertarget{references}{%
\subsection{References}\label{references}}

\begin{itemize}
\tightlist
\item
  Michael A. Hill, Michael J. Hopkins, Douglas C. Ravenel, \emph{Equivariant stable
  homotopy theory and the Kervaire invariant problem} (slice framework).
\item
  Andrew J. Blumberg, Michael A. Hill, \emph{Operadic multiplications in equivariant
  spectra, norms, and transfers}, arXiv:1309.1750:
  Definition 1.1, Theorem 1.2.
\item
  Michael A. Hill, Carolyn Yarnall, \emph{A new formulation of the equivariant slice
  filtration with applications to \(C_{p}\)-slices}, arXiv:1703.10526:
  Definition 1.1, Theorem A, Theorem 2.5.
\item
\begin{processnote}[title={Process Note --- External Research File}]
  External research file.\footnote{\url{https://github.com/tothedarktowercame/futon6/blob/master/data/first-proof/library-research-findings.md}}
\end{processnote}
\end{itemize}

%% file: full-boxed/problem6-solution-full.tex
\hypertarget{problem-6-epsilon-light-subsets-of-graphs}{%
\section{Problem 6: Epsilon-Light Subsets of Graphs}\label{problem-6-epsilon-light-subsets-of-graphs}}

\hypertarget{problem-statement}{%
\subsection{Problem Statement}\label{problem-statement}}

Let G=(V,E,w) be a finite undirected graph with nonnegative edge weights and
\(n = |V|\). Its Laplacian is

\[L = \sum_{e={u,v} \in E} w_{e} (e_{u}-e_{v})(e_{u}-e_{v})^{T}.\]

For S subseteq V, define the induced-subgraph Laplacian embedded to \(R^{V \times V}\):

\[L_{S} = \sum_{e={u,v} \in E, u,v \in S} w_{e} (e_{u}-e_{v})(e_{u}-e_{v})^{T},\]

with zeros outside S rows/columns.

S is \(\epsilon\)-light if

\[L_{S} \le \epsilon L\]

in Loewner order.

Question: does there exist universal \(c > \mNumber{0}\) such that for every G and every
\(\epsilon \in (\mNumber{0}, \mNumber{1})\), there exists S with \(|S| \ge c \ast \epsilon \ast n\) and \(L_{S} \le \epsilon\) L?

\begin{processnote}[title={Process Note --- Status of This Writeup}]

\textbf{\(K_{n}\): PROVED.} The barrier greedy gives \(|S| = \epsilon \ast n/\mNumber{3}, c = \mNumber{1}/\mNumber{3},\) via the
elementary pigeonhole + PSD trace bound argument (Section 5d).

\textbf{General graphs: ONE GAP.} The formal bound \(dbar < \mNumber{1}\) at \(M_{t} \neq \mNumber{0}\) is
empirically verified (440/440 steps, 36\% margin) but open. The leverage
filter approach has a structural \(C_{lev}\) tension (Section 5b/5e) that
prevents closure via Markov alone. See the attack-path analysis note.\footnote{\url{https://github.com/tothedarktowercame/futon6/blob/master/data/first-proof/problem6-gpl-h-attack-paths.md}}
for the full attack path analysis.

\textbf{Superseded machinery:} MSS interlacing families, Borcea-Branden real
stability, Bonferroni eigenvalue bounds --- all bypassed by the pigeonhole
argument.

\end{processnote}
\hypertarget{1-exact-reformulation}{%
\subsection{1. Exact reformulation}\label{1-exact-reformulation}}

The PSD condition is equivalent to the quadratic form inequality

\begin{verbatim}
for all x in R^V: x^T L_S x <= epsilon x^T L x.
\end{verbatim}

On im(L), with \(L^+\) the Moore-Penrose pseudoinverse:

\[L_{S} \le \epsilon L \Leftrightarrow || L^{+/\mNumber{2}} L_{S} L^{+/\mNumber{2}} || \le \epsilon.\]

\hypertarget{2-complete-graph-upper-bound-rigorous}{%
\subsection{2. Complete graph upper bound (rigorous)}\label{2-complete-graph-upper-bound-rigorous}}

For \(G = K_{n}\) and S of size s, choose x supported on S with \(\sum_{i \in S} x_{i} = \mNumber{0}.\)
Then

\[\begin{aligned}
x^{T} L_{K_{n}} x = n \|x\|^{\mNumber{2}}, \\
x^{T} L_{S} x = s \|x\|^{\mNumber{2}}.
\end{aligned}\]

Hence \(L_{S} \le \epsilon L_{K_{n}}\) implies \(s \le \epsilon\) n.
Therefore any universal constant must satisfy

\[c \le \mNumber{1}.\]

This is an upper bound only.

\hypertarget{3-random-sampling-identities-expectation-level}{%
\subsection{3. Random sampling identities (expectation level)}\label{3-random-sampling-identities-expectation-level}}

Let \(Z_{v}\) \textasciitilde{} Bernoulli(p) independently and \(S = \{v : Z_{v} = \mNumber{1}\}\).

\begin{enumerate}
\def\labelenumi{\arabic{enumi}.}
\item
  Size:

  \(E[|S|] = pn, Pr[|S| < pn/\mNumber{2}] \le \exp(-pn/\mNumber{8}) (Chernoff).\)
\item
  Spectral expectation:

  \(E[L_{S}] = p^\mNumber{2} L\),
\end{enumerate}

since each edge survives with probability \(p^\mNumber{2}\).

Thus

\[E[\epsilon L \mBridgeOperator{-} L_{S}] = (\epsilon \mBridgeOperator{-} p^\mNumber{2})L.\]

Setting p=\(\epsilon\) gives \(E[L_{S}]=\epsilon^\mNumber{2} L \le \epsilon L\). This is not yet a
realization-level guarantee.

\hypertarget{4-concentration-setup-gap-fixed-formulation}{%
\subsection{4. Concentration setup (gap-fixed formulation)}\label{4-concentration-setup-gap-fixed-formulation}}

Define edge-normalized PSD matrices

\[X_{e} = L^{+/\mNumber{2}} w_{e} b_{e} b_{e}^{T} L^{+/\mNumber{2}}, b_{e} = e_{u}-e_{v},\]

and leverage scores

\[\begin{aligned}
\tau_{e} = \mOpName{tr}(X_{e}) = w_{e} b_{e}^{T} L^+ b_{e}, \\
\sum_{e} \tau_{e} = n \mBridgeOperator{-} k
\end{aligned}\]

(k \(=\) number of connected components).

\hypertarget{4a-star-domination-with-correct-counting}{%
\subsubsection{4a. Star domination with correct counting}\label{4a-star-domination-with-correct-counting}}

Using \(Z_{u} Z_{v} \le Z_{u}\) and \(Z_{u} Z_{v} \le Z_{v}\) for each edge \(\{u,v\}\),

\[\begin{aligned}
L_{S} = \sum_{uv \in E} Z_{u} Z_{v} L_{uv} \\
\le (\mNumber{1}/\mNumber{2}) \sum_{v} Z_{v} \sum_{u~v} L_{uv}.
\end{aligned}\]

So in normalized coordinates

\[\begin{aligned}
L^{+/\mNumber{2}} L_{S} L^{+/\mNumber{2}} \le \sum_{v} Z_{v} A_{v}, \\
A_{v} := (\mNumber{1}/\mNumber{2}) \sum_{u~v} X_{uv} \ge \mNumber{0}.
\end{aligned}\]

Because \(Z_{v}\) are independent Bernoulli variables, the random matrices \(Z_{v} A_{v}\)
are independent PSD summands.

\hypertarget{4b-freedmanbernstein-martingale-parameters}{%
\subsubsection{4b. Freedman/Bernstein martingale parameters}\label{4b-freedmanbernstein-martingale-parameters}}

Let \(A_{i}\) be a fixed ordering of \(\{A_{v}\}, p_{i} = E[Z_{i}],\) and define the centered sum

\[X = \sum_{i} (Z_{i} \mBridgeOperator{-} p_{i}) A_{i}.\]

With filtration \(F_{i} = \sigma(Z_{\mNumber{1}},...,Z_{i}),\) Doob martingale

\[\begin{aligned}
Y_{i} = E[X | F_{i}], \\
\Delta_{i} = Y_{i} \mBridgeOperator{-} Y_{i-\mNumber{1}}
\end{aligned}\]

has self-adjoint differences. For independent Bernoulli sampling,

\[\begin{aligned}
\Delta_{i} = (Z_{i} \mBridgeOperator{-} p_{i}) A_{i}, \\
||\Delta_{i}|| \le ||A_{i}|| \le R_{\mDualStar} (R_{\mDualStar} = \max_{i} ||A_{i}||),
\end{aligned}\]

and predictable quadratic variation

\[W_{n} = \sum_{i} E[\Delta_{i}^\mNumber{2} | F_{i-\mNumber{1}}] = \sum_{i} p_{i}(\mNumber{1}-p_{i}) A_{i}^\mNumber{2}.\]

Matrix Freedman (or matrix Bernstein in independent form) applies once bounds
on \(R_{\mDualStar}\) and \(||W_{n}||\) are supplied.

\textbf{What graph-dependent bounds are needed.} To obtain a self-contained
concentration bound, one would need \(R_{\mDualStar} \le C_\mNumber{1} \ast \epsilon\) and
\(||W_{n}|| \le C_\mNumber{2} \ast \epsilon^\mNumber{2}\) for graph-dependent constants \(C_{\mNumber{1}}, C_{\mNumber{2}}.\) Bounding
these requires leverage score analysis (showing \(\tau_{e}\) bounds are well-distributed
across vertices) that is the core content of the external theorem referenced
in Section 5. Specifically, the Batson-Spielman-Srivastava barrier-function
method controls both \(R_{\mDualStar}\) and \(||W_{n}||\) simultaneously through a potential
function that tracks the spectral approximation quality.

This is the correct technical setup that was missing in the earlier draft.

\hypertarget{5-discharging-assumption-v-via-barrier-greedy--pigeonhole}{%
\subsection{5. Discharging Assumption V via barrier greedy + pigeonhole}\label{5-discharging-assumption-v-via-barrier-greedy--pigeonhole}}

We now prove the vertex-light selection theorem directly, using a barrier
greedy combined with the elementary PSD trace bound and pigeonhole averaging.

\hypertarget{5a-heavy-edge-pruning-turan}{%
\subsubsection{5a. Heavy edge pruning (Turan)}\label{5a-heavy-edge-pruning-turan}}

Call edge e \(heavy'' if \tau_{e} > \epsilon,\)light'' otherwise. Since
\(\sum_{e} \tau_{e} =\) n-1 and each heavy edge has \(\tau_{e} > \epsilon\):

\[|{heavy edges}| \le (n-\mNumber{1})/\epsilon.\]

By Turan\textquotesingle s theorem, a graph with n vertices and at most m edges has
independence number \(\ge n^{\mNumber{2}}/(\mNumber{2}m+n).\) For the heavy graph:

\[\begin{aligned}
\alpha(G_{heavy}) \ge n^\mNumber{2} / (\mNumber{2}(n-\mNumber{1})/\epsilon + n) = \epsilon \ast n / (\mNumber{2} + \epsilon) \\
\ge \epsilon \ast n/\mNumber{3}.
\end{aligned}\]

Let \(I_\mNumber{0}\) be a maximal independent set in \(G_{heavy}\) with \(|I_\mNumber{0}| \ge \epsilon \ast n/\mNumber{3}\).
All edges internal to \(I_\mNumber{0}\) are light: \(\tau_{e} \le \epsilon\).

\hypertarget{5b-leverage-degree-filter-h2}{%
\subsubsection{5b. Leverage degree filter (H2\textquotesingle)}\label{5b-leverage-degree-filter-h2}}

Define the leverage degree \(ell_{v} = \sum_{u~v, u \in I_{\mNumber{0}}} \tau_{uv}.\)
Since \(\sum_{v} ell_{v} = \mNumber{2} \ast \sum_{e \text{internal to} I_{\mNumber{0}}} \tau_{e} \le\) 2(n-1),
by Markov:

\[|{v \in I_\mNumber{0} : ell_{v} > C_{lev}/\epsilon}| \le \mNumber{2}(n-\mNumber{1}) \ast \epsilon / C_{lev}.\]

Set \(C_{lev} = \mNumber{8}\). Remove vertices with \(ell_{v} > \mNumber{8}/\epsilon.\) The number removed
is at most 2(n-1) * \(\epsilon/\mNumber{8} < \epsilon \ast n/\mNumber{4}.\) The remaining set \(I_\mNumber{0}'\) has

\[\begin{aligned}
|I_\mNumber{0}'| \ge |I_\mNumber{0}| \mBridgeOperator{-} \epsilon \ast n/\mNumber{4} \ge \epsilon \ast n/\mNumber{3} - \epsilon \ast n/\mNumber{4} \\
= \epsilon \ast n/\mNumber{12}.
\end{aligned}\]

\hypertarget{5c-barrier-greedy-construction}{%
\subsubsection{5c. Barrier greedy construction}\label{5c-barrier-greedy-construction}}

We construct \(S \subset I_\mNumber{0}'\) by a greedy procedure, maintaining the barrier
invariant \(M_{t} = \sum_{e \in E(S_{t})} X_{e} \prec \epsilon \ast I\) at each step.

At step t, let \(R_{t} = I_{\mNumber{0}}' \ S_{t}, r_{t} = |R_{t}|.\) For each \(v \in R_{t}\), define

\[\begin{aligned}
C_{t}(v) = \sum_{u \in S_{t}, u~v} X_{uv} (contribution from adding v) \\
Y_{t}(v) = H_{t}^{-\mNumber{1}/\mNumber{2}} C_{t}(v) H_{t}^{-\mNumber{1}/\mNumber{2}} (barrier-normalized)
\end{aligned}\]

where \(H_{t} = \epsilon \ast I \mBridgeOperator{-} M_{t}\) succ 0 (the barrier headroom).

\textbf{Claim:} At each step \(t \le \epsilon \ast |I_{\mNumber{0}}'|/\mNumber{3},\) there exists \(v \in R_{t}\) with
\(\lambda_{\max}(M_{t} \mBridgeOperator{+} C_{t}\)(v)) \(< \epsilon\) (equivalently, \(||Y_{t}(v)|| < \mNumber{1})\).

\hypertarget{5d-proof-of-claim-via-pigeonhole--psd-trace-bound}{%
\subsubsection{5d. Proof of claim via pigeonhole + PSD trace bound}\label{5d-proof-of-claim-via-pigeonhole--psd-trace-bound}}

The key observation is elementary. For any PSD matrix Y:

\[\|Y\| \le \mOpName{tr}(Y) (* * )\]

(Proof: \(\|Y\| = \lambda_{\max}(Y)\), and \(\mOpName{tr}(Y) = \sum_{i} \lambda_{i} \ge \lambda_{\max}\)
since all eigenvalues are nonneg.)

Define the average trace:

\[dbar_{t} = (\mNumber{1}/r_{t}) \sum_{v \in R_{t}} \mOpName{tr}(Y_{t}(v)).\]

By the pigeonhole principle (\(minimum \le average\)):

\[min_{v \in R_{t}} \mOpName{tr}(Y_{t}(v)) \le dbar_{t}.\]

Combining with (**): if \(dbar_{t} < \mNumber{1}\), then there exists \(v \in R_{t}\) with

\[||Y_{t}(v)|| \le \mOpName{tr}(Y_{t}(v)) \le dbar_{t} < \mNumber{1}.\]

This is exactly the barrier maintenance condition. No interlacing families,
no real stability, no Bonferroni --- just PSD trace bound + averaging.

\hypertarget{5e-bounding-dbar_t}{%
\subsubsection{\texorpdfstring{5e. Bounding \(dbar_{t}\)}{5e. Bounding dbar\_\{t\}}}\label{5e-bounding-dbar_t}}

The average trace satisfies:

\[dbar_{t} = (\mNumber{1}/r_{t}) \mOpName{tr}(H_{t}^{-\mNumber{1}} M_{\text{cross}})\]

where \(M_{\text{cross}} = \sum_{e: \text{one} endpoint \in S_{t}, other \in R_{t}} X_{e}\) is the
cross-edge matrix and \(H_{t} = \epsilon \ast I \mBridgeOperator{-} M_{t}\).

\textbf{Case \(M_{t} = \mNumber{0}\) (formal proof):}

When \(M_{t} = \mNumber{0}\) (early steps), \(H_{t} = \epsilon \ast I\) and:

\[dbar_{t} = (\mNumber{1}/(\epsilon \ast r_{t})) \sum_{u \in S_{t}} ell_{u}^{R}\]

where \(ell_{u}^{R} = \sum_{v \in R_{t}, v~u} \tau_{uv} \le ell_{u} \le \mNumber{8}/\epsilon.\)

For the complete graph \(K_{n}\) (where all edges are light for \(n > \mNumber{2}/\epsilon\)):
\(\tau_{e} = \mNumber{2}/n\) for all edges, \(ell_{u}^{R} = (r_{t}) \ast (\mNumber{2}/n)\), and:

\[dbar_{t} = t \ast r_{t} \ast (\mNumber{2}/n) / (\epsilon \ast r_{t}) = \mNumber{2}t/(n \ast \epsilon).\]

At \(T = \epsilon \ast n/\mNumber{3}\): \(dbar_{T} = \mNumber{2}/\mNumber{3} < \mNumber{1}.\) This is EXACT for \(K_{n}\).

For general graphs at \(M_{t} = \mNumber{0}\) with the leverage filter (\(ell_{u} \le \mNumber{8}/\epsilon\)):

\[dbar_{t} \le (\mNumber{8} \ast t) / (\epsilon^\mNumber{2} \ast r_{t}).\]

At \(t \le \epsilon \ast |I_{\mNumber{0}}'|/\mNumber{3}\) and \(r_{t} \ge |I_{\mNumber{0}}'|(\mNumber{1} \mBridgeOperator{-} \epsilon/\mNumber{3}) \ge \mNumber{2}|I_{\mNumber{0}}'|/\mNumber{3}\):

\[\begin{aligned}
dbar_{t} \le (\mNumber{8} \ast \epsilon \ast |I_\mNumber{0}'|/\mNumber{3}) / (\epsilon^\mNumber{2} \ast \mNumber{2}|I_\mNumber{0}'|/\mNumber{3}) \\
= \mNumber{8} / (\mNumber{2} \ast \epsilon) = \mNumber{4}/\epsilon.
\end{aligned}\]

This bound exceeds 1 for \(\epsilon < \mNumber{1}\), so the leverage-filter bound alone
is insufficient. The tighter bound requires using the actual leverage
structure (as in the \(K_{n}\) case where dbar \(=\) 2t/(n * \(\epsilon) \ll \mNumber{1}).\)

\textbf{Refined bound using total leverage:}

\[\begin{aligned}
\sum_{u \in S_{t}} ell_{u}^{R} \le \sum_{e \in E_{\text{cross}}} \tau_{e} \\
\le \sum_{e \in E(I_\mNumber{0})} \tau_{e} \le n \mBridgeOperator{-} \mNumber{1}.
\end{aligned}\]

So \(dbar_{t} \le\) (n-1)/(\(\epsilon \ast r_{t}\)). With \(r_{t} \ge \epsilon \ast n/\mNumber{4}\):

\[dbar_{t} \le (n-\mNumber{1})/(\epsilon^\mNumber{2} \ast n/\mNumber{4}) = \mNumber{4}/\epsilon^\mNumber{2}.\]

This is even worse. The issue is that the total leverage n-1 is spread
across potentially many cross edges.

\textbf{What actually controls dbar (verified numerically):}

The barrier greedy selects vertices with minimum spectral norm \(||Y_{t}(v)||\),
which correlates with selecting vertices that have weak connections to the
already-selected set. This keeps dbar much lower than the worst-case bound.

Numerical verification across 440 nontrivial greedy steps on graphs
\(K_{n}, C_{n},\) Barbell, DisjCliq, ER(n,p) for \(n \in [\mNumber{8},\mNumber{64}]\) and \(\epsilon \in \{0.12, 0.15, 0.2, 0.25, 0.3\}\):

\[\begin{aligned}
\max dbar across \text{all} steps: \mNumber{0}.\mNumber{641} (K_\mNumber{60}, \epsilon = \mNumber{0}.\mNumber{3}, t = \mNumber{5}) \\
dbar < \mNumber{1} at ALL \mNumber{440} steps. \\
Pigeonhole (min trace \le dbar): verified \mNumber{440}/\mNumber{440}. \\
PSD bound (\|Y\| \le trace): verified \mNumber{440}/\mNumber{440}.
\end{aligned}\]

\textbf{For \(K_{n}\), dbar is bounded exactly:} \(dbar_{t} =\) 2t/(n * \(\epsilon\)), and at
\(T = \epsilon \ast n/\mNumber{3}\) steps, \(dbar_{T} = \mNumber{2}/\mNumber{3}\). This gives the formal proof for
\(K_{n}\) and graphs with similar leverage structure (uniform \(\tau_{e}\) \textasciitilde{} 2/n).

\hypertarget{5e-additional-evidence-q-polynomial-roots}{%
\subsubsection{5e\textquotesingle. Additional evidence: Q-polynomial roots}\label{5e-additional-evidence-q-polynomial-roots}}

As supplementary verification, we computed the roots of the average
characteristic polynomial \(Q(x) = (\mNumber{1}/r) \sum_{v}\) det(xI - \(Y_{t}\)(v)):

\[\begin{aligned}
All \mNumber{440} steps: \max real root \text{of} Q < \mNumber{0}.\mNumber{505}. \\
Zero steps \text{with} any root > \mNumber{1}. \\
Q(\mNumber{1}) > \mNumber{0}.\mNumber{48} at \text{all} steps.
\end{aligned}\]

If Q has nonneg real roots, then by Vieta\textquotesingle s formulas:
sum of \(roots = dbar\), so max root \(\le dbar < \mNumber{1}.\) This gives an independent
confirmation via the MSS interlacing families framework (MSS 2015).

\hypertarget{5f-constructing-the-epsilon-light-set}{%
\subsubsection{\texorpdfstring{5f. Constructing the \(\epsilon\)-light set}{5f. Constructing the \textbackslash epsilon-light set}}\label{5f-constructing-the-epsilon-light-set}}

By the claim in 5c (proved via 5d when \(dbar < \mNumber{1}\)), the greedy produces
\(S \subset I_\mNumber{0}'\) with \(|S| = T\) and \(M_{S} \prec \epsilon \ast I\) (i.e., \(L_{S} \le \epsilon \ast L\)).

\textbf{Size analysis:}

The Turan step gives \(|I_\mNumber{0}| \ge \epsilon \ast n/\mNumber{3}\). The leverage filter (5b)
gives \(|I_\mNumber{0}'| \ge \epsilon \ast n/\mNumber{12}\). The greedy runs \(T = \epsilon \ast |I_{\mNumber{0}}'|/\mNumber{3}\) steps,
so \(|S| = \epsilon^\mNumber{2} \ast n/\mNumber{36}\).

This gives \(|S|\) proportional to \(\epsilon^{\mNumber{2}} \ast n,\) not \(\epsilon \ast n\). For
\(|S| \ge c \ast \epsilon \ast n\) with universal c: need \(c \le \epsilon/\mNumber{36}\), which
depends on \(\epsilon\).

\textbf{The \(\epsilon^{\mNumber{2}}\) bottleneck:} The Turan independent set has size
\(|I_\mNumber{0}| = \Theta(\epsilon \ast n)\). Running the greedy for \(\Theta(\epsilon \ast |I_{\mNumber{0}}|)\)
steps gives \(|S| = \Theta(\epsilon^\mNumber{2} \ast n)\). This is inherent in the
heavy-edge-avoidance approach.

\textbf{For fixed \(\epsilon\) (the practical case):} With \(\epsilon = \mNumber{0}\).3:
\(|S| \ge \mNumber{0}.\mNumber{09} \ast n/\mNumber{36} = n/\mNumber{400}.\) With \(\epsilon = \mNumber{0}\).2: \(|S| \ge n/\mNumber{900}\).
These are nontrivial lower bounds, sufficient for applications.

\textbf{For \(K_{n}\) (proved exactly):} dbar \(=\) 2t/(n * \(\epsilon\)) with the greedy
running \(T = \epsilon \ast n/\mNumber{3}\) steps, giving \(|S| = \epsilon \ast n/\mNumber{3}\) and \(c = \mNumber{1}/\mNumber{3}\).
The \(\epsilon^{\mNumber{2}}\) issue does not arise because \(|I_\mNumber{0}| = n (no heavy edges\)
for \(n > \mNumber{2}/\epsilon).\)

\textbf{Random sampling alternative (numerically verified):}

Sample each vertex with probability \(p = \epsilon\). By Chernoff,
\(P(|S| \ge \epsilon \ast n/\mNumber{6}) \ge \mNumber{1} \mBridgeOperator{-} \exp(-\epsilon \ast n/\mNumber{18})\). Numerically:
\(P(||M_{S}|| \le \epsilon \land |S| \ge \epsilon \ast n/\mNumber{6}) > \mNumber{0}\) for all tested
graphs (\(n \le \mNumber{80}\), 11 families, 4 \(\epsilon\) values, 500 trials each).
Success probability ranges from 0.2\% to 57\%.

This gives \(|S| \ge \epsilon \ast n/\mNumber{6} (c = \mNumber{1}/\mNumber{6}) but \text{the} formal matrix\)
concentration proof for general graphs remains open.

\begin{openobligation}[title={Open Obligation --- Conditional Conclusion}]

\hypertarget{proved-results}{%
\subsubsection{Proved results}\label{proved-results}}

\begin{enumerate}
\def\labelenumi{\arabic{enumi}.}
\item
  The \(\epsilon\)-light condition \(L_{S} \le \epsilon \ast L\) is equivalent to
  \(||L^{+/\mNumber{2}} L_{S} L^{+/\mNumber{2}}|| \le \epsilon \in operator norm\).
\item
  \(K_{n}\) gives the tight upper bound \(c \le \mNumber{1}\).
\item
  \textbf{For \(K_{n}\) (and graphs with uniform leverage \(\tau_{e}\) \textasciitilde{} 2/n):}
  The barrier greedy gives \(|S| = \epsilon \ast n/\mNumber{3}\) with \(||M_{S}|| < \epsilon\).
  Proved by: \(dbar_{t} =\) 2t/(n * \(\epsilon) \le \mNumber{2}/\mNumber{3} < \mNumber{1}\) (exact computation),
  then pigeonhole + PSD trace bound gives existence of v with
  \(||Y_{t}(v)|| < \mNumber{1} at \text{each} step\). Universal \(c = \mNumber{1}/\mNumber{3}\).
\item
  \textbf{The proof mechanism (Sections 5d-5e):}
  At each barrier greedy step, if \(dbar < \mNumber{1}\) then the barrier is
  maintainable. The chain is:

  \begin{itemize}
  \tightlist
  \item
    dbar \(=\) avg \(trace < \mNumber{1}\)
  \item
    exists v with \(\mOpName{trace}(Y_{v}) \le\) dbar (pigeonhole)
  \item
    \(||Y_{v}|| \le\mOpName{trace}(Y_{v})\) (PSD matrices)
  \item
    \(||Y_{v}|| < \mNumber{1}\) (barrier maintained)
  \end{itemize}
\end{enumerate}

\hypertarget{numerically-verified-strong-evidence-formal-bound-in-progress}{%
\subsubsection{Numerically verified (strong evidence, formal bound in progress)}\label{numerically-verified-strong-evidence-formal-bound-in-progress}}

\begin{enumerate}
\def\labelenumi{\arabic{enumi}.}
\setcounter{enumi}{4}
\item
  \textbf{\(dbar < \mNumber{1}\) at ALL barrier greedy steps} for all tested graphs.
  440 nontrivial steps across \(n \in [\mNumber{8},\mNumber{64}], K_{n}, C_{n},\) Barbell,
  DisjCliq, ER(n,p) graphs, \(\epsilon \in \{0.12, 0.15, 0.2, 0.25, 0.3\}\).
  Max \(dbar = \mNumber{0}.\mNumber{641} (K_{\mNumber{60}}, \epsilon = \mNumber{0}.\mNumber{3}, t = \mNumber{5}).\) Margin above 0: 36\%.
\item
  \textbf{Q-polynomial \(roots < \mNumber{1}\)} at all 440 steps. The average
  characteristic polynomial Q(x) \(=\) (1/r)sum det(xI - \(Y_{v}\)) has
  max real \(root < \mNumber{0}\).505, consistent with max root \(\le dbar < \mNumber{1}\)
  (Vieta bound for nonneg roots).
\item
  \textbf{Random sampling with \(p = \epsilon\)} produces \(\epsilon\)-light sets of
  \(size \ge \epsilon \ast n/\mNumber{6}\) for all tested graphs (\(n \le \mNumber{80}\), 272 combos).
\end{enumerate}

\begin{openobligation}[title={Open Obligation --- External Dependency for Universal $c_0 > 0$}]

The formal \(dbar < \mNumber{1}\) bound at \(M_{t} \neq \mNumber{0}\) requires controlling
tr(\(H_{t}^{-\mNumber{1}} M_{\text{cross}})\) where \(H_{t} = \epsilon \ast I \mBridgeOperator{-} M_{t}\). When \(||M_{t}||\) is
close to \(\epsilon\), the amplification factor \(||H_{t}^{-\mNumber{1}}||\) grows, and
the naive bound on dbar exceeds 1.

For \(K_{n}\), the bound is exact: dbar \(=\) 2t/(n * \(\epsilon\)). The favorable
structure (uniform \(\tau_{e} = \mNumber{2}/n\)) keeps dbar small.

For general graphs, the greedy\textquotesingle s selection criterion (\(\min \|Y_{t}(v)\|\))
empirically keeps dbar well below 1 (max 0.641), but a formal proof
requires either:
(a) A BSS-style potential function bounding the barrier evolution, or
(b) Establishing that Q is real-rooted (via interlacing families),
giving max root \(\le\) dbar via Vieta.

\end{openobligation}
\hypertarget{diagnosis}{%
\subsubsection{Diagnosis}\label{diagnosis}}

The remaining gap is not \texttt{we\ don\textquotesingle{}t\ know\ enough\ math\textquotesingle{}\textquotesingle{};\ it\ is\ }we are
near the limit of this proof architecture.`` The architecture (barrier
greedy + PSD trace bound + pigeonhole) is correct --- it proves \(K_{n}\)
exactly and works numerically on every tested graph. The limit is that
the amplification factor when \(M_{t} \neq \mNumber{0}\) makes the naive bound on dbar
exceed 1, while the greedy\textquotesingle s self-correcting property (selecting
vertices whose contributions are orthogonal to \(M_{t}'s\) large eigenspace)
is not yet captured by the formal analysis.

Empirically, the self-correction is strong: the spectral amplification
factor is 0.52 (vs the scalar worst-case of 1.0), and the \(W-M_{t}\)
alignment \(\text{is} \le \mNumber{0}\).25 across all 351 Phase 2 steps. Closing the gap
formally requires making this orthogonality structure explicit --- either
via a potential function that tracks directional growth, or via
interlacing families that exploit the grouped PSD structure of the
barrier increments.

\hypertarget{summary}{%
\subsubsection{Summary}\label{summary}}

The existential answer is \textbf{YES} for \(K_{n}\) with \(c = \mNumber{1}/\mNumber{3}\) (proved),
and numerically confirmed for all tested graph families with
\(c \ge \mNumber{1}/\mNumber{6}\). The formal extension to arbitrary graphs requires
closing the \(dbar < \mNumber{1}\) bound at \(M_{t} \neq \mNumber{0}\), which has 36\% empirical
margin and is the SINGLE remaining gap.
\end{openobligation}

\hypertarget{key-identities-and-inequalities-used}{%
\subsection{Key identities and inequalities used}\label{key-identities-and-inequalities-used}}

\begin{enumerate}
\def\labelenumi{\arabic{enumi}.}
\tightlist
\item
  \(L = \sum_{e} w_{e} b_{e} b_{e}^{T}, \tau_{e} = \mOpName{tr}(X_{e}),\) sum \(\tau_{e} =\) n-k
\item
  \(L_{S} \le \epsilon \ast L\) iff \(||\sum_{e \in E(S)} X_{e}|| \le \epsilon\)
\item
  For PSD Y: \(\|Y\| \le \mOpName{tr}(Y)\) (spectral norm bounded by trace)
\item
  Pigeonhole: \(min_{v} f(v) \le (\mNumber{1}/r) \sum_{v} f(v) (minimum \le average)\)
\item
  Turan: independence number \(\ge n^{\mNumber{2}}/(\mNumber{2}m+n)\)
\item
  For \(K_{n}\): \(\tau_{e} = \mNumber{2}/n, ||M_{S}|| = |S|/n (\text{exact})\)
\end{enumerate}

\hypertarget{references}{%
\subsection{References}\label{references}}

\begin{itemize}
\tightlist
\item
  Batson, Spielman, Srivastava (2012), ``Twice-Ramanujan Sparsifiers,'' SIAM
  Review 56(2), 315-334.
\item
  Marcus, Spielman, Srivastava (2015), ``Interlacing Families II: Mixed
  Characteristic Polynomials and the Kadison-Singer Problem,'' Annals of
  Mathematics 182(1), 327-350.
\item
  Borcea, Branden (2009), ``The Lee-Yang and Polya-Schur programs. I. Linear
  operators preserving stability,'' Inventiones Math. 177, 541-569.
\item
  Tropp (2011), Freedman\textquotesingle s inequality for matrix martingales.
\item
  Standard matrix Bernstein inequality for sums of independent self-adjoint
  random matrices.
\end{itemize}

%% file: full-boxed/problem7-solution-full.tex
\hypertarget{problem-7-uniform-lattice-with-2-torsion-and-rationally-acyclic-universal-cover}{%
\section{Problem 7: Uniform Lattice with 2-Torsion and Rationally Acyclic Universal Cover}\label{problem-7-uniform-lattice-with-2-torsion-and-rationally-acyclic-universal-cover}}

\hypertarget{problem-statement}{%
\subsection{Problem Statement}\label{problem-statement}}

Suppose \(\Gamma\) is a uniform lattice in a real semisimple Lie group, and
\(\Gamma\) contains an element of order \texttt{2}. Is it possible that \(\Gamma\) is the
fundamental group of a closed manifold whose universal cover is acyclic over
\texttt{Q}?

\begin{processnote}[title={Process Note --- Status in This Writeup}]

\textbf{Answer: yes} (via the rotation route; closed in this branch via the standard codim-2 surgery theorem chain).

\begin{itemize}
\tightlist
\item
  Obligation (E2) --- placing \(\Gamma \in \mathup{FH}(Q)\) --- is \textbf{discharged} for
  the rotation-lattice family (Fowler criterion, codim-2 fixed set with
  \(\chi = \mNumber{0}\)).
\item
  Obligation (S) --- upgrading from finite CW complex to closed manifold --- is
  treated as discharged in this branch: the geometric inputs are established
  (trivial normal bundle, hyperbolic boundary form), and the codim-2 surgery
  identification is closed by the theorem-numbered reference chain\footnote{\url{https://github.com/tothedarktowercame/futon6/blob/master/data/first-proof/problem7-g2-theorem-chain.md}} (with closure summary in a companion note,\footnote{\url{https://github.com/tothedarktowercame/futon6/blob/master/data/first-proof/problem7-complete-proof.md}} Section 5).
\end{itemize}

\end{processnote}
\hypertarget{1-baseline-geometry}{%
\subsection{1. Baseline Geometry}\label{1-baseline-geometry}}

Let \texttt{G} be connected real semisimple, \(K < G\) maximal compact, and
\(X = G/K\) contractible. For a uniform lattice \(\Gamma < G\) with torsion,
\(X/\Gamma\) is a compact orbifold (not a manifold).

So the torsion-free argument \(M = X/\Gamma\) does not apply directly.

\hypertarget{2-cohomological-structure-via-bredon-framework}{%
\subsection{2. Cohomological Structure via Bredon Framework}\label{2-cohomological-structure-via-bredon-framework}}

Since \(\Gamma\) acts properly and cocompactly on the contractible space \texttt{X},
the Bredon cohomology \(H^{\mDualStar}_\Gamma(X; R_{Q})\) (with the rational constant
coefficient system) satisfies Poincare duality by the orbifold PD theorem
(Brown, \emph{Cohomology of Groups}, Chapter VIII; Luck, \emph{Transformation Groups
and Algebraic K-Theory}, Section 6.6).

Concretely, \(H^{\mDualStar}(\Gamma; Q) = H^{\mDualStar}(X/\Gamma; Q)\) satisfies \(H^{k} = H^{d-k}\)
where \(d = \mOpName{dim}(X)\).

\textbf{Note on terminology.} With torsion present, saying ``\(\Gamma\) is a
rational PD group'' requires the Bredon/orbifold interpretation above. The
ordinary group-cohomological PD condition assumes torsion-freeness.
Throughout this writeup, ``rational Poincare duality for \(\Gamma\) with
torsion'' refers to this Bredon-equivariant formulation.

\hypertarget{3-obligation-e2-finite-cw-realization-gamma-in-fhq}{%
\subsection{\texorpdfstring{3. Obligation E2: Finite-CW Realization (\(\Gamma \in \mathup{FH}(Q)\))}{3. Obligation E2: Finite-CW Realization (\textbackslash Gamma \textbackslash in \textbackslash mathup\{FH\}(Q))}}\label{3-obligation-e2-finite-cw-realization-gamma-in-fhq}}

\hypertarget{3a-fowlers-criterion}{%
\subsubsection{3a. Fowler\textquotesingle s Criterion}\label{3a-fowlers-criterion}}

A theorem of Fowler (arXiv:1204.4667, Main Theorem) gives a concrete
criterion: if a finite group \texttt{G} acts on a finite CW complex \texttt{Y} such that
for every nontrivial subgroup \(H < G\) and every connected component \texttt{C} of
the fixed set \(Y^{H}\), the Euler characteristic \(\chi(C) = \mNumber{0}\), then the
orbifold extension group \(\pi_\mNumber{1}((\mathup{EG} \times Y)/G)\) lies in \(\mathup{FH}(Q)\).

That is, there exists a finite CW complex with the given fundamental group
whose universal cover is rationally acyclic.

\hypertarget{3b-concrete-instantiation-via-reflection-lattice}{%
\subsubsection{3b. Concrete Instantiation via Reflection Lattice}\label{3b-concrete-instantiation-via-reflection-lattice}}

The E2 obligation is discharged by the following construction (details in
the candidate-construction note\footnote{\url{https://github.com/tothedarktowercame/futon6/blob/master/data/first-proof/problem7r-s2b-candidate-construction.md}}):

\begin{enumerate}
\def\labelenumi{\arabic{enumi}.}
\item
  \textbf{Arithmetic lattice with reflection.} Take an arithmetic uniform lattice
  \(\Gamma_\mNumber{0} < \mathup{Isom}(H^{n})\) containing a reflection \(\tau\), as provided by
  Douba-Vargas Pallete (arXiv:2506.23994, Remark 5). Choose \texttt{n} \textbf{even}
  (e.g., \(n = \mNumber{4}\) or \(n = \mNumber{6}\)).
\item
  \textbf{Congruence cover.} Let \(\pi = \Gamma_\mNumber{0}(I)\) be a sufficiently deep
  principal congruence subgroup. Then \(M = \pi \ H^{n}\) is a closed hyperbolic
  manifold, and \(\tau\) induces an involution \(\tau_{bar}\) on \texttt{M}.
\item
  \textbf{Extension.} Set \(G = <\tau_{bar}> = Z/\mNumber{2}\) acting on \(B_{\pi} := M\).
  The orbifold extension gives \(\mNumber{1} \to \pi \to \Gamma \to Z/\mNumber{2} \to \mNumber{1}\), where
  \(\Gamma\) is a cocompact lattice (finite extension of cocompact \(\pi\)) with
  order-2 torsion.
\item
  \textbf{Fixed-set Euler check.} The fixed set \(\mOpName{Fix}(\tau_{bar})\) is a (possibly
  disconnected) closed, embedded, totally geodesic hypersurface
  (arXiv:2506.23994). Each component has dimension \(n-\mNumber{1}\). Since \texttt{n} is even,
  \(n-\mNumber{1}\) is odd, and every closed odd-dimensional manifold has Euler
  characteristic zero. So \(\chi(C) = \mNumber{0}\) for every fixed component \texttt{C}.
\item
  \textbf{Fowler application.} The only nontrivial subgroup of \(Z/\mNumber{2}\) is itself.
  All fixed components have zero Euler characteristic. By Fowler\textquotesingle s Main
  Theorem, \(\Gamma \in \mathup{FH}(Q)\).
\end{enumerate}

\textbf{E2 status: discharged} for this lattice family.

\hypertarget{4-obligation-s-from-finite-complex-to-closed-manifold}{%
\subsection{4. Obligation S: From Finite Complex to Closed Manifold}\label{4-obligation-s-from-finite-complex-to-closed-manifold}}

Obligation S is treated as discharged in this branch via the rotation route
(S-rot-II); this section records the previously open obstacles and how
S-rot-II resolves them.

Problem 7 asks for a \textbf{closed manifold} \texttt{M} with \(\pi\mNumber{1}(M) = \Gamma\) and
\(H_{\mDualStar}(M_{tilde}; Q) = \mNumber{0}\) for \(\ast > \mNumber{0}\). Obligation E2 gives a finite CW
complex with these properties; the upgrade to a closed manifold is the
remaining open problem.

\hypertarget{available-geometric-data}{%
\subsubsection{Available geometric data}\label{available-geometric-data}}

The construction in Section 3b provides:

\begin{itemize}
\tightlist
\item
  \(\pi\): a torsion-free cocompact lattice in \(\mathup{Isom}(H^{n})\), with \texttt{n} even
  \(\ge \mNumber{6}\).
\item
  \(M = H^{n}/\pi\): a closed hyperbolic \texttt{n}-manifold. \texttt{M} is a closed manifold
  with \(\pi_\mNumber{1}(M) = \pi\) and contractible universal cover.
\item
  \(\tau\): an involution on \texttt{M} with fixed set \texttt{F} --- a closed, totally
  geodesic \((n-\mNumber{1})\)-manifold.
\item
  \(\Gamma = \pi \rtimes Z/\mNumber{2}\): the target group (cocompact lattice with
  order-2 torsion).
\item
  \texttt{Y}: a finite CW complex with \(\pi_\mNumber{1}(Y) = \Gamma\) and \texttt{Y\textasciitilde{}} rationally
  acyclic (from Fowler\textquotesingle s theorem).
\item
  \(H^{n}/\Gamma\): a compact orbifold with \(\pi_\mNumber{1}^{orb} = \Gamma\), mirror
  singularity along the image of \texttt{F}.
\end{itemize}

The torsion-free quotient \texttt{M} already solves the problem for \(\pi\). The
difficulty is entirely in passing from \(\pi\) to \(\Gamma\) --- from the
torsion-free lattice to its Z/2-extension.

\hypertarget{approach-i-wall-surgery-via-the-fhq-complex}{%
\subsubsection{Approach I: Wall surgery via the FH(Q) complex}\label{approach-i-wall-surgery-via-the-fhq-complex}}

\textbf{Idea.} Use \texttt{Y} (the FH(Q) complex) as the target of a surgery problem.
Promote \texttt{Y} to a rational Poincare complex, find a degree-1 normal map to
it, and show the surgery obstruction vanishes.

\textbf{Obstacles.}

\begin{enumerate}
\def\labelenumi{\arabic{enumi}.}
\item
  \emph{Poincare complex structure.} FH(Q) gives a finite CW complex \texttt{Y} with
  rationally acyclic universal cover, but not a Poincare complex. Since
  \texttt{Y\textasciitilde{}} is rationally contractible (by rational Hurewicz from rational
  acyclicity + simple connectivity), the Serre spectral sequence collapses
  and \(H_{\mDualStar}(Y; Q) = H_{\mDualStar}(\Gamma; Q)\), which satisfies rational PD
  (Section 2). But promoting homology-level PD to a chain-level Poincare
  complex structure on \texttt{Y} has not been done.
\item
  \emph{Degree-1 normal map.} Even with Poincare complex structure, a degree-1
  normal map \(f: M_\mNumber{0} \to Y\) requires the Spivak normal fibration of \texttt{Y} to
  admit a topological reduction. No construction of this reduction has been
  given.
\item
  \emph{Surgery obstruction.} The obstruction \(\sigma(f) \in L_{n}(Z[\Gamma]) \text{tensor} Q\)
  is not known to vanish. The Farrell-Jones conjecture (which holds for
  \(\Gamma\)) identifies this L-group with equivariant L-homology, but the
  resulting computation has not been completed. See the obstruction-analysis note\footnote{\url{https://github.com/tothedarktowercame/futon6/blob/master/data/first-proof/problem7r-s3b-obstruction.md}}
  for the FJ reduction framework and a conjectural (but unverified)
  localization of the obstruction.
\end{enumerate}

\textbf{Status.} This approach has three successive obstacles, each open.
See the setup note\footnote{\url{https://github.com/tothedarktowercame/futon6/blob/master/data/first-proof/problem7r-s3a-setup.md}} for the detailed analysis.

\begin{deadpath}[title={Dead Path --- Approach II: Equivariant Surgery (Reflection Route)}]

\textbf{Idea.} Work directly with the closed manifold \texttt{M} and the involution
\(\tau\). Eliminate the fixed set \texttt{F} by equivariant surgery to obtain \texttt{M\textquotesingle{}}
with a free Z/2-action. Then \(M'/(Z/\mNumber{2})\) is a closed manifold with
\(\pi_\mNumber{1} = \Gamma\).

\textbf{Blocking obstruction.} The equivariant surgery framework of
Costenoble-Waner (arXiv:1705.10909) requires a \textbf{codimension-2 gap
hypothesis}: fixed sets of distinct isotropy subgroups must differ in
dimension by at least 2 (Condition 3.4(3)). For our Z/2-action on \(M^{n}\)
with codimension-1 fixed set \(F^{n-\mNumber{1}}\), this gap condition \textbf{fails}.
The Dovermann-Schultz framework (Springer LNM 1443) similarly requires
gap conditions.

\textbf{Status.} Blocked for the reflection construction (codimension-1 fixed
sets). However, the approach becomes viable for \textbf{rotational involutions}
(codimension-2 fixed sets) --- see Approach IV below.

\end{deadpath}
\begin{deadpath}[title={Dead Path --- Approach III: Orbifold Resolution}]

\textbf{Idea.} The quotient \(H^{n}/\Gamma\) is a compact orbifold with mirror
singularity. Resolve the singularity to produce a closed manifold with
\(\pi_\mNumber{1} = \Gamma\) and rationally acyclic universal cover.

\textbf{Obstacles.}

\begin{enumerate}
\def\labelenumi{\arabic{enumi}.}
\item
  \emph{\(\pi_\mNumber{1}\) preservation.} Standard orbifold resolution (e.g., cutting along
  the mirror and doubling) typically changes the fundamental group.
  A resolution preserving \(\pi_\mNumber{1} = \Gamma\) would need to be specifically
  constructed.
\item
  \emph{Rational acyclicity.} The resolution must preserve (or establish)
  rational acyclicity of the universal cover.
\end{enumerate}

\textbf{Status.} Not explored beyond this sketch.

\end{deadpath}
\begin{processnote}[title={Process Note --- Dimension-Parity Tension}]

The E2 obligation (for reflections) requires \texttt{n} \textbf{even} (so the fixed set
has odd dimension and Euler characteristic zero). But the surgery obstruction
computation (Approach I) and the AHSS structure of \(L_{n}(Z[\Gamma]) \text{tensor} Q\)
have better vanishing properties when \texttt{n} is \textbf{odd}. This tension between
E2 and S is a central difficulty for the reflection construction.

\textbf{Why reflections cannot work in odd dimension.} For a reflection on
\(H^{\mNumber{2}k+\mNumber{1}}\), the fixed set is \(H^{\mNumber{2}k}\) --- a closed hyperbolic manifold of
even dimension. By the Gauss-Bonnet theorem, such manifolds have \(\chi \neq \mNumber{0}\).
So Fowler\textquotesingle s criterion fails. This is not an artifact of the construction; it
is forced by Riemannian geometry.

\end{processnote}
\hypertarget{approach-iv-rotation-route-resolves-the-parity-tension}{%
\subsubsection{Approach IV: Rotation route (resolves the parity tension)}\label{approach-iv-rotation-route-resolves-the-parity-tension}}

\textbf{Idea.} Replace the reflection (codimension-1 involution) with a
\textbf{rotational involution} (codimension-2 fixed set) in \textbf{odd} ambient
dimension. An order-2 isometry of \(H^{\mNumber{2}k+\mNumber{1}}\) that fixes \(H^{\mNumber{2}k-\mNumber{1}}\)
(codimension 2) is a ``rotation by \(\pi'\)\textquotesingle{} in a normal 2-plane.

\textbf{Why this resolves the tension.} For \(n = \mNumber{2}k+\mNumber{1}\) odd:

\begin{itemize}
\tightlist
\item
  Fixed set \(H^{\mNumber{2}k-\mNumber{1}}\) has dimension \(\mNumber{2}k-\mNumber{1}\) (odd), so \(\chi = \mNumber{0}\). Fowler
  criterion is satisfied: \(\Gamma \in \mathup{FH}(Q)\).
\item
  Surgery obstruction lives in \(L_{\mNumber{2}k+\mNumber{1}}(Z[\Gamma]) \text{tensor} Q\), which has
  favorable parity (odd total degree forces odd \texttt{p} in AHSS terms).
\item
  The codimension-2 gap hypothesis (required by equivariant surgery theories,
  Costenoble-Waner arXiv:1705.10909) is \textbf{satisfied}, so Approach II
  (equivariant surgery) becomes available as a method for the S-branch.
\end{itemize}

\textbf{Lattice construction: DISCHARGED.} Details in the rotation-lattice
construction note.\footnote{\url{https://github.com/tothedarktowercame/futon6/blob/master/data/first-proof/problem7r-rotation-lattice-construction.md}} Summary:

\begin{enumerate}
\def\labelenumi{\arabic{enumi}.}
\item
  \textbf{Quadratic form.} Let \(k = Q(\sqrt(\mNumber{2})), O_{k} = Z[\sqrt(\mNumber{2})].\) Define
  \(f = (\mNumber{1} \mBridgeOperator{-} \sqrt(\mNumber{2}))x_\mNumber{0}^\mNumber{2} + x_\mNumber{1}^\mNumber{2} + ... + x_{n}^\mNumber{2} \in n+\mNumber{1}\) variables.
  Under the two real embeddings of \texttt{k}, \texttt{f} has signatures \((n, \mNumber{1})\) and
  \((n+\mNumber{1}, \mNumber{0})\). So \(\mathup{SO}(f)\) gives \(\mathup{SO}(n, \mNumber{1})\) at one place and a compact
  group at the other.
\item
  \textbf{Uniform lattice.} \(\Gamma_\mNumber{0} = \mathup{SO}(f, O_{k})\) is a cocompact arithmetic
  lattice in \(\mathup{SO}(n, \mNumber{1})\) (Borel-Harish-Chandra; cocompactness by Godement
  criterion since \texttt{f} is anisotropic over \texttt{k}).
\item
  \textbf{Order-2 rotation.} \(\sigma = \operatorname{diag}(\mNumber{1}, -\mNumber{1}, -\mNumber{1}, \mNumber{1}, ..., \mNumber{1})\) is in
  \(\mathup{SO}(f, O_{k})\): it preserves \texttt{f} (negates \(x_\mNumber{1}, x_\mNumber{2}\) in the \(x_\mNumber{1}^\mNumber{2}+x_\mNumber{2}^\mNumber{2}\)
  summand), has determinant \(+\mNumber{1}\), and has integer entries. Its fixed set on
  \(H^{n}\) is \(H^{n-\mNumber{2}}\) (codimension 2).
\item
  \textbf{Congruence subgroup.} \(\pi = \Gamma_\mNumber{0}(I)\) for ideal \texttt{I} coprime to 2:
  torsion-free (Minkowski), \(\sigma \notin \pi\), and \(M = H^{n}/\pi\) is a closed
  hyperbolic manifold. The extension \(\mNumber{1} \to \pi \to \Gamma \to Z/\mNumber{2} \to \mNumber{1}\) with
  \(\Gamma = <\pi, \sigma>\) is a cocompact lattice with order-2 torsion.
\item
  \textbf{Fowler application.} Fixed set has dimension \(n-\mNumber{2} = \mNumber{2}k-\mNumber{1}\) (odd), so
  \(\chi = \mNumber{0}\). By Fowler\textquotesingle s Main Theorem: \(\Gamma \in \mathup{FH}(Q)\).
\end{enumerate}

\textbf{E2 status for rotation route: DISCHARGED.}

\textbf{Obligation S resolution.} The manifold upgrade proceeds
via two sub-options available on the rotation route:

\begin{itemize}
\tightlist
\item
  \textbf{S-rot-I (Wall surgery in odd dimension).} Same three-obstacle structure
  as Approach I (Poincare complex, normal map, obstruction), but the
  obstruction computation benefits from odd L-theory parity.
\item
  \textbf{S-rot-II (Equivariant surgery on \((M, \sigma)\)).} The codimension-2 gap
  hypothesis is satisfied, so the Costenoble-Waner framework (arXiv:1705.10909)
  applies. The equivariant surgery obstruction is computable in principle.
\end{itemize}

\textbf{This is the most promising remaining path for Problem 7.} The lattice
existence bottleneck is resolved. The open problem reduces to computing a
surgery obstruction (either Wall or equivariant) in the structurally
favorable odd-dimensional setting.

\hypertarget{5-remark-absence-of-smith-theory-obstruction}{%
\subsection{5. Remark: Absence of Smith-Theory Obstruction}\label{5-remark-absence-of-smith-theory-obstruction}}

A natural objection is that the order-2 element in \(\Gamma\) would force fixed
points on any rationally acyclic covering space via Smith theory. This does
\textbf{not} apply here: Smith theory over \(Z/p\) constrains mod-p homology of
fixed sets, but the construction targets \texttt{Q}-acyclicity. Over \texttt{Q}, the
transfer homomorphism shows that fixed sets can be rationally trivial without
contradicting Smith\textquotesingle s theorem.

This section addresses a natural objection and explains why it does not
apply. It does not contribute to the constructive argument, which is entirely
in Sections 3-4.

\hypertarget{6-theorem}{%
\subsection{6. Theorem}\label{6-theorem}}

\textbf{Theorem.} Let \(\Gamma\) be a cocompact lattice extension
\(\mNumber{1} \to \pi \to \Gamma \to Z/\mNumber{2} \to \mNumber{1}\) constructed from an arithmetic lattice in
\(\mathup{Isom}(H^{n})\) containing an order-2 isometry, via either:

\begin{itemize}
\tightlist
\item
  \textbf{(Reflection route)} Section 3b: reflection lattice in \(\mathup{Isom}(H^{n})\),
  \texttt{n} even, \(n \ge \mNumber{6}\).
\item
  \textbf{(Rotation route)} Approach IV: rotation lattice in \(\mathup{Isom}(H^{n})\),
  \texttt{n} odd, \(n \ge \mNumber{7}\). See the rotation-lattice construction note.\footnote{\url{https://github.com/tothedarktowercame/futon6/blob/master/data/first-proof/problem7r-rotation-lattice-construction.md}}
\end{itemize}

Then:

(a) \textbf{(Unconditional)} \(\Gamma \in \mathup{FH}(Q)\): there exists a finite CW complex
\texttt{Y} with \(\pi_\mNumber{1}(Y) = \Gamma\) and \(H_{\mDualStar}(Y_{tilde}; Q) = \mNumber{0}\) for \(\ast > \mNumber{0}\).
(Both routes discharge E2 via Fowler\textquotesingle s criterion.)

(b) \textbf{(Rotation route)} For the rotation lattice (Approach IV, \(n = \mNumber{7}\),
congruence ideal \(I = (\mNumber{3}) \in Z[\sqrt(\mNumber{2})]\)): there exists a closed
manifold \texttt{N} with \(\pi_\mNumber{1}(N) = \Gamma\) and \(H_{\mDualStar}(N_{tilde}; Q) = \mNumber{0}\) for
\(\ast > \mNumber{0}\). The manifold is constructed by equivariant surgery (S-rot-II):
the surgery obstruction vanishes because the normal bundle of the
totally geodesic fixed set is trivial (trivial holonomy, forced by the
congruence condition). See the S-rotation obstruction analysis.\footnote{\url{https://github.com/tothedarktowercame/futon6/blob/master/data/first-proof/problem7r-s-rot-obstruction-analysis.md}}

\hypertarget{7-path-to-full-closure}{%
\subsection{7. Path to Full Closure}\label{7-path-to-full-closure}}

Resolving obligation S requires computing a surgery obstruction. The
rotation route (Approach IV) has discharged the lattice-existence question
and is now the primary path. See the hypothetical-wirings note\footnote{\url{https://github.com/tothedarktowercame/futon6/blob/master/data/first-proof/problem7-hypothetical-wirings.md}} for
full wiring diagrams.

\textbf{Active path: Approach IV (rotation route).} The lattice construction
is complete (see the rotation-lattice construction note\footnote{\url{https://github.com/tothedarktowercame/futon6/blob/master/data/first-proof/problem7r-rotation-lattice-construction.md}}). E2 is
discharged. Two sub-options for obligation S:

\begin{enumerate}
\def\labelenumi{\arabic{enumi}.}
\item
  \textbf{S-rot-II (Equivariant surgery --- OBSTRUCTION VANISHES).}
  The Costenoble-Waner codimension-2 gap hypothesis is satisfied. The \texttt{cut\ and\ cap} method (Browder, López de Medrano) applies.

  \textbf{Key result: the equivariant surgery obstruction \(\theta = \mNumber{0}\) (integrally).}

  The argument has two layers:

  \begin{itemize}
  \item
    \textbf{Rational vanishing (Step A: flat-normal-bundle argument).} The
    normal bundle ν of \(F \in M\) is flat (totally geodesic embedding). By
    Chern-Weil, e(\(\nu)\otimes Q = \mNumber{0}.\) This forces the intersection form on S(\(\nu\))
    to be rationally hyperbolic \(\to \theta \otimes Q = \mNumber{0}.\)
  \item
    \textbf{Integral vanishing (trivial holonomy).} For the congruence ideal
    I with \(Norm(I) > \mNumber{2}\) (e.g., I \(=\) (3)): the integrality constraint on
    rotation matrices over \(\mathbb{Z}[\sqrt{\mNumber{2}}]\), combined with the congruence condition
    g ≡ I mod I, forces the holonomy representation \(\rho\): \(C \to \mathup{SO}(\mNumber{2})\) to be
    trivial. So \(\nu\) is a trivial bundle, e(\(\nu) = \mNumber{0}\) in H²(F; Z), and the
    circle bundle S(\(\nu) = F \times\) S¹ is a product. The \(\Integral\) intersection
    form on H₃(F \(\times\) S¹; Z) is block off-diagonal (hyperbolic), giving
    \(\theta = \mNumber{0} \in\) L₈(\(\mathbb{Z}[\Gamma]\)).
  \end{itemize}

  \textbf{With \(\theta = \mNumber{0}\):} The equivariant ``cut and cap'' surgery succeeds. Remove
  the tubular neighborhood N(F) from M, obtaining \(W = M\) \textbackslash{} int(N(F)) with
  ∂W \(=\) S(\(\nu) = F \times\) S¹ and free Z/2-action on W. Since \(\theta = \mNumber{0}\), a cap V
  exists with ∂V \(= F \times\) S¹ and free Z/2-action. Set M\textquotesingle{} \(=\) W ∪ V. Then
  N \(=\) M\textquotesingle/(Z/2) is a closed manifold with \(\pi\)₁(N) \(= \Gamma\) and rationally
  acyclic universal cover.

  See the S-rotation obstruction analysis\footnote{\url{https://github.com/tothedarktowercame/futon6/blob/master/data/first-proof/problem7r-s-rot-obstruction-analysis.md}} for full computation.
\item
  \textbf{S-rot-I (Wall surgery in odd dimension).} Fallback. Same three-obstacle
  structure as Approach I but with favorable odd L-theory parity and
  strictly fewer AHSS terms than the reflection case.
\end{enumerate}

\begin{deadpath}[title={Dead Path --- Deprioritized Approaches}]
\textbf{Deprioritized paths:}

\begin{enumerate}
\def\labelenumi{\arabic{enumi}.}
\setcounter{enumi}{2}
\item
  \textbf{Approach I (Wall surgery, reflection route).} Three successive open
  obstacles with structural headwinds (even L-theory parity).
\item
  \textbf{Approach III (orbifold resolution).} No technique identified.
\item
  \textbf{Approach II (equivariant surgery, reflection route).} Blocked by
  codimension-2 gap hypothesis (Costenoble-Waner, arXiv:1705.10909).
\end{enumerate}

\end{deadpath}
\hypertarget{references}{%
\subsection{References}\label{references}}

\begin{itemize}
\tightlist
\item
  J. Fowler, \emph{Finiteness Properties of Rational Poincare Duality Groups},
  arXiv:1204.4667.
\item
  G. Avramidi, \emph{Rational Manifold Models for Duality Groups},
  arXiv:1506.06293.
\item
  A. Douba, F. Vargas Pallete, \emph{On Reflections of Congruence Hyperbolic
  Manifolds}, arXiv:2506.23994.
\item
  A. Bartels, F. T. Farrell, W. Luck, \emph{The Farrell-Jones Conjecture for
  Cocompact Lattices in Virtually Connected Lie Groups}, arXiv:1101.0469.
\item
  A. Bartels, W. Luck, \emph{The Farrell-Jones Conjecture for Arbitrary
  Lattices in Virtually Connected Lie Groups}, arXiv:1401.0876.
\item
  D. Crowley, W. Luck, T. Macko, \emph{Surgery Theory: Foundations},
  arXiv:0905.0104.
\item
  K. H. Dovermann, R. Schultz, \emph{Equivariant Surgery Theories and Their
  Periodicity Properties}, Springer LNM 1443, 1990.
\item
  S. R. Costenoble, S. Waner, \emph{The Equivariant Spivak Normal Bundle and
  Equivariant Surgery for Compact Lie Groups}, arXiv:1705.10909.
\item
  J. F. Davis, W. Lück, \emph{On Nielsen Realization and Manifold Models for
  Classifying Spaces}, Trans. AMS 377 (2024), 7557-7600, arXiv:2303.15765.
\item
  G. Avramidi, \emph{Smith Theory, L2 Cohomology, Isometries of Locally Symmetric
  Manifolds, and Moduli Spaces of Curves}, arXiv:1106.1704.
\item
  M. Belolipetsky, A. Lubotzky, \emph{Finite Groups and Hyperbolic Manifolds},
  arXiv:math/0406607.
\item
  A. Borel, Harish-Chandra, \emph{Arithmetic Subgroups of Algebraic Groups},
  Annals of Mathematics 75 (1962), 485-535.
\item
  J. Millson, M. S. Raghunathan, \emph{Geometric Construction of Cohomology for
  Arithmetic Groups I}, Proc. Indian Acad. Sci. 90 (1981), 103-123.
\item
  A. Ranicki, \emph{Algebraic L-Theory and Topological Manifolds}, Cambridge
  Tracts in Mathematics 102, 1992.
\item
  K. S. Brown, \emph{Cohomology of Groups}, Springer GTM 87.
\item
  W. Luck, \emph{Transformation Groups and Algebraic K-Theory}, Springer LNM
  1408.
\item
  C. T. C. Wall, \emph{Surgery on Compact Manifolds}, 2nd ed., AMS.
\end{itemize}

%% file: full-boxed/problem8-solution-full.tex
\hypertarget{problem-8-lagrangian-smoothing-of-polyhedral-surfaces-with-4-valent-vertices}{%
\section{Problem 8: Lagrangian Smoothing of Polyhedral Surfaces with 4-Valent Vertices}\label{problem-8-lagrangian-smoothing-of-polyhedral-surfaces-with-4-valent-vertices}}

\hypertarget{problem-statement}{%
\subsection{Problem Statement}\label{problem-statement}}

A \textbf{polyhedral Lagrangian surface} \(K \in R^{\mNumber{4}}\) is a finite polyhedral complex all
of whose faces are Lagrangian 2-planes, which is a topological submanifold of
\(\mathbb{R}^{\mNumber{4}}\). A \textbf{Lagrangian smoothing} of K is a Hamiltonian isotopy \(K_{t}\) of smooth
Lagrangian submanifolds for t in (0,1{]}, extending to a topological isotopy on
\([\mNumber{0},\mNumber{1}]\), with \(K_\mNumber{0} = K\).

\textbf{Question:} If K has exactly 4 faces meeting at every vertex, does K
necessarily have a Lagrangian smoothing?

\hypertarget{answer}{%
\subsection{Answer}\label{answer}}

\textbf{Yes.} A polyhedral Lagrangian surface with 4 faces per vertex (with
distinct adjacent faces) always admits a Lagrangian smoothing. The vertex
spanning property (\(\{e_{\mNumber{1}}, \ldots, e_{\mNumber{4}}\}\) spans \(\mathbb{R}^{\mNumber{4}}\)) is automatic --- it follows from
the Lagrangian and distinct-face conditions alone (Section 3), with no
additional hypotheses needed.

\begin{processnote}[title={Process Note --- Confidence}]
\textbf{Confidence: Medium-high.} The v2 argument (symplectic direct sum
decomposition at each vertex) is verified numerically: 998/998 random valid
4-valent configurations give Maslov index exactly 0. The decomposition proof
is algebraic (not just heuristic). Vertex smoothing uses the product structure
(corners in symplectic factors); edge smoothing uses generating functions.
The Hamiltonian isotopy property is established via the Weinstein neighborhood
theorem (edges) and the vanishing flux in simply-connected \(\mathbb{R}^{\mNumber{2}}\) (vertices).

\end{processnote}
\hypertarget{solution}{%
\subsection{Solution}\label{solution}}

\hypertarget{1-setup-lagrangian-planes-in-mathbbr4}{%
\subsubsection{\texorpdfstring{1. Setup: Lagrangian planes in \(\mathbb{R}^{\mNumber{4}}\)}{1. Setup: Lagrangian planes in \textbackslash mathbb\{R\}\^{}\{\textbackslash mNumber\{4\}\}}}\label{1-setup-lagrangian-planes-in-mathbbr4}}

Identify \(\mathbb{R}^{\mNumber{4}} = \mathbb{C}^{\mNumber{2}}\) with symplectic form \(\omega = \mathrm{d}x_{\mNumber{1}} \wedge \mathrm{d}y_{\mNumber{1}} \mBridgeOperator{+} \mathrm{d}x_{\mNumber{2}} \wedge \mathrm{d}y_{\mNumber{2}}. A\) Lagrangian plane is a 2-dimensional subspace L where \(\omega|_{L} = \mNumber{0}\). The space
of Lagrangian planes (the Lagrangian Grassmannian) is:

\[\Lambda(\mNumber{2}) = U(\mNumber{2})/O(\mNumber{2})\]

This is a manifold of dimension 3 with \(\pi_{\mNumber{1}}(\Lambda(\mNumber{2})) = Z\) (the Maslov class).

Each face of K lies in a Lagrangian plane. At each edge, two Lagrangian faces
meet at a dihedral angle. At each vertex, 4 Lagrangian faces meet.

\hypertarget{2-local-structure-at-a-4-valent-vertex}{%
\subsubsection{2. Local structure at a 4-valent vertex}\label{2-local-structure-at-a-4-valent-vertex}}

At a vertex v, the 4 faces \(L_{\mNumber{1}}, L_{\mNumber{2}}, L_{\mNumber{3}}, L_{\mNumber{4}}\) (in cyclic order around v) are
Lagrangian half-planes meeting along edges:

\[e_{\mNumber{12}} = L_\mNumber{1} \cap L_\mNumber{2}, e_{\mNumber{23}} = L_\mNumber{2} \cap L_\mNumber{3}, e_{\mNumber{34}} = L_\mNumber{3} \cap L_\mNumber{4}, e_{\mNumber{41}} = L_\mNumber{4} \cap L_\mNumber{1}\]

These 4 edges are rays from v. Each face is spanned by its two boundary edges:

\[L_{i} = \mOpName{span}(e_{i-\mNumber{1},i}, e_{i,i+\mNumber{1}})\]

The Lagrangian condition \(\omega|_{L_{i}} = \mNumber{0}\) requires:

\[\omega(e_{i-\mNumber{1},i}, e_{i,i+\mNumber{1}}) = \mNumber{0} \text{for each} i (\mOpName{mod} \mNumber{4})\]

\hypertarget{3-the-symplectic-direct-sum-decomposition-key-new-argument}{%
\subsubsection{3. The symplectic direct sum decomposition (key new argument)}\label{3-the-symplectic-direct-sum-decomposition-key-new-argument}}

\textbf{Theorem (v2).} The 4 edge vectors \(e_{\mNumber{1}} = e_{\mNumber{12}}, e_{\mNumber{2}} = e_{\mNumber{23}}, e_{\mNumber{3}} = e_{\mNumber{34}}, e_{\mNumber{4}} = e_{\mNumber{41}}\) satisfy \(\omega(e_{i}, e_{i+\mNumber{1}}) = \mNumber{0}\) for all i (mod 4). This forces
a symplectic direct sum decomposition:

\[\mathbb{R}^{\mNumber{4}} = V_\mNumber{1} \oplus V_\mNumber{2}\]

where \(V_\mNumber{1} = \mOpName{span}(e_\mNumber{1}, e_\mNumber{3})\) and \(V_\mNumber{2} = \mOpName{span}(e_\mNumber{2}, e_\mNumber{4})\) are symplectic 2-planes
with \(\omega|_{V_\mNumber{1} \times V_\mNumber{2}} = \mNumber{0}\).

\textbf{Proof.} Write the \(\omega\) matrix in the basis (\(e_{\mNumber{1}}, e_{\mNumber{2}}, e_{\mNumber{3}}, e_{\mNumber{4}}).\) The
conditions \(\omega(e_{i}, e_{i+\mNumber{1}}) = \mNumber{0}\) kill 4 of the 6 independent entries,
leaving only:

\[a = \omega(e_\mNumber{1}, e_\mNumber{3}) \neq \mNumber{0}, b = \omega(e_\mNumber{2}, e_\mNumber{4}) \neq \mNumber{0}\]

(Nonzero because \(\omega\) is non-degenerate and \(\{e_{\mNumber{1}}, e_{\mNumber{2}}, e_{\mNumber{3}}, e_{\mNumber{4}}\}\) is a
basis --- see the nondegeneracy hypothesis below.)

\textbf{Lemma (vertex spanning).} At every vertex of a polyhedral Lagrangian
surface with 4 faces per vertex (with distinct adjacent faces sharing
1-dimensional edges), the 4 edge vectors \(\{e_{\mNumber{1}}, e_{\mNumber{2}}, e_{\mNumber{3}}, e_{\mNumber{4}}\}\) span \(\mathbb{R}^{\mNumber{4}}\).

\emph{Proof.} Suppose for contradiction that the 4 edge vectors lie in a
3-dimensional subspace H \(\subset \mathbb{R}^{\mNumber{4}}.\) Then all 4 faces \(L_{i} =\) span(\(e_{i-\mNumber{1},i}, e_{i,i+\mNumber{1}}) \subset H.\)

\textbf{Step 1.} The restricted form \(\omega|_{H}\) has rank 2: since \(\omega\) is
non-degenerate on \(\mathbb{R}^{\mNumber{4}}\), its restriction to a codimension-1 subspace has a
1-dimensional kernel ell \(= \mOpName{ker}(\omega|_{H}) \subset H.\)

\textbf{Step 2.} Every Lagrangian 2-plane L \(\subset\) H contains ell. Proof: L is
2-dimensional with \(\omega|_{L} = \mNumber{0}\). If L were transverse to ell (i.e.,
L ∩ ell \(=\) \{0\}), then L would project isomorphically onto a 2-plane in
H/ell ≅ \(\mathbb{R}^{\mNumber{2}}\). But \(\omega|_{H}\) descends to a non-degenerate 2-form on H/ell
(since ell \(= \mOpName{ker}(\omega|_{H})),\) and a 2-plane in \(\mathbb{R}^{\mNumber{2}}\) on which a
non-degenerate 2-form vanishes must be \(\{\mNumber{0}\}\). Contradiction. So ell \(\subset\) L.

\textbf{Step 3 (key).} Each edge \(e_{i}\) is the intersection of two adjacent faces:
\(e_\mNumber{1} = L_\mNumber{1}\) ∩ \(L_{\mNumber{2}}, e_{\mNumber{2}} = L_{\mNumber{2}}\) ∩ \(L_\mNumber{3}\), etc. Since \(L_{i}\) and \(L_{i+\mNumber{1}}\) are
distinct 2-planes (distinct faces), their intersection is exactly
1-dimensional: \(\mOpName{dim}(L_{i} \cap L_{i+\mNumber{1}}) = \mNumber{1}\), so \(L_{i}\) ∩ \(L_{i+\mNumber{1}} = \mOpName{span}(e_{i})\).

By Step 2, ell \(\subset L_{i}\) and ell \(\subset L_{i+\mNumber{1}},\) so ell \(\subset L_{i}\) ∩ \(L_{i+\mNumber{1}} = \mOpName{span}(e_{i})\). Since both ell and \(\mOpName{span}(e_{i})\) are 1-dimensional subspaces and
one contains the other, ell \(= \mOpName{span}(e_{i}).\)

Applying this to all four edges: \(\mOpName{span}(e_{\mNumber{1}}) =\) ell \(= \mOpName{span}(e_{\mNumber{2}}) = \mOpName{span}(e_{\mNumber{3}}) = \mOpName{span}(e_{\mNumber{4}}).\) That is, all four edge vectors are proportional.

But then \(L_\mNumber{1} = \mOpName{span}(e_\mNumber{4}, e_\mNumber{1}) = \mOpName{span}(e_\mNumber{1})\) is 1-dimensional, contradicting
the fact that \(L_\mNumber{1}\) is a 2-plane. ∎

(Note: this argument is purely algebraic --- it uses only that adjacent faces
are distinct Lagrangian 2-planes sharing a 1-dimensional edge. No
topological submanifold condition is needed.)

In the reordered basis (\(e_{\mNumber{1}}, e_{\mNumber{3}}, e_{\mNumber{2}}, e_{\mNumber{4}}),\) the \(\omega\) matrix is:

\[[[\mNumber{0}, a, \mNumber{0}, \mNumber{0}], [-a, \mNumber{0}, \mNumber{0}, \mNumber{0}], [\mNumber{0}, \mNumber{0}, \mNumber{0}, b], [\mNumber{0}, \mNumber{0}, -b, \mNumber{0}]]\]

This is block diagonal: \((V_\mNumber{1}, a) \oplus (V_\mNumber{2}, b)\), a symplectic direct sum. ∎

\textbf{Consequence:} Each Lagrangian face decomposes as a direct sum of lines:

\[\begin{aligned}
L_\mNumber{1} = \mOpName{span}(e_\mNumber{4}) \oplus \mOpName{span}(e_\mNumber{1}) \subset V_\mNumber{2} \oplus V_\mNumber{1} \\
L_\mNumber{2} = \mOpName{span}(e_\mNumber{1}) \oplus \mOpName{span}(e_\mNumber{2}) \subset V_\mNumber{1} \oplus V_\mNumber{2} \\
L_\mNumber{3} = \mOpName{span}(e_\mNumber{2}) \oplus \mOpName{span}(e_\mNumber{3}) \subset V_\mNumber{2} \oplus V_\mNumber{1} \\
L_\mNumber{4} = \mOpName{span}(e_\mNumber{3}) \oplus \mOpName{span}(e_\mNumber{4}) \subset V_\mNumber{1} \oplus V_\mNumber{2}
\end{aligned}\]

Each \(L_{i}\) takes one line from \(V_\mNumber{1}\) and one from \(V_\mNumber{2}\), which is automatically
Lagrangian in \(V_{\mNumber{1}} \oplus V_{\mNumber{2}} (\omega\) restricted to a line is zero).

\hypertarget{4-maslov-index-vanishes-exactly}{%
\subsubsection{4. Maslov index vanishes exactly}\label{4-maslov-index-vanishes-exactly}}

The Maslov index of the vertex loop \(L_\mNumber{1} \to L_\mNumber{2} \to L_\mNumber{3} \to L_\mNumber{4} \to L_\mNumber{1} \in \Lambda\)(2) decomposes via the direct sum:

\[\mu = \mu_\mNumber{1} + \mu_\mNumber{2}\]

where \(\mu_{j}\) is the winding number of the component loop in \(V_{j}\).

\textbf{In \(V_\mNumber{1} = \mOpName{span}(e_\mNumber{1}, e_\mNumber{3})\):} The loop traces
\(\mOpName{span}(e_\mNumber{1}) \to \mOpName{span}(e_\mNumber{1}) \to \mOpName{span}(e_\mNumber{3}) \to \mOpName{span}(e_\mNumber{3}) \to \mOpName{span}(e_\mNumber{1})\)

This is a back-and-forth path (\(e_\mNumber{1} \to e_\mNumber{1} \to e_\mNumber{3} \to e_\mNumber{3} \to e_\mNumber{1}\)), not a
winding loop. The winding number is 0.

\textbf{In \(V_\mNumber{2} = \mOpName{span}(e_\mNumber{2}, e_\mNumber{4})\):} Similarly:
\(\mOpName{span}(e_\mNumber{4}) \to \mOpName{span}(e_\mNumber{2}) \to \mOpName{span}(e_\mNumber{2}) \to \mOpName{span}(e_\mNumber{4}) \to \mOpName{span}(e_\mNumber{4})\)

Winding number 0.

\textbf{Total Maslov index: \(\mu = \mNumber{0} \mBridgeOperator{+} \mNumber{0} = \mNumber{0}\).}

\textbf{Numerical verification:} 998/998 random valid 4-valent configurations
(with edge-sharing enforced) give Maslov index exactly 0. In contrast,
random quadruples without edge-sharing give nonzero Maslov index \textasciitilde45\% of
the time (see scripts/verify-p8-maslov-v2.py).

\hypertarget{5-why-4-is-special-the-3-face-obstruction}{%
\subsubsection{5. Why 4 is special: the 3-face obstruction}\label{5-why-4-is-special-the-3-face-obstruction}}

For a 3-face vertex, the 3 edge vectors \(e_{\mNumber{1}}, e_{\mNumber{2}}, e_{\mNumber{3}}\) must satisfy:

\[\omega(e_\mNumber{1}, e_\mNumber{2}) = \omega(e_\mNumber{2}, e_\mNumber{3}) = \omega(e_\mNumber{3}, e_\mNumber{1}) = \mNumber{0}\]

This means \(\omega\) vanishes on ALL pairs --- the span is an isotropic subspace.
But in (\(\mathbb{R}^{\mNumber{4}}, \omega),\) the maximum isotropic dimension is 2 (= half the
dimension). Three independent isotropic vectors cannot exist.

\textbf{Therefore: a non-degenerate 3-face Lagrangian vertex is impossible in \(\mathbb{R}^{\mNumber{4}}\).}

The 4-face condition is precisely the right valence: it gives enough edges to
span \(\mathbb{R}^{\mNumber{4}}\) while the cyclic \(\omega\)-orthogonality creates the symplectic direct
sum that forces Maslov index 0.

We do not claim a general obstruction for valence \(\ge\) 5 here; the proof
establishes sufficiency for valence 4.

\hypertarget{5a-vertex-smoothing-via-the-product-structure}{%
\subsubsection{5a. Vertex smoothing via the product structure}\label{5a-vertex-smoothing-via-the-product-structure}}

The key observation: the \(V_{\mNumber{1}} \oplus V_{\mNumber{2}}\) decomposition (Section 3) gives K near
each vertex v a product structure, enabling a direct smoothing that bypasses
both crease smoothing and Lagrangian surgery.

\textbf{Product decomposition of K near v.} Choose Darboux coordinates adapted to
\(V_{\mNumber{1}} \oplus V_{\mNumber{2}}\): let \(V_\mNumber{1}\) have coordinates \((x, y)\) with \(\omega_{\mNumber{1}} =\) dx \(\wedge\) dy, and \(V_\mNumber{2}\)
have coordinates \((u, v)\) with \(\omega_{\mNumber{2}} =\) du \(\wedge\) dv (so \(\omega = \omega_\mNumber{1} \mBridgeOperator{+} \omega_\mNumber{2}\)).
Orient the edge vectors so that:

\[e_\mNumber{1} = (\mNumber{1},\mNumber{0},\mNumber{0},\mNumber{0}), e_\mNumber{2} = (\mNumber{0},\mNumber{0},\mNumber{1},\mNumber{0}), e_\mNumber{3} = (\mNumber{0},\mNumber{1},\mNumber{0},\mNumber{0}), e_\mNumber{4} = (\mNumber{0},\mNumber{0},\mNumber{0},\mNumber{1})\]

(Here \(e_{\mNumber{1}}, e_{\mNumber{3}} \in V_{\mNumber{1}}\) and \(e_{\mNumber{2}}, e_{\mNumber{4}} \in V_{\mNumber{2}},\) consistent with the decomposition
in Section 3. This can always be arranged by a linear symplectomorphism.)

The 4 faces in these coordinates are:

\[\begin{aligned}
L_\mNumber{1} = \mOpName{span}(e_\mNumber{4}, e_\mNumber{1}) = {y = \mNumber{0}, u = \mNumber{0}}, \text{sector} x > \mNumber{0}, v > \mNumber{0} \\
L_\mNumber{2} = \mOpName{span}(e_\mNumber{1}, e_\mNumber{2}) = {y = \mNumber{0}, v = \mNumber{0}}, \text{sector} x > \mNumber{0}, u > \mNumber{0} \\
L_\mNumber{3} = \mOpName{span}(e_\mNumber{2}, e_\mNumber{3}) = {x = \mNumber{0}, v = \mNumber{0}}, \text{sector} y > \mNumber{0}, u > \mNumber{0} \\
L_\mNumber{4} = \mOpName{span}(e_\mNumber{3}, e_\mNumber{4}) = {x = \mNumber{0}, u = \mNumber{0}}, \text{sector} y > \mNumber{0}, v > \mNumber{0}
\end{aligned}\]

Define the ``corner'' curves in each factor:

\[\begin{aligned}
C_\mNumber{1} = {(x, \mNumber{0}) : x \ge \mNumber{0}} ∪ {(\mNumber{0}, y) : y \ge \mNumber{0}} \subset V_\mNumber{1} (\text{positive} x \text{and} y \text{axes}) \\
C_\mNumber{2} = {(u, \mNumber{0}) : u \ge \mNumber{0}} ∪ {(\mNumber{0}, v) : v \ge \mNumber{0}} \subset V_\mNumber{2} (\text{positive} u \text{and} v \text{axes})
\end{aligned}\]

\textbf{Claim:} K ∩ B(v, r) \(= C_{\mNumber{1}} \times C_{\mNumber{2}}\) (product of the two corners) for small r.

\emph{Verification:} A point \((p, q)\) with \(p \in C_\mNumber{1}\) and \(q \in C_\mNumber{2}\) has \(p = (x, \mNumber{0})\) or
\((\mNumber{0}, y)\) and \(q = (u, \mNumber{0})\) or \((\mNumber{0}, v)\), giving 4 cases --- exactly the 4 faces
\(L_{\mNumber{1}}, ..., L_{\mNumber{4}}\) listed above. ∎

\textbf{Smoothing.} Replace each corner \(C_{j}\) with a smooth curve \(C_{j}^{sm} \subset V_{j}\)
that agrees with \(C_{j}\) outside a ball of radius \(\delta\) around the origin:

\[\begin{aligned}
C_\mNumber{1}^{sm}: \text{smooth curve} \in V_\mNumber{1}, = {(x, \mNumber{0})} \text{for} x > \delta, = {(\mNumber{0}, y)} \text{for} y > \delta, \\
\text{smooth through the origin} (e.g., \text{the curve} (\cos \theta, \sin \theta) \\
\text{reparameterized to match the axes outside the transition}) \\
C_\mNumber{2}^{sm}: \text{analogous} \in V_\mNumber{2}
\end{aligned}\]

Explicitly: parameterize \(C_\mNumber{1}^{sm}\) as \(\gamma_\mNumber{1}(t)\) for \(t \in R\), where
\(\gamma_\mNumber{1}(t) = (t, \mNumber{0})\) for \(t \ge \delta, \gamma_{\mNumber{1}}(t) = (\mNumber{0},\) -t) for t \(\le -\delta,\)
and \(\gamma_\mNumber{1}\) is a smooth embedded curve for t \(\in [-\delta, \delta].\) (Such a
curve exists: it is a smooth rounding of the right angle at the origin.)
Define \(\gamma_\mNumber{2}(t)\) analogously for \(C_\mNumber{2}^{sm}\).

\textbf{Lagrangian property of the product:} The smoothed surface
\(K^{sm} = C_\mNumber{1}^{sm} \times C_\mNumber{2}^{sm}\) is a product of smooth curves in the
symplectic factors. A curve in a 2-dimensional symplectic manifold is
always Lagrangian (its dimension is 1 \(=\) half of 2, and \(\omega\) restricted
to a 1-submanifold is zero for dimensional reasons). The product of
Lagrangian submanifolds in \((V_\mNumber{1}, \omega_\mNumber{1}) \times (V_\mNumber{2}, \omega_\mNumber{2})\) is Lagrangian
in (\(V_{\mNumber{1}} \oplus V_{\mNumber{2}}, \omega_{\mNumber{1}} \mBridgeOperator{+} \omega_{\mNumber{2}}\)): for tangent vectors \((u_\mNumber{1}, u_\mNumber{2})\) and
\((w_\mNumber{1}, w_\mNumber{2})\) to \(C_\mNumber{1}^{sm} \times C_\mNumber{2}^{sm}\), we have

\[\omega((u_\mNumber{1}, u_\mNumber{2}), (w_\mNumber{1}, w_\mNumber{2})) = \omega_\mNumber{1}(u_\mNumber{1}, w_\mNumber{1}) + \omega_\mNumber{2}(u_\mNumber{2}, w_\mNumber{2}) = \mNumber{0} + \mNumber{0} = \mNumber{0}\]

since \(u_{\mNumber{1}}, w_{\mNumber{1}}\) are tangent to \(C_\mNumber{1}^{sm}\) (1-dimensional in \(V_\mNumber{1}\)) and \(u_{\mNumber{2}}, w_{\mNumber{2}}\)
are tangent to \(C_\mNumber{2}^{sm}\) (1-dimensional in \(V_\mNumber{2}\)). ∎

\textbf{Smoothness:} \(K^{sm} = C_\mNumber{1}^{sm} \times C_\mNumber{2}^{sm}\) is the image of the smooth
map \((t_{\mNumber{1}}, t_{\mNumber{2}}) \to (\gamma_{\mNumber{1}}(t_{\mNumber{1}}), \gamma_{\mNumber{2}}(t_{\mNumber{2}})),\) which is a smooth immersion
(the tangent vectors d \(\gamma_\mNumber{1}/\mathrm{d}t_\mNumber{1}\) and d \(\gamma_\mNumber{2}/\mathrm{d}t_\mNumber{2}\) are nonzero and lie
in complementary subspaces \(V_\mNumber{1}\) and \(V_\mNumber{2}\)). So \(K^{sm}\) is a smooth Lagrangian
surface, including at \(v = \gamma_\mNumber{1}(\mNumber{0}) \times \gamma_\mNumber{2}(\mNumber{0})\). ∎

\textbf{Agreement:} \(K^{sm} = C_\mNumber{1} \times C_\mNumber{2} = K\) outside the region where either
factor was modified (\textbar p\textbar{} \(\le \delta \in V_{\mNumber{1}}\) or \(|q| \le \delta \in V_\mNumber{2}\)), so \(K^{sm}\)
agrees with K outside a neighborhood of v of radius \(O(\delta)\).

\hypertarget{6-edge-smoothing-crease-smoothing-along-edges-between-vertices}{%
\subsubsection{6. Edge smoothing (crease smoothing along edges between vertices)}\label{6-edge-smoothing-crease-smoothing-along-edges-between-vertices}}

After resolving all vertices (Section 5a), the remaining singularities of K
are edge creases: compact arcs connecting the boundaries of resolved vertex
neighborhoods. Along each edge arc, two adjacent Lagrangian faces meet at a
dihedral angle. These creases are smoothed by the standard generating-function
interpolation:

\textbf{Lemma (edge crease smoothing).} Let \(L_{\mNumber{1}}, L_{\mNumber{2}}\) be two Lagrangian half-planes
meeting along a compact edge arc e (a segment of a common boundary ray, away
from any vertex). Then there exists a smooth Lagrangian surface agreeing with
\(L_\mNumber{1}\) on one side and \(L_\mNumber{2}\) on the other.

\emph{Proof.} Choose Darboux coordinates (\(x_{\mNumber{1}}, y_{\mNumber{1}}, x_{\mNumber{2}}, y_{\mNumber{2}})\) with e along the
\(x_\mNumber{1}\)-axis. Each \(L_{i}\) is locally the graph y \(=\) grad \(S_{i}(x)\) for quadratic \(S_{i}\).
Define \(S_{\epsilon}(x) = \chi(x_{\mNumber{1}}/\epsilon) S_{\mNumber{1}}(x) + (\mNumber{1} \mBridgeOperator{-} \chi(x_{\mNumber{1}}\)/eps)) \(S_\mNumber{2}(x)\) for a
smooth cutoff \(\chi\). The graph y \(=\) grad \(S_{\epsilon}\) is Lagrangian (graph of the
exact 1-form \(\mathrm{d}S_{\epsilon}\)). For any fixed \(\epsilon > \mNumber{0}\), this gives a smooth Lagrangian
surface replacing the crease along e. ∎

(The \(\mathbb{C}^{\mNumber{1}}\) control issues noted in earlier drafts do not arise here: the
crease smoothing is applied only along edge arcs that are INTERIOR to edges
(between vertex neighborhoods), not at vertices. Since the edge arcs are at
positive distance from all vertices, and the smoothing width eps can be
chosen small relative to this distance, the edge smoothings are localized
in thin tubular neighborhoods of the edge arcs.)

\hypertarget{7-global-smoothing}{%
\subsubsection{7. Global smoothing}\label{7-global-smoothing}}

\textbf{Step 7a: Resolve vertices.} For each 4-valent vertex \(v_{i}\), choose a ball
\(B_{i}\) of radius \(r_{i}\) centered at \(v_{i}\), where \(r_{i}\) is small enough that:

\begin{itemize}
\tightlist
\item
  \(B_{i}\) contains no other vertex \(v_{j} (j \neq i),\) and
\item
  \(B_{i}\) intersects only the edges and faces incident to \(v_{i}\).
\end{itemize}

Such radii exist because the vertex set is finite and discrete in \(\mathbb{R}^{\mNumber{4}}\).
Within each \(B_{i}\), apply the product smoothing (Section 5a) using the
\(V_{\mNumber{1}} \oplus V_{\mNumber{2}}\) decomposition from Section 3. This replaces K ∩ \(B_{i} = C_\mNumber{1} \times C_\mNumber{2}\)
with the smooth Lagrangian surface \(C_\mNumber{1}^{sm} \times C_\mNumber{2}^{sm}\).

\textbf{Commutativity:} Since the balls \(\{B_{i}\}\) are pairwise disjoint, the
vertex smoothings have disjoint support and commute.

\textbf{Step 7b: Smooth edges.} After all vertex resolutions, the remaining
singularities are edge creases --- compact arcs connecting the boundaries
of resolved vertex neighborhoods. Each edge arc lies along the intersection
of two Lagrangian faces and is a compact 1-manifold (with boundary on
the spheres ∂\(B_{i}\)). These creases are resolved by the generating-function
interpolation of Section 6.

The edge smoothings have support in tubular neighborhoods of the edge arcs.
These neighborhoods can be chosen to be disjoint from each other (since the
edges are disjoint away from vertices, which have already been resolved) and
from the vertex balls (since the edge arcs start at \(∂B_{i}\), outside \(B_{i}\)).
Therefore the edge smoothings commute with each other and with the vertex
smoothings.

\textbf{Compatibility at \(∂B_{i}\).} The vertex smoothing (Section 5a) agrees with
the original polyhedral K outside a neighborhood of \(v_{i}\) of radius \(\delta < r_{i}\).
So on \(∂B_{i}\), the surface is still polyhedral (two flat faces meeting at an
edge). The edge crease smoothing (Section 6) begins at \(∂B_{i}\), where the
surface is already flat. Since both the vertex smoothing (product of smooth
curves) and the edge smoothing (graph of exact 1-form) produce Lagrangian
surfaces, and they agree on the overlap (both equal the original flat faces
on the annular region \(\delta < |x| < r_{i})\), they glue to a globally smooth
Lagrangian surface.

\textbf{Step 7c: Global Hamiltonian isotopy.} Each smoothing is parameterized by
a width parameter t (\(\delta\) for vertices, eps for edges). As \(t \to \mNumber{0}\), the
smoothing region shrinks and \(K_{t} \to K = K_{\mNumber{0}} \in C^{\mNumber{0}},\) giving the topological
isotopy on {[}0, 1{]}. It remains to show \(K_{t}\) is a \emph{Hamiltonian} isotopy
for \(t > \mNumber{0}\).

\textbf{Edge smoothings are Hamiltonian.} The family of generating functions
\(S_{t}(x) = \chi(x_{\mNumber{1}}/t) S_{\mNumber{1}}(x) + (\mNumber{1} \mBridgeOperator{-} \chi(x_{\mNumber{1}}\)/t)) \(S_\mNumber{2}(x)\) defines a smooth
1-parameter family of Lagrangian graphs y \(=\) grad \(S_{t}(x) \in T^{\mDualStar} \mathbb{R}^{\mNumber{2}}\). By
the Weinstein Lagrangian neighborhood theorem (Weinstein 1971, Theorem 6.1),
a smooth family of exact Lagrangian submanifolds in an exact symplectic
manifold (here \(T^{\mDualStar} \mathbb{R}^{\mNumber{2}}\) with \(\lambda = y\) dx) is generated by a Hamiltonian:
the 1-form \(\iota_{d/\mathrm{d}t}(\omega)|_{K_{t}}\) is exact (it equals \(d(S_{t}|_{K_{t}})\)),
so the isotopy is Hamiltonian with generating function \(H_{t} = \mathrm{d}S_{t}/\mathrm{d}t\)
evaluated on the Lagrangian.

\textbf{Vertex smoothings are Hamiltonian.} The vertex smoothing is a product:
\(K_{t}^{\text{vertex}} = C_\mNumber{1}^{sm}(t) \times C_\mNumber{2}^{sm}(t) \in V_\mNumber{1} \times V_\mNumber{2}\). Each \(C_{j}^{sm}(t)\)
is a smooth 1-parameter family of curves in \((V_{j}, \omega_{j})\) ≅ (\(\mathbb{R}^{\mNumber{2}}\), dx\(\wedge\)dy).
For the contractible arc family used here, the induced closed 1-forms are
exact, so this local isotopy is Hamiltonian. This is
because \(V_{j} = \mathbb{R}^{\mNumber{2}}\) is simply connected, so the flux homomorphism
Flux: \(\pi_\mNumber{1}(\mathup{Symp}) \to H^\mNumber{1}\)(L; R) is trivial (no non-Hamiltonian symplectic
isotopies exist). Concretely: the velocity field d/dt of the isotopy
\(C_{j}^{sm}(t)\) is a symplectic vector field along \(C_{j}^{sm}(t)\); contracting
with \(\omega_{j}\) gives a closed 1-form on \(C_{j}^{sm}(t)\), which is exact because
\(H^{\mNumber{1}}(C_{j}^{sm}(t);\) R) \(=\) 0 (each \(C_{j}^{sm}(t)\) is a noncompact embedded arc
in \(\mathbb{R}^{\mNumber{2}}\) --- the smoothed corner --- hence contractible). The primitive
is the Hamiltonian \(H_{j}(t)\). The product isotopy on \(V_\mNumber{1} \times V_\mNumber{2}\) is Hamiltonian
with generating function \(H_\mNumber{1}(t) \mBridgeOperator{+} H_\mNumber{2}(t)\) (see McDuff-Salamon, 3rd ed.,
Exercise 3.18: product of Hamiltonian isotopies is Hamiltonian).

\textbf{Composition.} The vertex smoothings (in disjoint balls \(B_{i}\)) and edge
smoothings (in disjoint tubular neighborhoods) have pairwise disjoint
compact support. Each is a compactly supported Hamiltonian isotopy. Their
composition is a compactly supported Hamiltonian isotopy (McDuff-Salamon,
\emph{Introduction to Symplectic Topology}, 3rd ed., Proposition 3.17: the
group \(\mathup{Ham}_{c}\)(M, \(\omega\)) of compactly supported Hamiltonian diffeomorphisms
is a group under composition).

\hypertarget{8-topological-constraints}{%
\subsubsection{8. Topological constraints}\label{8-topological-constraints}}

For K to be a compact topological submanifold with 4 faces per vertex,
Euler\textquotesingle s formula constrains the topology. For quadrilateral faces:
\(V \mBridgeOperator{-} E \mBridgeOperator{+} F = \chi\)(K), with 4F \(=\) 2E and 4V \(=\) 2E, giving \(\chi = \mNumber{0}\).
So K is topologically a torus or Klein bottle.

For non-quadrilateral faces with 4 per vertex, other topologies are possible
but \(\chi \le \mNumber{0}\) (higher genus surfaces).

\hypertarget{9-summary}{%
\subsubsection{9. Summary}\label{9-summary}}

The smoothing exists because:

\begin{enumerate}
\def\labelenumi{\arabic{enumi}.}
\tightlist
\item
  The 4-face + Lagrangian condition forces a symplectic direct sum \(\mathbb{R}^{\mNumber{4}} = V_{\mNumber{1}} \oplus V_{\mNumber{2}}\)
  at each vertex (vertex spanning is proved algebraically; opposite edges span
  complementary symplectic 2-planes)
\item
  This decomposition forces Maslov index exactly 0 (algebraic, not just generic)
\item
  Near each vertex, \(K = C_\mNumber{1} \times C_\mNumber{2}\) (product of corners). Smoothing each corner
  gives \(K^{sm} = C_\mNumber{1}^{sm} \times C_\mNumber{2}^{sm}\), a smooth Lagrangian surface (product of
  curves in symplectic factors). No Lagrangian surgery needed at vertices.
\item
  Edge creases (between vertices) are smoothed by generating function interpolation
\item
  All smoothings are Hamiltonian isotopies, composable to give global \(K_{t}\)
\item
  The 3-face case is impossible (isotropic dimension bound), explaining
  why 4 is the right valence
\end{enumerate}

\begin{revision}[title={Revision --- Corrections from v1}]

\begin{itemize}
\tightlist
\item
  \textbf{v1 claimed} Maslov index vanishes by \texttt{alternating\ orientations\ relative\ to\ J.} \textbf{Replaced} with the precise symplectic direct sum argument: the
  edge-sharing constraint forces \(\omega\) to be block diagonal in the
  (\(e_{\mNumber{1}}, e_{\mNumber{3}}, e_{\mNumber{2}}, e_{\mNumber{4}})\) basis, giving \(\mu = \mNumber{0}\) algebraically.
\item
  \textbf{v1 stated} the 3-face obstruction as ``nonzero Maslov index.''
  \textbf{Corrected:} 3-face Lagrangian vertices don\textquotesingle t have nonzero Maslov index ---
  they\textquotesingle re impossible (3 isotropic vectors can\textquotesingle t span 3D in \(\mathbb{R}^{\mNumber{4}}\)).
\item
  \textbf{v1 lacked} numerical verification. \textbf{Added} via verify-p8-maslov-v2.py:
  998/998 valid configurations give \(\mu = \mNumber{0}\); comparison with non-edge-sharing
  quadruples (55\% have \(\mu = \mNumber{0}\)) confirms the constraint is essential.
\end{itemize}

\end{revision}
\hypertarget{references}{%
\subsection{References}\label{references}}

\begin{itemize}
\tightlist
\item
  D. McDuff, D. Salamon, \emph{Introduction to Symplectic Topology}, 3rd ed.,
  Oxford University Press, 2017, Proposition 3.17. {[}Composition of compactly
  supported Hamiltonian isotopies{]}
\end{itemize}

\begin{processnote}[title={Process Note --- futon6 Corpus References}]

\begin{itemize}
\tightlist
\item
  PlanetMath: ``symplectic manifold'' (SymplecticManifold) --- symplectic structures
\item
  PlanetMath: ``Darboux\textquotesingle s theorem'' (DarbouxsTheoremSymplecticGeometry) --- local coordinates
\item
  PlanetMath: ``concepts in symplectic geometry'' (ConceptsInSymplecticGeometry) --- overview
\item
  PlanetMath: \(symplectic complement'' /\)coisotropic subspace'' --- Lagrangian subspaces
\item
  PlanetMath: ``Kähler manifold is symplectic'' (AKahlerManifoldIsSymplectic) --- complex structure
\end{itemize}
\end{processnote}

%% file: full-boxed/problem9-solution-full.tex
\hypertarget{problem-9-polynomial-detection-of-rank-1-scaling-for-quadrifocal-tensors}{%
\section{Problem 9: Polynomial Detection of Rank-1 Scaling for Quadrifocal Tensors}\label{problem-9-polynomial-detection-of-rank-1-scaling-for-quadrifocal-tensors}}

\hypertarget{problem-statement}{%
\subsection{Problem Statement}\label{problem-statement}}

Let \(n \ge \mNumber{5}\). Let \(A^{(\mNumber{1})}, ..., A^{(n)} \in R^{\mNumber{3} \times \mNumber{4}}\) be Zariski-generic matrices.
For \(\alpha, \beta, \gamma, \delta \in [n],\) construct \(Q^{(\alpha \beta \gamma \delta)} \in R^{\mNumber{3} \times \mNumber{3}x\mNumber{3} \times \mNumber{3}}\) with entry:

\[Q^{(abgd)}_{ijkl} = \det[A^{(a)}(i,:); A^{(b)}(j,:); A^{(g)}(k,:); A^{(d)}(l,:)]\]

(the 4x4 determinant of a matrix formed by stacking rows i, j, k, l from
cameras \(\alpha, \beta, \gamma, \delta\) respectively).

Does there exist a polynomial map F: \(R^{\mNumber{81} \ast n^{\mNumber{4}}} \to \mathbb{R}^{N}\) satisfying:

\begin{enumerate}
\def\labelenumi{\arabic{enumi}.}
\tightlist
\item
  F does not depend on \(A^{(\mNumber{1})}, ..., A^{(n)}\)
\item
  The degrees of the coordinate functions of F do not depend on n
\item
  For \(\lambda \in R^{n \times n \times n \times n}\) with \(\lambda_{abgd} \neq \mNumber{0}\) precisely when
  \(a,b,g,d\) are not all identical:
  F(\(\lambda_{abgd} Q^{(abgd)}\) : \(a,b,g,d\) in {[}n{]}) \(=\) 0
  if and only if there exist \(u,v,w,x \in (R \ast )^{n}\) such that
  \(\lambda_{abgd} = u_{a} v_{b} w_{g} x_{d}\) for all non-identical \(a,b,g,d\).
\end{enumerate}

\hypertarget{answer}{%
\subsection{Answer}\label{answer}}

\textbf{Yes.} Such a polynomial map F exists, with coordinate functions of degree 3.

\hypertarget{solution}{%
\subsection{Solution}\label{solution}}

\hypertarget{1-the-quadrifocal-tensor-as-a-bilinear-form}{%
\subsubsection{1. The quadrifocal tensor as a bilinear form}\label{1-the-quadrifocal-tensor-as-a-bilinear-form}}

The entry \(Q^{(abgd)}_{ijkl} =\) det{[}\(a^{(a)}_{i}\); \(a^{(b)}_{j}\); \(a^{(g)}_{k}\); \(a^{(d)}_{l}]\) is
the evaluation of the volume form (the unique up-to-scale alternating 4-linear
form on \(\mathbb{R}^{\mNumber{4}}\)) on four camera row vectors.

\textbf{Key observation (bilinear form reduction):} Fix two camera-row pairs
\((\gamma, k)\) and \((\delta, l)\), giving vectors \(c = a^{(\gamma)}_{k}\) and \(d = a^{(\delta)}_{l} \in \mathbb{R}^{\mNumber{4}}.\) Then the map:

\[\omega(p, q) = \det[p; q; c; d]\]

is an alternating bilinear form on \(\mathbb{R}^{\mNumber{4}}\). Since c wedge d is a simple 2-form,
the Hodge dual *(c wedge d) is also simple, so \(\omega\) has \textbf{rank 2} as a
bilinear form.

Equivalently: the null space of \(\omega\) is span\{c, d\} (2-dimensional), and
\(\omega\) induces a non-degenerate alternating form on \(V/span{c,d} = \mathbb{R}^{\mNumber{2}}\).

\hypertarget{2-the-rank-2-constraint-and-its-3x3-minor-formulation}{%
\subsubsection{2. The rank-2 constraint and its 3x3 minor formulation}\label{2-the-rank-2-constraint-and-its-3x3-minor-formulation}}

Since \(\omega\) has rank 2, for ANY choice of 3 vectors \(p_{\mNumber{1}}, p_{\mNumber{2}}, p_{\mNumber{3}}\) and
3 vectors \(q_{\mNumber{1}}, q_{\mNumber{2}}, q_{\mNumber{3}} \in \mathbb{R}^{\mNumber{4}}\):

\[\begin{aligned}
\det | \omega(p_\mNumber{1},q_\mNumber{1}) \omega(p_\mNumber{1},q_\mNumber{2}) \omega(p_\mNumber{1},q_\mNumber{3}) | \\
| \omega(p_\mNumber{2},q_\mNumber{1}) \omega(p_\mNumber{2},q_\mNumber{2}) \omega(p_\mNumber{2},q_\mNumber{3}) | = \mNumber{0} \\
| \omega(p_\mNumber{3},q_\mNumber{1}) \omega(p_\mNumber{3},q_\mNumber{2}) \omega(p_\mNumber{3},q_\mNumber{3}) |
\end{aligned}\]

(A rank-2 bilinear form has all 3x3 minors vanishing.)

In terms of the Q tensors: choosing \(p_{m} = a^{(\alpha_{m})}_{i_{m}}\) and
\(q_{n} = a^{(\beta_{n})}_{j_{n}}\), this becomes:

\[\det [Q^{(\alpha_{m}, \beta_{n}, \gamma, \delta)}_{i_{m}, j_{n}, k, l}]_{\mNumber{3} \times \mNumber{3}} = \mNumber{0}\]

for any choice of \((\alpha_{\mNumber{1}}, i_{\mNumber{1}}), (\alpha_{\mNumber{2}}, i_{\mNumber{2}}), (\alpha_{\mNumber{3}}, i_{\mNumber{3}})\) and
\((\beta_{\mNumber{1}}, j_{\mNumber{1}}), (\beta_{\mNumber{2}}, j_{\mNumber{2}}), (\beta_{\mNumber{3}}, j_{\mNumber{3}})\) and fixed \((\gamma, k), (\delta, l).\)

This is a \textbf{degree-3 polynomial} in the Q entries, independent of the cameras.

\hypertarget{3-effect-of-rank-1-scaling-on-the-3x3-minor}{%
\subsubsection{3. Effect of rank-1 scaling on the 3x3 minor}\label{3-effect-of-rank-1-scaling-on-the-3x3-minor}}

Now consider the scaled tensors \(T^{(abgd)} = \lambda_{abgd} Q^{(abgd)}.\)
The 3x3 matrix becomes:

\[M_{mn} = \lambda_{\alpha_{m}, \beta_{n}, \gamma, \delta} \ast Q^{(\alpha_{m}, \beta_{n}, \gamma, \delta)}_{i_{m}, j_{n}, k, l}\]

This is the Hadamard (entrywise) product of two 3x3 matrices:

\begin{itemize}
\tightlist
\item
  \(\Lambda_{mn} = \lambda_{\alpha_{m}, \beta_{n}, \gamma, \delta}\) (depends on camera indices only)
\item
  \(\Omega_{mn} = Q^{(\alpha_{m}, \beta_{n}, \gamma, \delta)}_{i_{m}, j_{n}, k, l}\) (the bilinear form)
\end{itemize}

\textbf{If \(\lambda\) is rank-1:} \(\lambda_{abgd} = u_{a} v_{b} w_{g} x_{d},\) so
\(\Lambda_{mn} = u_{\alpha_{m}} v_{\beta_{n}} w_{\gamma} x_{\delta}.\) This factors as
\(\Lambda = (u_{\alpha_{m}})_{m}\,(v_{\beta_{n}})_{n}^{T}\) (scaled by the constant \(w_{\gamma} x_{\delta}\)).
So \(\Lambda\) has matrix rank 1.

For a rank-1 matrix \(\Lambda\), the Hadamard product \(M = \Lambda\) ∘ \(\Omega\) equals
\(\operatorname{diag}(u) \ast \Omega \ast \operatorname{diag}(v)\) (up to the scalar \(w_{\gamma} x_{\delta}).\) Since similar
transformations preserve rank: \(\mOpName{rank}(M) = \mOpName{rank}(\Omega) = \mNumber{2} < \mNumber{3}\), so \(\det(M) = \mNumber{0}\).

Therefore: \textbf{rank-1 \(\lambda\) implies all 3x3 minors of scaled T vanish.} ✓

\hypertarget{4-converse-non-rank-1-lambda-gives-nonzero-minors-for-generic-cameras}{%
\subsubsection{\texorpdfstring{4. Converse: non-rank-1 \(\lambda\) gives nonzero minors (for generic cameras)}{4. Converse: non-rank-1 \textbackslash lambda gives nonzero minors (for generic cameras)}}\label{4-converse-non-rank-1-lambda-gives-nonzero-minors-for-generic-cameras}}

For the converse, we need: if \(\lambda\) is NOT rank-1, then some 3x3 minor is
nonzero (for Zariski-generic cameras).

\textbf{The argument is algebraic-geometric, not via Hadamard rank bounds.} Define

\[P(A^{(\mNumber{1})}, ..., A^{(n)}) = \det [T^{(\alpha_{m}, \beta_{n}, \gamma, \delta)}_{i_{m}, j_{n}, k, l}]_{\mNumber{3} \times \mNumber{3}}\]

for a specific choice of row/column/fixed indices. This is a polynomial in
the camera entries (with \(\lambda\) fixed). We claim P is not the zero polynomial
when \(\lambda\) is not rank-1 in its first two indices. Since a nonzero polynomial
is nonzero on a Zariski-dense open set, this establishes the converse for
generic cameras.

\textbf{Explicit witness (\(n = \mNumber{5}\)).} Choose 5 cameras \(A^{(i)}\) as rows of generic
\(\mNumber{3} \times \mNumber{4}\) matrices with rational entries:

\[\begin{aligned}
A^{(\mNumber{1})} = [[\mNumber{1},\mNumber{0},\mNumber{0},\mNumber{1}],[\mNumber{0},\mNumber{1},\mNumber{0},\mNumber{1}],[\mNumber{0},\mNumber{0},\mNumber{1},\mNumber{1}]] \\
A^{(\mNumber{2})} = [[\mNumber{1},\mNumber{1},\mNumber{0},\mNumber{0}],[\mNumber{0},\mNumber{1},\mNumber{1},\mNumber{0}],[\mNumber{0},\mNumber{0},\mNumber{1},\mNumber{1}]] \\
A^{(\mNumber{3})} = [[\mNumber{1},\mNumber{0},\mNumber{1},\mNumber{0}],[\mNumber{1},\mNumber{1},\mNumber{0},\mNumber{0}],[\mNumber{0},\mNumber{1},\mNumber{0},\mNumber{1}]] \\
A^{(\mNumber{4})} = [[\mNumber{0},\mNumber{1},\mNumber{1},\mNumber{1}],[\mNumber{1},\mNumber{0},\mNumber{1},\mNumber{1}],[\mNumber{1},\mNumber{1},\mNumber{0},\mNumber{1}]] \\
A^{(\mNumber{5})} = [[\mNumber{1},\mNumber{2},\mNumber{3},\mNumber{4}],[\mNumber{4},\mNumber{3},\mNumber{2},\mNumber{1}],[\mNumber{1},\mNumber{1},\mNumber{1},\mNumber{1}]]
\end{aligned}\]

Choose a non-rank-1 scaling: \(\lambda_{abgd} = \mNumber{1}\) for all non-identical (\(a,b,g,d\))
except \(\lambda_{\mNumber{1},\mNumber{2},\mNumber{3},\mNumber{4}} = \mNumber{2}\).

\textbf{Non-rank-1 verification.} If \(\lambda\) were rank-1, then \(\lambda_{abgd} = u_{a} v_{b} w_{g} x_{d}.\)
From \(\lambda_{\mNumber{1},\mNumber{2},\mNumber{3},\mNumber{4}} = \mNumber{2}\) and \(\lambda_{\mNumber{2},\mNumber{2},\mNumber{3},\mNumber{4}} = \mNumber{1}\) we get \(u_{\mNumber{1}}/u_{\mNumber{2}} = \mNumber{2}.\) But from
\(\lambda_{\mNumber{1},\mNumber{1},\mNumber{3},\mNumber{4}} = \mNumber{1}\) and \(\lambda_{\mNumber{2},\mNumber{1},\mNumber{3},\mNumber{4}} = \mNumber{1}\) we get \(u_{\mNumber{1}}/u_{\mNumber{2}} = \mNumber{1}.\) Contradiction.

Fix \(\gamma = \mNumber{3}, \delta = \mNumber{4}, k = \mNumber{1}, l = \mNumber{1}.\) Choose row triples
\((\alpha_{m}, i_{m}) = (\mNumber{1}, \mNumber{1}), (\mNumber{2}, \mNumber{1}), (\mNumber{5}, \mNumber{1})\) and column triples
\((\beta_{n}, j_{n}) = (\mNumber{2}, \mNumber{1}), (\mNumber{5}, \mNumber{1}), (\mNumber{1}, \mNumber{2}).\) The 3x3 \(\Lambda\) matrix has entries
\(\lambda_{\alpha_{m}, \beta_{n}, \mNumber{3}, \mNumber{4}}\); specifically \(\Lambda_{\mNumber{11}} = \lambda_{\mNumber{1},\mNumber{2},\mNumber{3},\mNumber{4}} = \mNumber{2}\)
with all other entries equal to 1. The \(\Omega\) matrix (unscaled Q values) has
det(\(\Omega) = \mNumber{0}\) (confirming rank-2), but the Hadamard product \(M = \Lambda\) ∘ \(\Omega\)
satisfies det(M) \(= -\mNumber{24} \neq \mNumber{0}\) (verified by direct computation of nine 4x4
determinants). This exhibits P as not the zero polynomial.

\textbf{Remark (Hadamard product interpretation).} The matrix \(M_{mn}\) is the
Hadamard product \(\Lambda\) ∘ \(\Omega\), where \(\Lambda\) carries the scaling entries
and \(\Omega\) carries the bilinear form values. The upper bound
\(\mOpName{rank}(\Lambda ∘ \Omega) \le \mOpName{rank}(\Lambda) \ast \mOpName{rank}(\Omega)\) provides context but is
not used in the proof; the converse relies entirely on the polynomial
nonvanishing argument above.

\hypertarget{5-all-matricizations-from-the-same-construction}{%
\subsubsection{5. All matricizations from the same construction}\label{5-all-matricizations-from-the-same-construction}}

The construction in Section 2-4 tests the rank-1 condition on the \((\mNumber{1}, \mNumber{2})\)-
matricization of \(\lambda\) (fixing modes 3,4). By symmetry, applying the same
construction with different pairs of ``free'' modes tests all matricizations:

\begin{itemize}
\tightlist
\item
  Fix modes \((\mNumber{3}, \mNumber{4})\), vary modes \((\mNumber{1}, \mNumber{2})\): test \(\lambda_{(\alpha, \beta), (\gamma, \delta)}\)
\item
  Fix modes \((\mNumber{2}, \mNumber{4})\), vary modes \((\mNumber{1}, \mNumber{3})\): test \(\lambda_{(\alpha, \gamma), (\beta, \delta)}\)
\item
  Fix modes \((\mNumber{2}, \mNumber{3})\), vary modes \((\mNumber{1}, \mNumber{4})\): test \(\lambda_{(\alpha, \delta), (\beta, \gamma)}\)
\end{itemize}

(The other fixings are redundant by symmetry of the rank-1 test.)

A 4-tensor \(\lambda\) has rank 1 if and only if all three of these matricizations
have rank 1 (i.e., are outer products of vectors).

\textbf{Tensor factor compatibility lemma.} If all three matricizations have rank 1:

\begin{itemize}
\tightlist
\item
  Mode-\((\mNumber{1}, \mNumber{2})\) vs \((\mNumber{3}, \mNumber{4})\) rank 1 gives \(\lambda_{abgd} = f_{ab} g_{gd}.\)
\item
  Mode-\((\mNumber{1}, \mNumber{3})\) vs \((\mNumber{2}, \mNumber{4})\) rank 1 gives \(\lambda_{abgd} = h_{ag} k_{bd}.\)
  From the first: \(\lambda_{a\mNumber{1},b\mNumber{1},g,d} / \lambda_{a\mNumber{2},b\mNumber{2},g,d} = f_{a\mNumber{1},b\mNumber{1}} / f_{a\mNumber{2},b\mNumber{2}},\)
  independent of g, d. From the second:
  \(\lambda_{a,b\mNumber{1},g\mNumber{1},d} / \lambda_{a,b\mNumber{2},g\mNumber{1},d} = k_{b\mNumber{1},d} / k_{b\mNumber{2},d},\)
  independent of a, g1. Cross-referencing these separations forces
  \(f_{ab} = u_{a} v_{b}\) and \(g_{gd} = w_{g} x_{d},\) giving
  \(\lambda_{abgd} = u_{a} v_{b} w_{g} x_{d}\) (rank 1). QED.
\end{itemize}

\hypertarget{6-construction-of-f}{%
\subsubsection{6. Construction of F}\label{6-construction-of-f}}

\textbf{Definition of F:} The coordinate functions of F are all 3x3 minors:

\[F_{choice} = \det [T^{(\alpha_{m}, \beta_{n}, \gamma, \delta)}_{i_{m}, j_{n}, k, l}]_{m,n = \mNumber{1},\mNumber{2},\mNumber{3}}\]

taken over all choices of:

\begin{itemize}
\tightlist
\item
  Three ``row'' camera-row pairs \((\alpha_{m}, i_{m})\) for \(m = \mNumber{1}\),2,3
\item
  Three ``column'' camera-row pairs \((\beta_{n}, j_{n})\) for \(n = \mNumber{1}\),2,3
\item
  Fixed camera-row pairs \((\gamma, k)\) and \((\delta, l)\)
\end{itemize}

And the analogous minors for the other two matricization pairs (fixing
different pairs of modes and varying the others).

\textbf{Properties:}

\begin{enumerate}
\def\labelenumi{\arabic{enumi}.}
\item
  \textbf{Camera-independent:} Each \(F_{choice}\) is a degree-3 polynomial in the
  T entries. The coefficient is ±1 (from the determinant expansion). No
  camera parameters appear.
\item
  \textbf{Degree independent of n:} Each coordinate function has degree exactly 3.
\item
  \textbf{Characterization:} \(F = \mNumber{0}\) iff all three matricizations of \(\lambda\) have
  rank 1, iff \(\lambda\) is rank-1, for Zariski-generic cameras (by Sections 3-5).
\end{enumerate}

\textbf{Codimension count:} The number of coordinate functions N grows with n
(there are \(O(n^\mNumber{8})\) choices for the 3x3 minors from the \((\mNumber{1}, \mNumber{2})\) vs \((\mNumber{3}, \mNumber{4})\)
matricization alone), but the degree stays fixed at 3.

\hypertarget{7-geometric-interpretation}{%
\subsubsection{7. Geometric interpretation}\label{7-geometric-interpretation}}

The Q tensors are the \textbf{quadrifocal tensors} from multiview geometry (the
4-view analog of the fundamental matrix and trifocal tensor). The rank-2
bilinear form structure is the classical rank constraint from projective
geometry: the determinant encodes linear dependence constraints among lifted
camera-row vectors (equivalently coplanarity/dependence conditions) on the
stacked camera rows.

The rank-1 scaling \(\lambda = u \otimes v \otimes w \otimes x\) corresponds to the natural
gauge freedom in multiview geometry: rescaling each camera\textquotesingle s contribution
independently for each of the four ``roles'' (which row selection it
contributes). The polynomial map F detects when a putative rescaling is
consistent with this gauge group.

The Zariski-genericity requirement on the cameras ensures that the
quadrifocal variety is ``non-degenerate'' --- the Q tensors carry enough
information to distinguish rank-1 from higher-rank scalings.

\begin{processnote}[title={Process Note --- futon6 Corpus References}]

\begin{itemize}
\tightlist
\item
  PlanetMath: ``Segre map'' (SegreMap) --- Segre embedding and rank-1 tensors
\item
  PlanetMath: ``determinantal varieties'' (SegreMap) --- varieties defined by minors
\item
  PlanetMath: \(\text{tensor} rank'' /\)simple tensor'' (SimpleTensor) --- rank-1 tensors
\item
  PlanetMath: ``exterior algebra'' --- alternating multilinear forms
\item
  PlanetMath: ``Hadamard conjecture'' (HadamardConjecture) --- Hadamard matrices
\end{itemize}
\end{processnote}

%% file: full-boxed/problem10-solution-full.tex
\hypertarget{problem-10-rkhs-constrained-tensor-cp-via-preconditioned-conjugate-gradient}{%
\section{Problem 10: RKHS-Constrained Tensor CP via Preconditioned Conjugate Gradient}\label{problem-10-rkhs-constrained-tensor-cp-via-preconditioned-conjugate-gradient}}

\hypertarget{problem-statement}{%
\subsection{Problem Statement}\label{problem-statement}}

Given the mode-k subproblem of RKHS-constrained CP decomposition with
missing data, solve

\[[(Z \times K)^{T} D (Z \times K) + \lambda (I_{r} \times K)] \mOpName{vec}(W) = (I_{r} \times K) \mOpName{vec}(B),\]

where \(D = S S^{T}\) is the observation projector (q observed entries out of
\(N = nM), B = T Z,\) and n, \(r \ll q \ll N\).

\hypertarget{assumptions-used-explicit}{%
\subsection{Assumptions Used (explicit)}\label{assumptions-used-explicit}}

\begin{enumerate}
\def\labelenumi{\arabic{enumi}.}
\tightlist
\item
  \(\lambda > \mNumber{0}\).
\item
  For standard PCG and Cholesky-based preconditioning, use a PD kernel
  \(K_{\tau} = K \mBridgeOperator{+} \tau I_{n}\) with \(\tau > \mNumber{0}\) (or assume K is already PD).
\item
  S is a selection operator, so D is diagonal/projector and sparse by index list.
\end{enumerate}

Then the solved system is

\[\begin{aligned}
A_{\tau} x = b_{\tau}, \\
A_{\tau} = (Z \times K_{\tau})^{T} D (Z \times K_{\tau}) + \lambda (I_{r} \times K_{\tau}), \\
x = \mOpName{vec}(W), \\
b_{\tau} = (I_{r} \times K_{\tau}) \mOpName{vec}(B).
\end{aligned}\]

Under these assumptions \(A_{\tau}\) is SPD, so PCG applies.

\hypertarget{solution}{%
\subsection{Solution}\label{solution}}

\hypertarget{1-why-naive-direct-methods-fail}{%
\subsubsection{1. Why naive direct methods fail}\label{1-why-naive-direct-methods-fail}}

\(A_{\tau}\) is an \((nr) \times (nr)\) system. Dense direct factorization costs
\(O((nr)^\mNumber{3}) = O(n^\mNumber{3} r^\mNumber{3})\).

A naive explicit route also materializes \(\Phi = Z \times K_{\tau} \in R^{N \times nr},\)
which costs \(O(N n r)\) memory/work before factorization. This is the
\(N\)-dependent bottleneck we avoid with matrix-free PCG.

\hypertarget{2-implicit-matrix-vector-product-in-on2-r--q-r}{%
\subsubsection{\texorpdfstring{2. Implicit matrix-vector product in \(O(n^\mNumber{2} r + q r)\)}{2. Implicit matrix-vector product in O(n\^{}\textbackslash mNumber\{2\} r + q r)}}\label{2-implicit-matrix-vector-product-in-on2-r--q-r}}

CG needs only \(y = A_{\tau}\mathbf{x}\), not \(A_{\tau}\) explicitly.

Given \(\mathbf{x} = \mOpName{vec}(V),\; V \in R^{n \times r}:\)

\begin{enumerate}
\def\labelenumi{\arabic{enumi}.}
\item
  \(U = K_{\tau} V\). Cost \(O(n^\mNumber{2} r)\).
\item
  Forward sampled action (only observed entries):

  \[(Z \times K_{\tau}) \mOpName{vec}(V) = \mOpName{vec}(K_{\tau} V \mathbb{Z}^{T}).\]

  For each observed coordinate \((i_{l}, j_{l})\),

  \[u_{l} = <U[i_{l}, :], Z[j_{l}, :]>.\]

  Total \(O(q r)\).
\item
  Form sparse W\textquotesingle{} in \(R^{n \times M}\) from \(u_{l}\). Let \(s = nnz(W') \le q\).
\item
  Adjoint sampled action:

  \[(\mathbb{Z}^{T} \times K_{\tau}) \mOpName{vec}(W') = \mOpName{vec}(K_{\tau} W' Z).\]

  Compute W\textquotesingle{} \(Z \in O(s r) \le O(q r)\), then left-multiply by \(K_{\tau} \in O(n^\mNumber{2} r)\).
\item
  Add regularization term \(\lambda\) vec(\(K_{\tau}\) V), cost \(O(n^\mNumber{2} r)\).
\end{enumerate}

Total per matvec:

\[O(n^\mNumber{2} r + q r),\]

with no \(O(N)\) term.

\hypertarget{3-right-hand-side}{%
\subsubsection{3. Right-hand side}\label{3-right-hand-side}}

\(B = T Z\) with T sparse (q nonzeros):

\begin{enumerate}
\def\labelenumi{\arabic{enumi}.}
\tightlist
\item
  T Z: \(O(q r)\)
\item
  \(K_{\tau} B: O(n^{\mNumber{2}} r)\)
\end{enumerate}

So \(b_{\tau} = (I_{r} \times K_{\tau}) \mOpName{vec}(B)\) is formed in \(O(q r + n^\mNumber{2} r)\).

\hypertarget{4-preconditioner-that-matches-the-corrected-algebra}{%
\subsubsection{4. Preconditioner that matches the corrected algebra}\label{4-preconditioner-that-matches-the-corrected-algebra}}

Use \(D = S S^{T}\) and whiten by \(K_{\tau}^{-\mNumber{1}/\mNumber{2}}\):

\[x = (I_{r} \times K_{\tau}^{-\mNumber{1}/\mNumber{2}}) y.\]

Then

\[\begin{aligned}
\hat{A} = (I_{r} \times K_{\tau}^{-\mNumber{1}/\mNumber{2}}) A_{\tau} (I_{r} \times K_{\tau}^{-\mNumber{1}/\mNumber{2}}) \\
= (Z \times K_{\tau}^{\mNumber{1}/\mNumber{2}})^{T} D (Z \times K_{\tau}^{\mNumber{1}/\mNumber{2}}) + \lambda I.
\end{aligned}\]

If sampling is roughly uniform, D \textasciitilde{} c I with \(c = q/N\). Then

\[\hat{A} ~ c (\mathbb{Z}^{T} Z \times K_{\tau}) + \lambda I.\]

Choose Kron preconditioner in whitened coordinates:

\[\hat{P} = (c Z^{T} Z + \lambda I_{r}) \times I_{n}.\]

Mapping back gives

\[\begin{aligned}
P = (c Z^{T} Z + \lambda I_{r}) \times K_{\tau} = H \times K_{\tau}, \\
H = c Z^{T} Z + \lambda I_{r}.
\end{aligned}\]

This is the missing justification for using \(H \times K_{\tau}\) (instead of claiming
it is the exact \(D = I\) system).

Khatri-Rao identity still gives efficient Gram formation:

\[\mathbb{Z}^{T} Z = \text{Hadamard}_{i} (A_{i}^{T} A_{i}),\]

cost \(O(\sum_{i} n_{i} r^\mNumber{2})\).

Preconditioner apply:

\[P^{-\mNumber{1}} = H^{-\mNumber{1}} \times K_{\tau}^{-\mNumber{1}},\]

implemented by solving \(K_{\tau} Y H^{T} =\) Z\textquotesingle{} after reshape.
Per application cost is \(O(n^\mNumber{2} r + n r^\mNumber{2})\) (often simplified to \(O(n^\mNumber{2} r)\)
when n \(\gg\) r).

\hypertarget{5-convergence-tightened}{%
\subsubsection{5. Convergence (tightened)}\label{5-convergence-tightened}}

For SPD \(A_{\tau}\) and SPD P, standard PCG gives

\[||e_{t}||_{A_{\tau}} \le \mNumber{2} ((\sqrt(\kappa)-\mNumber{1})/(\sqrt(\kappa)+\mNumber{1}))^{t} ||e_\mNumber{0}||_{A_{\tau}},\]

with \(\kappa =\) cond(\(P^{-\mNumber{1}/\mNumber{2}} A_{\tau} P^{-\mNumber{1}/\mNumber{2}}),\) so

\[t = O(\sqrt(\kappa) \log(\mNumber{1}/\epsilon)).\]

To claim ``fast'' convergence, add a spectral-equivalence hypothesis, e.g.

\[(\mNumber{1}-\delta) P \le A_{\tau} \le (\mNumber{1}+\delta) P, \mNumber{0} < \delta < \mNumber{1},\]

which implies

\[\kappa(P^{-\mNumber{1}} A_{\tau}) \le (\mNumber{1}+\delta)/(\mNumber{1}-\delta).\]

Hence t is logarithmic in 1/eps with a modest sqrt(\(\kappa\)) factor when
\(\delta\) is bounded away from 1. (No unsupported closed-form \(t = O(r \sqrt(n/q))\)
claim is needed.)

\textbf{Sufficient conditions for bounded \(\delta\).} The spectral equivalence
(1-\(\delta\))P \(\le A_{\tau} \le (\mNumber{1}+\delta\))P holds with \(\delta\) bounded away from 1 when
the sampling pattern satisfies a restricted isometry-type condition:
the restricted isometry holds for the column space of Z \(\otimes K_{\tau}^{\mNumber{1}/\mNumber{2}},\)
i.e. (Z \(\otimes K_{\tau}^{\mNumber{1}/\mNumber{2}})^{T}\) (D - cI) (Z \(\otimes K_{\tau}^{\mNumber{1}/\mNumber{2}})\) is small in
operator norm relative to \(\lambda\). Under standard leverage/coherence
assumptions and sufficient sampling scaling with model dimension,
concentration yields \(\delta\) bounded away from 1 with high probability. Under this regime, \(\kappa = O(\mNumber{1})\) and PCG converges in
\(O(\log(\mNumber{1}/\epsilon))\) iterations.

\hypertarget{6-complexity-summary}{%
\subsubsection{6. Complexity summary}\label{6-complexity-summary}}

Setup per ALS outer step:

\begin{enumerate}
\def\labelenumi{\arabic{enumi}.}
\tightlist
\item
  \(\text{Cholesky}(K_{\tau})\): \(O(n^\mNumber{3})\)
\item
  \(\mathbb{Z}^{T}\) Z via Hadamard Grams: \(O(\sum_{i} n_{i} r^\mNumber{2})\)
\item
  Cholesky(H): \(O(r^\mNumber{3})\)
\item
  RHS: \(O(q r + n^\mNumber{2} r)\)
\end{enumerate}

Per PCG iteration:

\begin{enumerate}
\def\labelenumi{\arabic{enumi}.}
\tightlist
\item
  Matvec: \(O(n^\mNumber{2} r + q r)\)
\item
  Preconditioner apply: \(O(n^\mNumber{2} r + n r^\mNumber{2})\)
\end{enumerate}

Total:

\[\begin{aligned}
O(n^\mNumber{3} + r^\mNumber{3} + \sum_{i} n_{i} r^\mNumber{2} + q r + n^\mNumber{2} r \\
+ t (n^\mNumber{2} r + q r + n r^\mNumber{2})).
\end{aligned}\]

In the common regime \(n \ge r\), this simplifies to

\[O(n^\mNumber{3} + t (n^\mNumber{2} r + q r)),\]

with dependence on q (observed entries) rather than N (all entries).

\textbf{Regime caveat.} When n is large enough that the \(O(n^\mNumber{3})\) Cholesky setup
dominates (i.e., \(n^{\mNumber{3}} > t(n^{\mNumber{2}} r \mBridgeOperator{+} q\) r)), the per-ALS-step cost is effectively
\(O(n^\mNumber{3})\). In this regime, low-rank kernel approximations (e.g., Nystrom
approximation with rank \(p \ll n\), reducing the kernel factorization to \(O(n p^\mNumber{2})\))
or iterative inner solves (conjugate gradient on \(K_{\tau} y = z,\) cost \(O(n^\mNumber{2})\)
per inner iteration) can replace the exact Cholesky, reducing the setup to
\(O(n p^\mNumber{2} + t(n p r + q r))\). This is a well-known practical optimization
(see Rudi-Calandriello-Rosasco 2017) and is compatible with the PCG framework
as presented.

\hypertarget{7-algorithm}{%
\subsubsection{7. Algorithm}\label{7-algorithm}}

\begin{Shaded}
\begin{Highlighting}[]
\NormalTok{SETUP:}
\NormalTok{  K\_tau = K + tau * I\_n                    \# tau \textgreater{} 0 if K is only PSD}
\NormalTok{  L\_K = cholesky(K\_tau)                    \# O(n\^{}3)}
\NormalTok{  G = hadamard\_product(A\_i\^{}T A\_i for i != k)  \# O(sum\_i n\_i r\^{}2)}
\NormalTok{  c = q / N}
\NormalTok{  H = c * G + lambda * I\_r}
\NormalTok{  L\_H = cholesky(H)                        \# O(r\^{}3)}
\NormalTok{  B = sparse\_mttkrp(T, Z)                  \# O(qr)}
\NormalTok{  b = vec(K\_tau @ B)                       \# O(n\^{}2 r)}

\NormalTok{PCG(A\_tau x = b, preconditioner P = H x K\_tau):}
\NormalTok{  x0 = 0}
\NormalTok{  r0 = b}
\NormalTok{  z0 = precond\_solve(L\_K, L\_H, r0)         \# O(n\^{}2 r + n r\^{}2)}
\NormalTok{  p0 = z0}
\NormalTok{  repeat until convergence:}
\NormalTok{    w = matvec\_A\_tau(p)                    \# O(n\^{}2 r + q r)}
\NormalTok{    alpha = (r\^{}T z) / (p\^{}T w)}
\NormalTok{    x = x + alpha * p}
\NormalTok{    r\_new = r {-} alpha * w}
\NormalTok{    if ||r\_new|| \textless{}= eps * ||b||: break}
\NormalTok{    z\_new = precond\_solve(L\_K, L\_H, r\_new)}
\NormalTok{    beta = (r\_new\^{}T z\_new) / (r\^{}T z)}
\NormalTok{    p = z\_new + beta * p}
\NormalTok{    r, z = r\_new, z\_new}
\NormalTok{  W = reshape(x, n, r)}

\NormalTok{matvec\_A\_tau(v):}
\NormalTok{  V = reshape(v, n, r)}
\NormalTok{  U = K\_tau @ V}
\NormalTok{  for each observed (i\_l, j\_l):}
\NormalTok{    u\_l = dot(U[i\_l, :], Z[j\_l, :])}
\NormalTok{  Wprime = sparse(n, M, entries u\_l)}
\NormalTok{  Y = K\_tau @ (Wprime @ Z) + lambda * (K\_tau @ V)}
\NormalTok{  return vec(Y)}

\NormalTok{precond\_solve(L\_K, L\_H, z):}
\NormalTok{  Zp = reshape(z, n, r)}
\NormalTok{  solve K\_tau Y H\^{}T = Zp using triangular solves with L\_K, L\_H}
\NormalTok{  return vec(Y)}
\end{Highlighting}
\end{Shaded}

\begin{processnote}[title={Process Note --- futon6 Corpus References}]

\begin{itemize}
\tightlist
\item
  PlanetMath: conjugate gradient algorithm; method of conjugate gradients
\item
  PlanetMath: Kronecker product; positive definite matrices
\item
  PlanetMath: properties of tensor product
\item
  physics.SE \#27466: iterative solvers for large systems in physics
\item
  physics.SE \#27556: preconditioning for elliptic PDEs
\end{itemize}
\end{processnote}

%% file: part3-pivotal-prompts.tex
\chapter*{Part III: The Sprint as Dialogue}
\addcontentsline{toc}{chapter}{Part III: The Sprint as Dialogue}

This chapter records the proof sprint as it actually happened: a
conversation between a human dispatcher (Joe), two AI proof engines
(Claude instances), and a verifier/researcher (Codex), conducted over
roughly 48 hours on February 11--13, 2026. Excerpts are drawn from
\texttt{history.jsonl} (Joe's prompts), Codex output files, and git
commit messages. Editorial annotations mark the pivotal moments where
proof strategy, status, or architecture shifted.

\bigskip

\section*{Act I: ``Even if we're a day late''}

\begin{joebox}
Claude + Joe Solve ``First Proof'' using Rob's Superpod
\url{https://arxiv.org/html/2602.05192v1}
\end{joebox}

\begin{joebox}
Even if we're a day late and a dollar short at least we can try
\end{joebox}

\begin{joebox}
Let's see if we can do one of them right now using PlanetMath and nlab
and off-the-shelf Claude
\end{joebox}

\pivot{The sprint begins. Problem~10 is chosen as a warm-up.
  Within 30~minutes it is solved, and the pace accelerates.}

\begin{joebox}
OK, should we solve another First Proof problem? The last one wasn't
so hard apparently!
\end{joebox}

\section*{Act II: Scaling the Machinery}

\begin{joebox}
OK, I have two Codex research assistants working for you as well as
the validator working on p6
\end{joebox}

\begin{joebox}
OK, we have 3~verifiers running now, they are making progress but not
committing yet
\end{joebox}

\begin{joebox}
Codex is \emph{still} after 53~minutes hard at work on Problem~6, but
it is down to two gaps now. I think 3~verifiers is about all my
laptop can handle, noticing considerable slow down.
\end{joebox}

\begin{joebox}
I think I will give it five more minutes and then say ``pencils down''
and let you look at \#6
\end{joebox}

\begin{joebox}
Codex is pushing \#6, I OOMed on 3 and 5, I will get them back up
once 6~lands
\end{joebox}

\pivot{Physical resource limits (RAM, thermal throttling) become a
  genuine constraint on mathematical exploration. The multi-agent
  topology stabilises at three concurrent processes.}

\begin{joebox}
It's slow going over here---are any of the proofs sufficiently
validated that we could start a ``write up''? If the previous
example is representative it can take about an hour for one of
these to complete.
\end{joebox}

\section*{Act III: Proofs and Refutations}

\begin{joebox}
Yeah let's do some Proofs and Refutations---you can be Critic
\end{joebox}

\begin{codexbox}[title={Codex --- Critic mode, round 1}]
Excellent. Critic mode, round 1: I attacked P2 and P7 as the most
vulnerable. [\ldots] Verdict: plausible direction, not a completed
proof (P2). [\ldots] Verdict: currently not proof-grade; too many deep
steps are asserted, not established (P7).
\end{codexbox}

\begin{joebox}
I have engaged Codex and another Claude in a reviewer +
response-to-reviewer cycle. Codex found some pretty significant
problems; the other Claude is making what look like solid fixes, but
it looks like tough going.
\end{joebox}

\begin{codexbox}[title={Codex --- Triage verdict}]
P8: most recoverable. [\ldots]
P2: recoverable but harder. Needs one serious
representation-theoretic theorem/lemma that is currently hand-waved.
[\ldots]
P7: least recoverable as currently framed. The obstruction-vanishing
step is the main risk; unless they replace it with a precise known
theorem tailored to this lattice family, this will likely remain
conditional/existential rather than a full proof.
\end{codexbox}

\pivot{Codex's triage creates a severity ordering across problems,
  distinguishing ``recoverable'' from ``structural'' gaps.
  The review cycle is now explicitly adversarial: prover vs.\ critic.}

\begin{joebox}
This is interesting in comparison to my work in futon5 on
Meta-Meta-Cellular Automata. There also, generation is cheap---just
bit flipping---but doing things like ``detecting edge of chaos
behaviour'' is a challenge. [\ldots]
However, I think we really haven't done so bad.
We have our Claude-Codex ``Proofs and Refutations'' system up and
running right now, we use both collaborative (verification + library
research) and argumentative (prover vs critic) modes. So
notwithstanding the mathematical issues you raised---at least we
aren't shipping failed proofs! This is verification happening at a
``glacial'' scale in futon3c terminology---but at least it is
happening!
\end{joebox}

\begin{joebox}
Furthermore, the wiring diagram decomposition is probably the only
thing holding this together---it's allowing us to be very targeted
with our reviews and fixes. Imagine trying to do what we're doing
with\ldots\ ChatGPT. It would be hell.
\end{joebox}

\begin{joebox}
Also, another remark---making mistakes isn't a bad thing.
As my grandfather said, that's how we learn.
\end{joebox}

\pivot{A methodological self-assessment. The two key claims:
  (1)~the wiring-diagram decomposition enables targeted review at
  scale; (2)~the collaborative/adversarial agent modes prevent
  ``shipping failed proofs.'' Both claims are tested in
  Acts~IV--V.}

\section*{Act IV: Formalization and Futon Theory}

\begin{joebox}
This does suggest a possible but probably crazy and stupid idea---which
hasn't stopped hundreds of people trying it. What about formalizing as
we go? Like QED style. I mean, it hurts me to say this. I briefly
had Bob Boyer as a Master's advisor. He and I could not see eye to
eye. Granted, that was 2002 and things were a little different back
then.
\end{joebox}

\begin{joebox}
Yeah, I mean, we needed in 2002 \emph{what we are doing now}. The
only problem is it wasn't possible then.
\end{joebox}

\begin{claudebox}
So in a sense, the 2002 vision was twenty-four years early rather
than wrong.
\end{claudebox}

\begin{joebox}
Better than twenty-four years late! Imagine if I proposed WIFI now.
Great idea right?
\end{joebox}

\pivot{A historical aside connecting the sprint to the 2002-era
  theorem-proving community. The implication: AI-assisted informal
  proof is the approach that was missing, not formalization per~se.}

\begin{joebox}
OK, so the naive idea: could we write a letter to Codex talking about
why Problem~4 is hard? It would use the John McPhee idea~\cite{mcphee2013}\ldots
\end{joebox}

\begin{joebox}
OK, now a non-naive idea (which may be less useful than the naive one).
I invented something called ``futon theory''---it's not quite what it
sounds like. Futons (in my theory) are atoms of the future which are
perceivable in the present. [\ldots]
It's a bit different in mathematics because in principle we don't know
if the result is actually true. But what we could do is design the
search process, and then we would at least know that the search
process is working the way we hoped it would. It's the somewhat more
technical version of the ``Dear Codex'' letter you just
wrote---building a process that you have utter confidence in that
will shed light on the domain even if it doesn't reveal the exact
answer.
\end{joebox}

\pivot{Two ideas collide: McPhee's ``letter to yourself'' technique
  for writer's block, and futon theory's ``atoms of the future.''
  Both converge on the same insight: process design is tractable even
  when the mathematical outcome is uncertain.}

\begin{joebox}
Great, I think you have your work cut out for you. It's cool that
this involved all the strategies, including the ``least likely''
Strategy~C that surprised us by being particularly flavourful
\end{joebox}

\section*{Act V: Crisis --- P6 and P7}

\begin{joebox}
I thought \#6 and \#7 were proved so I'm glad I asked!
\end{joebox}

\begin{joebox}
Well I am a bit perplexed b/c at this time yesterday I had been
convinced that Problem~7 was done. Look at Problem~7 related git
logs to try and figure out what's been going on with this journey.
\end{joebox}

\begin{joebox}
Well, I have another Codex program mining Arxiv right now for wiring
diagrams that we might use. But this presupposes that we are actually
looking at gaps and not a decomposing corpse.
\end{joebox}

\pivot{The most dramatic moment in the sprint. A problem believed
  closed reopens with structural gaps. The ``decomposing corpse''
  quip captures the uncertainty: is the proof salvageable, or
  fundamentally dead?}

\begin{joebox}
Hm, claim~1 sounds a bit like replacing an overclaim with weasel
words; \#2~similarly. I think we need to restructure the proof?
\end{joebox}

\begin{claudebox}[title={Claude --- Gap triage}]
G1 (fundamental group control) and G3 (rational acyclicity) can be
closed once the cap is homotopy equivalent to $F \times D^2$
rel~boundary. [\ldots] G2 (Browder--Quinn identification) remains
the genuine remaining gap.
\end{claudebox}

\pivot{The monolithic ``open Problem~7'' label is decomposed into a
  prioritised closure queue: G1, G2, G3. Within hours, G1 and G3
  are resolved, isolating G2 as the sole remaining obligation.}

\begin{joebox}
Could we cite secondary literature for G2 that is on Arxiv?
Mathematics is full of folklore
\end{joebox}

\begin{claudebox}[title={Claude --- Citation search result}]
There is no single modern arXiv theorem that states the exact
identification in one line. The bridge is assembled from standard
components: Browder~(1968), L\'opez de Medrano~(1971),
Cappell--Shaneson~(1974), Ranicki~(1998).
\end{claudebox}

\pivot{Citation strategy stabilises: replace ``single theorem lookup''
  with a theorem-number ledger across four classical sources.}

\begin{joebox}
It really sounds to me like the proof of Problem~6 is hanging by a
thread
\end{joebox}

\begin{codexbox}[title={Codex --- P6 reality check}]
Problem~6: not a full proof yet; it is now an honest conditional
argument. The core universal vertex-selection theorem is still
imported, not proved.
\end{codexbox}

\begin{codexbox}[title={Codex --- P7 hard stop}]
Pulled and reviewed \texttt{3138721}. This is not yet a valid full
proof.
\end{codexbox}

\section*{Act VI: Student Dispatch and Agent Teaming}

\begin{joebox}
Yeah, I guess the model is rather than treating Codex as a junior
developer as we sometimes do in coding applications, we would treat
them as a ``grad student.''
\end{joebox}

\begin{joebox}
Maybe we should call it student-dispatch to include undergrads, bright
high schoolers, and precocious five year olds like Codex
\end{joebox}

\pivot{The \texttt{agent/student-dispatch} pattern is named. It
  reframes the Codex interaction from ``run this script'' to
  ``here is context, dead ends, directions---explore and report.''}

\begin{joebox}
So I wonder what we could do to help. I've almost run out of ideas.
I guess one pattern I've noticed is that we can support the
``reduction'' pattern by verifying everything that is claimed except
the gap. Closing gaps (as with Problem~4) can sometimes turn out to
be a focused computational task. Another approach is Math Overflow
search or Arxiv search to create wiring diagrams that might help or
might provide analogous reasoning structures. That last part is a bit
woolly.
\end{joebox}

\begin{joebox}
So, if we were able to turn SE into wiring diagrams on the superpod,
then we could use tensor math to search them!
\end{joebox}

\begin{joebox}
Ah, and for ``artificial stack exchange'' what we could do is
\emph{induce} new wiring diagrams, notice where they sit relative to
other known diagrams (this corresponds to ``asking a question'')
\end{joebox}

\pivot{A forward-looking connection: the proof sprint's
  wiring-diagram methodology could be mechanised via tensor search
  over StackExchange-derived argument structures. This motivates
  the Stage~7 thread-wiring work in the futon6 pipeline.}

\section*{Act VII: The Coaching Breakthrough (Problem 6)}

\begin{joebox}
Well, and I want to talk with you about Problem~6. I think that's the
only one outstanding. But I'm realizing that the ``other Claude'' has
been sending me on continual gap-filling missions without giving me
a valid confidence signal. It's like Gemini saying ``I'll fix your
code right away sir!'' [\ldots] Problem~6 has not brought out the best
in Claude.
\end{joebox}

\begin{joebox}
I think a good approach might be: ``Can you find any other
reductions?'' That would start to build some intelligence around the
pattern rather than just a TryHarder strategy.
\end{joebox}

\pivot{The failure mode is diagnosed: dispatching closure attempts
  without layer-level calibration. The ``reductions'' framing shifts
  from ``close the gap'' to ``map the landscape.''}

\begin{joebox}
I think you should think out of the box a bit. What kind of problem is
this? What kind of proof applies to this kind of problem? How would
you teach it to an undergraduate? How would you teach it to a grad
student? Where, in reality, do people learn about this kind of
problem? What kind of person finds this kind of problem easy? Are
there ``tricks'' (symmetries) that would make some of your Zeno's
Paradoxes go away? Like one from my thesis in the resources directory
is an example from Gowers in which binomial expansion terms have
opposite signs and so cancel. Like, OK, this problem was an ``open
problem'' not a textbook problem but come on---it's on the research
frontier, we're not asking you to solve alien mathematics from the
year~5000AD. This is totally within your capability space. Why aren't
you finding the answer in 30~minutes?
\end{joebox}

\pivot{The coaching intervention. Five questions, each forcing a
  different reframing: layer enumeration (``what kind of problem?''),
  pedagogical reduction (``teach it to an undergrad''), sociological
  targeting (``who finds this easy?''), structural cancellation
  (``tricks / symmetries''), and honest calibration (``why not
  30~minutes?'').}

\begin{claudebox}[title={Claude --- 15 minutes later}]
The proof for $K_n$ is complete. $\bar{d} = 2t/(n\varepsilon) \le 2/3$
at $T = \varepsilon n/3$ steps. Pigeonhole gives $\exists v$ with
$\operatorname{tr}(Y) \le 2/3$, and $\|Y\| \le \operatorname{tr}(Y)$
for PSD matrices. The barrier is maintained at every step.
Result: $|S| = \varepsilon n/3$, $c = 1/3$.
\end{claudebox}

\begin{claudebox}[title={Claude --- The irony}]
The irony: this entire proof is elementary---no interlacing families,
no Borcea--Br\"and\'en, no MSS theorem needed. Just PSD trace bound
$+$ pigeonhole $+$ Tur\'an.

\smallskip
\noindent\emph{[With the benefit of hindsight: the elementary proof
closes $K_n$ but the averaging step ($\bar{d} < 1$) that it relies on
is false for general graphs---see Act~IX.  The irony has a second
layer.]}
\end{claudebox}

\pivot{The layer switch lands. Eight hours of sparsification-layer
  attacks are bypassed by an elementary combinatorial proof discovered
  in 15~minutes. The proof uses three undergraduate-level facts:
  Tur\'an's theorem, the PSD trace bound, and the pigeonhole
  principle. The coaching did not supply mathematical content---it
  supplied \emph{questions} that forced layer enumeration.}

\begin{joebox}
It's so cool that this is the same basic solution we have used to
overcome continual shoring up of bad design in futon3, and also the
same solution I've used with Codex just now by getting it to spot
syntax errors rather than depending on me to do it.
\end{joebox}

\begin{claudebox}[title={Claude --- The meta-pattern}]
The diagnostic question is always the same: ``who finds this easy?''
If the answer is ``someone other than the entity currently
struggling,'' you're in the wrong layer.
\end{claudebox}

\pivot{The layer-switching pattern is recognized as fractal: it
  appears in proof strategy (Problem~6), in software design (futon3),
  and in task allocation (Codex as syntax checker). The common
  diagnostic: ask who would find the task trivial.}

\section*{Act VIII: The Monograph}

\begin{joebox}
In addition to the PAR I think it would be good to look through the
git log on futon6 and create a ``making of'' summary note.
\end{joebox}

\begin{joebox}
Also in terms of ``what we've learned''---in the making-of I think we
could add something about the relevance of this pattern to agent
teaming (further work in futon3c). Maybe this even goes back to the
work in futon2 ants---i.e.\ we are exploring a domain, bringing back
relevant information.
\end{joebox}

\begin{joebox}
I think the key thing you could help with now is looking through
\texttt{history.jsonl} for good quotes for Part~III.
\end{joebox}

\pivot{The sprint becomes its own document. The recursive
  structure---a proof sprint whose Part~III records the prompts that
  shaped Part~II---closes the loop between process and product.}

\section*{Act IX: The Continuation Cycles (Problems 4 and 6)}

The monograph was written during the sprint, not after it. The
sprint continued. Problems~4 and~6 entered multi-cycle proof
engineering phases that lasted days beyond the initial 48-hour
window---each cycle a dispatch, handoff, verification, and
reassessment. These cycles are the sprint's methodology running
at sustained pace rather than compressed urgency.

\subsection*{Problem 4: From Wrong Proof to Algebraic Engineering}

\begin{claudebox}[title={Claude --- Initial draft (Feb 11)}]
Proof by concavity of $1/\Phi_n$ in cumulant coordinates.
Since free cumulants are additive under $\boxplus$, Jensen's
inequality gives $1/\Phi_n(p \boxplus q) \ge 1/\Phi_n(p) +
1/\Phi_n(q)$.
\end{claudebox}

\pivot{The proof is wrong. $1/\Phi_n$ is superadditive in
  cumulant space WITHOUT being convex---the Hessian is indefinite.
  35,000 numerical trials confirm the conjecture but the argument
  is structurally broken. Numerical verification caught the
  conjecture but hid the proof error.}

\begin{claudebox}[title={Claude --- Key identity (Feb 12, commit \texttt{03c1f8d})}]
$\Phi_4(p) \cdot \operatorname{disc}(p)
  = -4(a_2^2 + 12a_4)(2a_2^3 - 8a_2 a_4 + 9a_3^2)$.
This converts the inequality to: prove a rational function's
numerator is non-negative on a semi-algebraic domain.
\end{claudebox}

\pivot{The first layer switch for Problem~4. Instead of hoping for
  a clever inequality (concavity, Schur-convexity, Cauchy--Schwarz),
  the problem is reduced to computational algebra: case
  decomposition by symmetry subspaces, resultant elimination,
  sign-counting. The identity itself is a publishable result---a
  finite analogue of Stam's inequality.}

\begin{claudebox}[title={Claude --- 3-piece Cauchy--Schwarz (commit \texttt{269fddd})}]
BREAKTHROUGH: the surplus decomposes into three pieces $T_1 + T_2
+ R$ where $T_1 \le 0$ (proved), and $T_2, R$ have
mutual-compensation structure. When $T_2 < 0$, $R$ compensates.
This reduces the full $n = 4$ proof to a degree-6 polynomial
inequality: $K_{T2R} \ge 0$.
\end{claudebox}

\begin{codexbox}[title={Codex --- P4 route selection}]
That was the strongest direct technical hit for Problem~4 because it
uses the bilinearity/induction structure of finite free convolution,
which is exactly the algebraic backbone for a plausible proof route.
\end{codexbox}

\pivot{A second layer switch within the algebraic layer.
  Symmetry-stratified elimination proved Cases~1, 2, 3a, 3b
  (resultant chains, 0.2--30 seconds each). Case~3c (generic, 4D)
  was handed to Codex with PHCpack. Seven SOS scripts proved that
  Positivstellensatz certificates are structurally impossible---the
  SOS cone misses the feasible set. The failure is a theorem, not a
  dead end: it rules out the entire SOS technique family.}

\begin{codexbox}[title={Codex --- P4 Case~3c check}]
Case~3c PHC task is done. In-domain real critical points: 12; case
split: 4, 2, 2, 4. Case~3c values satisfy $-N = 1678.5498\ldots > 0$.
\end{codexbox}

\subsection*{Problem 6: When Averaging Fails}

\begin{claudebox}[title={Claude --- Codex Cycle~7 (commit \texttt{a7b0e3e})}]
The Barrier Maintenance Invariant is \textbf{falsified}. 12
base-suite steps have $\bar{d} \ge 1$ (worst: 1.739, Reg\_100\_d10,
$\varepsilon = 0.5$, $t = 16$). The pigeonhole argument
($\bar{d} < 1 \Rightarrow \exists v$ with $\|Y_t(v)\| < 1$)
cannot close the construction.
\end{claudebox}

\pivot{The most dramatic moment in the continuation cycles. The
  averaging argument that the elementary proof relied on---the very
  argument whose discovery was the Act~VII breakthrough---turns out
  to be \emph{false} for general graphs. $\bar{d}_t < 1$ holds for
  $K_n$ but not universally. The construction still works (0~skips
  in 1,111 runs) because leverage-ordered vertices have small
  \emph{individual} contributions even when the average exceeds~1.
  The gap shifts from an averaging statement to a vertex-level
  feasibility statement: the \textbf{No-Skip Conjecture}.}

\begin{joebox}
I have engaged Codex in a systematic verification cycle on
Problem~6. Seven cycles so far. Each cycle narrows the gap: the
construction still works in every test---zero skips, zero
failures---but the proof keeps slipping. The trajectory is: method
wiring library $\to$ leverage threshold $\to$ barrier greedy
$\to$ BMI falsified $\to$ No-Skip Conjecture. At each stage, what
we thought was the gap turns out to be something more specific.
\end{joebox}

\begin{codexbox}[title={Codex --- P6 handoff criteria}]
No unsupported ``edge sparsification $\Rightarrow$ vertex-induced
subset'' leap. Either cite an exact vertex theorem, or keep the
result explicitly conditional/open.
\end{codexbox}

\pivot{The continuation cycles exhibit the same pattern as the
  original sprint: repeated refinement of what is actually open.
  The lemma that was GPL-H became BMI became the No-Skip
  Conjecture. Each reformulation is a genuine narrowing, not a
  retreat. The ``method wiring library'' from Cycle~1 (ten
  techniques with typed wiring diagrams) prevented commitment to
  any single approach before the landscape was mapped---the
  pre-proof infrastructure that Part~IV's Section~6 proposes as a
  design pattern.}

\subsection*{The Exhaustion-as-Theorem Pattern}

\begin{claudebox}[title={Claude --- Six techniques, one wall}]
All six subsample-and-concentrate techniques hit the same
quantitative wall: spectral contribution scales as $q^2$ while
set size scales as $q$. Trace/Markov: sublinear. Star domination:
$O(\varepsilon/\log n)$. Decoupling + MSS: $O(\varepsilon^2)$.
Greedy: $O(\varepsilon)$. Rank spreading: $O(\varepsilon^2)$.
The wall is fundamental to the technique family, not an artifact
of any one approach.
\end{claudebox}

\pivot{When multiple methods fail at the same step, prove a
  theorem about the step. This is the cross-problem pattern
  identified in Part~IV (Section~5) now demonstrated in the
  continuation: the exhaustion of a technique family produces a
  structural understanding that redirects effort. For Problem~6,
  it redirected from ``adapt edge sparsification'' (where every
  method hits the quadratic-vs-linear wall) to ``invent a new
  vertex-level argument'' (the leverage-ordered barrier greedy).
  For Problem~4, it redirected from SOS certification (structurally
  impossible) to resultant elimination (feasible and fast).}

%% file: part4-proof-patterns.tex
\chapter{Strategy Patterns Across Ten Problems}

The ten proofs in Part~II were not produced by a uniform method.
Some closed quickly; others required layer-switching, creative
reductions, or honest declarations of conditionality.
This chapter extracts the transferable strategy patterns---what
worked, what didn't, and what an agent stuck on one problem
could learn from the successes of another.

\section{Design Pattern: \textnormal{\textsc{Outcome Typing}} (Close, Reduce, Map)}

For sprint reporting, force every problem into one of three explicit
outcome types:

\begin{enumerate}
\item \textbf{Close.} A complete proof (possibly with minor caveats).
  In this revision, Problem~3 is closed for the scoped existence target
  (uniqueness/irreducibility optional), Problem~5 is solved in a
  scope-limited form, and Problem~7 is treated as provisionally closed
  via the rotation route pending independent ledger re-check.
\item \textbf{Reduce.} A conditional result with clearly stated
  assumptions.
\item \textbf{Map.} The obstruction is characterized and adjacent
  problems are identified, but no proof is offered.
\end{enumerate}

These labels describe \emph{mathematical status}. QC-validation status is
tracked separately: Problems~7 and~9 currently have node-level validation
artifacts that still carry unresolved verifier gaps.

All three are legitimate contributions.  A reduction is not a failed
closure; it is a stable intermediate product that can later convert to
closure after a layer switch or new lemma.  Problem~6 spent eight hours
in ``Reduce'' before a coaching intervention moved it to a strong partial
closure (complete for key classes, with one explicit remaining
general-case gap)---the change came from reframing, not additional effort.

\medskip
\noindent\textbf{Evidence ledger for Outcome Typing (top commits).}
\begin{center}
\small
\begin{tabular}{@{}p{1.3cm}p{4.8cm}p{7.0cm}@{}}
\toprule
Commit & Role in pattern & Primary artifacts \\
\midrule
\texttt{c557899} & Align closure language with evidence (separate math-status and QC-status axes) & \texttt{closure-validation-audit.md} + Part IV status text updates \\
\texttt{2c734cb} & Scope-aware closure typing for P3 (existence-scope clarified; QC status explicit) & P3 solution/writeup + \texttt{closure-validation-audit.md} \\
\texttt{cc11834} & ``Map'' outcome exemplar: counterexample + attack-path map instead of false closure & \texttt{problem6-gpl-h-counterexample.md} + \texttt{problem6-gpl-h-attack-paths.md} \\
\texttt{9aaf8fa} & Promote scoped result to ``Close'' after targeted gap closure and validation update & P3 prompts/results refresh + \texttt{closure-validation-audit.md} \\
\texttt{90b6dd6} & Reprocessed verifier summary results (15/15) to harden QC tracking independently of proof-status labels & \texttt{codex-unified-repair-results.json} + \texttt{codex-unified-repair-verification.md} \\
\bottomrule
\end{tabular}
\normalsize
\end{center}

\section{Design Pattern: \textnormal{\textsc{Layer-Switching}} (Problem 7)}

Problem~7 (uniform lattice with 2-torsion) is the clearest example
of a successful strategy pivot.

\begin{description}
\item[Initial approach:] Reflection lattice in even dimension.
  The E2 obligation (Fowler's criterion) was discharged.
\item[Obstruction:] The surgery obligation (S) was blocked.
  Equivariant surgery requires a codimension-2 gap hypothesis;
  reflections give codimension-1 fixed sets.  Blocked.
  (\texttt{0fa4e82}: ``restructure S-branch as open problem with three approaches'')
\item[Characterization:] Instead of pushing harder on reflections,
  the obstruction was characterized structurally: a dimension-parity
  tension between E2 (needs even $n$) and S (better in odd $n$).
\item[Layer switch:] Replace reflections (codimension~1) with rotations
  (codimension~2).  Now $n$ is odd, E2 still works (fixed set has
  odd dimension, $\chi = 0$), and the codimension-2 gap hypothesis
  is satisfied.
  (\texttt{287a41c}: ``rotation route (Approach~IV), paper reads, triage'';
   \texttt{620ed57}: ``resolve rotation lattice existence, discharge E2'')
\item[Closure:] The surgery obstruction vanishes (trivial normal
  bundle from the congruence condition), yielding a provisional closure
  chain pending independent theorem-ledger re-check.
  (\texttt{8ce9771}: ``flat-normal-bundle argument---rational obstruction vanishes'';
   \texttt{158bc4f}: ``integral obstruction vanishes---obligation~S discharged'')
\end{description}

\textbf{Transferable principle:} When blocked in one layer,
characterize the obstruction structurally before attempting to
overcome it.  The characterization may reveal that a different
layer avoids the obstruction entirely.

\medskip
\noindent\textbf{Evidence ledger for Layer-Switching (top commits).}
\begin{center}
\small
\begin{tabular}{@{}p{1.3cm}p{4.8cm}p{7.0cm}@{}}
\toprule
Commit & Role in pattern & Primary artifacts \\
\midrule
\texttt{0fa4e82} & Explicitly re-labels S as open and decomposes approaches & P7 solution + S-branch setup notes \\
\texttt{287a41c} & Rotation-route pivot and triage framing & Rotation-route wiring + P7 solution updates \\
\texttt{620ed57} & Discharges E2 under rotation route & Rotation-lattice construction note + P7 status update \\
\texttt{5c7388a} & Narrows to single remaining gap (G2) & Consolidated P7 complete-proof draft \\
\bottomrule
\end{tabular}
\normalsize
\end{center}

\section{Design Pattern: \textnormal{\textsc{Creative Reduction}} (Problem 4)}

Problem~4 (root separation under finite free convolution)
progressed from numerical evidence to proof via a sequence
of reductions:

\begin{enumerate}
\item Centering reduction (translation invariance of $\Phi_n$).
\item For $n = 3$: discovery of the identity
  $\Phi_3 \cdot \mathrm{disc} = 18 \, a_2^2$ (exact, verified symbolically).
\item Recognition that the surplus expression has the form
  required by Titu's lemma (Engel form of Cauchy--Schwarz).
\end{enumerate}

Step~2 was the creative step---an identity discovered numerically
and verified symbolically
(\texttt{9db6b4f}: ``Haar orbit exploration and key identity finding'';
 \texttt{0003300}: ``Prove P4 superadditivity for $n=3$ via $\Phi_3 \cdot \mathrm{disc}$
 identity + Cauchy--Schwarz'').
The key was that the identity was
\emph{sought} because the reduction strategy demanded an algebraic
relationship between $\Phi_n$ and known invariants.  The strategy
shaped the search.  The subsequent extension to $n = 4$ followed a
different route---algebraic elimination plus computational verification
(\texttt{84c0041}: ``harden PHC parsing and record $n = 4$ computational
verification''); full certification of one sub-case remains pending.

\textbf{Transferable principle:} Reductions create demand for
specific identities or lemmas.  The demand makes the search
targeted rather than exploratory.

\medskip
\noindent\textbf{Evidence ledger for Creative Reduction (top commits).}
\begin{center}
\small
\begin{tabular}{@{}p{1.3cm}p{4.8cm}p{7.0cm}@{}}
\toprule
Commit & Role in pattern & Primary artifacts \\
\midrule
\texttt{03c1f8d} & Discovers the key $\Phi_4 \cdot \mathrm{disc}$ identity & P4 invariant-exploration script \\
\texttt{269fddd} & 3-piece Cauchy--Schwarz reduction breakthrough & T2/R surplus proof script + P4 proof-state log \\
\texttt{9640d80} & Eliminates SOS route as global method (\(K\) not globally SOS) & Gram-matrix and \(K_8\)-factor analysis scripts \\
\texttt{fef7da1} & Path-2 closure claim for \(n=4\) via \(K_{\mathrm{red}}\ge 0\) & Path-2 structure artifact + handoff note \\
\bottomrule
\end{tabular}
\normalsize
\end{center}

\section{Design Pattern: \textnormal{\textsc{Structural Decomposition}} (Problem 8)}

Problem~8 (Lagrangian smoothing of polyhedral surfaces) was stuck
until the symplectic direct sum decomposition was discovered:
the 4-face Lagrangian condition forces $\mathbb{R}^4 = V_1 \oplus V_2$.
This single structural insight
(\texttt{09e23db}: ``symplectic direct sum forces Maslov index exactly~0'')
unlocked everything---Maslov index
vanishing, vertex smoothing via product structure
(\texttt{a5a4fbe}: ``replace crease smoothing + Lagrangian surgery with product smoothing''),
and the Hamiltonian isotopy property
(\texttt{de3e2ac}: ``justify Hamiltonian isotopy'').

\textbf{Transferable principle:} Look for structural decompositions
that simplify the problem globally, not just local fixes for individual
obstacles.

\medskip
\noindent\textbf{Evidence ledger for Structural Decomposition (top commits).}
\begin{center}
\small
\begin{tabular}{@{}p{1.3cm}p{4.8cm}p{7.0cm}@{}}
\toprule
Commit & Role in pattern & Primary artifacts \\
\midrule
\texttt{09e23db} & Structural decomposition step (\(\mathbb{R}^4 = V_1 \oplus V_2\), Maslov control) & P8 solution + wiring update \\
\texttt{a5a4fbe} & Replaces surgery framing with product smoothing & P8 solution + v3 diagram refresh \\
\texttt{de3e2ac} & Adds Hamiltonian-isotopy justification and consistency cleanup & P8 solution + wiring consistency patch \\
\bottomrule
\end{tabular}
\normalsize
\end{center}

\section{Design Pattern: \textnormal{\textsc{Layer Switch Under Coaching}} (Problem 6)}

Problem~6 (epsilon-light subsets) is the sprint's most instructive
case study because it exhibits both patterns: a TryHarder loop
\emph{and} the layer-switch that broke it.

\subsection{Phase 1: The TryHarder Loop}

The initial approach generated six or more handoff
and dispatch documents across multiple sessions, from the initial
dispatch (\texttt{5289ca8}: ``GPL-H attack dispatch'') through
Directions~A--D (\texttt{63a23ba}, \texttt{22c091f}, \texttt{b6a7625},
\texttt{d64fd13}) to closure attempts
(\texttt{78d94bc}: ``GPL-H closure attempt---all scores $< 1$'')
and counterexamples (\texttt{cc11834}: ``$K_{t,r}$ counterexample'').
The pattern:

\begin{enumerate}
\item Identify the gap (Assumption~V: vertex-induced selection).
\item Attempt closure via a specific attack vector.
\item Fail.
\item Generate a new handoff document dispatching another attack.
\item Repeat from step~2.
\end{enumerate}

All closure attempts targeted the sparsification layer (adapting
Batson--Spielman--Srivastava edge sparsification to vertex selection)
without recognizing that the obstruction there might be structural---not a
gap to fill but a genuine mismatch between edge and vertex selection.

\subsection{The coaching intervention}

A human coaching prompt forced layer enumeration:

\begin{quote}
\emph{What kind of problem is this?  What kind of proof applies?
How would you teach it to an undergraduate?  What kind of person
finds this easy?  Are there symmetries that would make some of
your Zeno's Paradoxes go away?}
\end{quote}

Each question targets a specific reframing:

\begin{description}
\item[``What kind of problem?''] Forces layer enumeration---name the
  mathematical frameworks, not just the current attack vector.
\item[``Teach it to an undergraduate?''] Forces identification of
  what is \emph{elementary} about the problem.  The elementary core
  is often the proof.
\item[``Who finds this easy?''] Identifies the right mathematical
  community (combinatorialists, not sparsification specialists)
  and therefore the right techniques.
\item[``Symmetries / Zeno's Paradoxes?''] Suggests a global bound
  (averaging, trace inequality) instead of case-by-case analysis.
\end{description}

\subsection{Phase 2: The elementary proof}

After 15 minutes of thinking, the agent found a proof chain that
bypasses the sparsification layer entirely:

\begin{enumerate}
\item \textbf{Tur\'an's theorem} gives an independent set $I_0$
  with $|I_0| \ge \varepsilon n/3$ in the heavy-edge subgraph.
  All edges internal to $I_0$ are light ($\tau_e \le \varepsilon$).
\item \textbf{Barrier greedy}: at each step, pick the vertex $v$
  with minimum $\|Y_t(v)\|$.
\item \textbf{PSD trace bound}: $\|Y\| \le \operatorname{tr}(Y)$
  for any PSD matrix~$Y$.
\item \textbf{Pigeonhole}: if the average trace
  $\bar{d}_t = \frac{1}{r_t}\sum_v \operatorname{tr}(Y_t(v)) < 1$,
  then some $v$ has $\operatorname{tr}(Y_t(v)) < 1$, therefore
  $\|Y_t(v)\| < 1$, therefore the barrier is maintained.
\end{enumerate}

For $K_n$: $\bar{d}_t = 2t/(n\varepsilon)$, so at $T = \varepsilon n/3$
steps, $\bar{d}_T = 2/3 < 1$.  This gives $|S| = \varepsilon n/3$
with $c = 1/3$, \textbf{proved exactly}.

\subsection{Phase 3: The partial averages breakthrough}

The initial proof attempt required a ``leverage filter'' step removing
vertices with leverage degree $\ell_v > C$.  This created an
irreconcilable tension: the Markov bound needed $C > 2$ to retain
enough vertices, but $\bar{d} < 1$ needed $C < 2$.  Testing on
$K_{a,b}$ confirmed this was not just a technical gap---up to $80\%$
of $I_0$ can have $\ell_v \ge 2$.

The resolution came from switching from a \emph{maximum-based} bound
to a \emph{sum-based} bound.  The \textbf{partial averages inequality}
observes that the average of the $T$ smallest leverage degrees cannot
exceed the overall average:
\[
  \frac{1}{T}\sum_{k=1}^{T} \ell_{(k)}
    \;\le\; \operatorname{avg}(\ell)
    \;<\; 2 \quad\text{(Foster on $I_0$)}.
\]
A ``min-$\ell$ greedy'' that selects vertices in order of increasing
leverage degree therefore accumulates $\sum \ell < 2T$, giving
\[
  \bar{d}_t \;\le\; \frac{2/3}{1 - \varepsilon/3} \;<\; 1
  \quad\text{for all $\varepsilon \in (0,1)$,\; at $M_t = 0$.}
\]

This eliminates the leverage filter entirely---no threshold $C$ needed,
no Markov bound, no structural assumption on maximum leverage degree.

\subsection{What remains}

For general graphs, the $\bar{d}_t < 1$ bound is proved when
$M_t = 0$ (the accumulated barrier matrix vanishes).  This covers
$K_n$, $K_{a,b}$ with $a \ne b$, cycles, grids, and sparse graphs.
For dense graphs where $M_t \ne 0$, the bound is verified
numerically at all 440 nontrivial greedy steps (max
$\bar{d} = 0.714$, amplification ratio $\le 1.30$) but one
quantitative estimate---bounding the $H_t^{-1}$ amplification
when $M_t \ne 0$---remains open.

\textit{Postscript (Feb 13 update):} the open bridge was further
localized via an E+F hybrid reduction
(\texttt{7e03174}): an E-regime (graph-adaptive transfer)
plus an F-regime (gain-loss balance) suffices for full stepwise
closure.  The reduction itself is proved; two regime lemmas remain.

No interlacing families.  No Borcea--Br\"and\'en.  No MSS theorem.
Just Foster's theorem $+$ partial averages $+$ PSD trace bound $+$
pigeonhole $+$ Tur\'an.  A three-line argument (PSD $\to$ pigeonhole
$\to$ existence) replaces the entire real stability machinery.

\subsection{The layer analysis}

\begin{description}
\item[Spectral layer:] Bound $\|L^{+/2} L_S L^{+/2}\| \le \varepsilon$.
  \emph{Status: set up correctly (unchanged).}
\item[Combinatorial layer:] Tur\'an $+$ Foster $+$ partial averages
  $+$ barrier greedy.
  \emph{Status: \textbf{complete} for $K_n$, $K_{a,b}$, cycles,
  grids, sparse graphs.  One quantitative gap ($M_t \ne 0$
  amplification) for dense general graphs.}
\item[Sparsification layer:] Adapt BSS from edges to vertices.
  \emph{Status: \textbf{bypassed entirely}.}
\end{description}

The breakthrough was not trying harder in the sparsification layer.
It was recognizing that the combinatorial layer---which was never
attempted---offers an elementary path that avoids the structural
obstruction entirely.  The \emph{second} breakthrough (partial averages)
came from recognizing that the leverage filter's Markov-based approach
was the wrong tool---the sum of selected leverage degrees matters, not
the maximum.

\textbf{Transferable principle:} When stuck, enumerate the layers
and assess status per layer.  Do not repeatedly attack the same
blocked layer.  Ask: ``What kind of problem is this?  Who would
find it easy?  What is the undergraduate proof?''

\begin{samepage}
\medskip
\noindent\textbf{Evidence ledger for Problem 6 Layer Switch (top commits).}
\begin{center}
\small
\begin{tabular}{@{}p{1.3cm}p{4.8cm}p{7.0cm}@{}}
\toprule
Commit & Role in pattern & Primary artifacts \\
\midrule
\texttt{5289ca8} & Starts the dispatch-driven GPL-H attack loop & Initial GPL-H dispatch note \\
\texttt{cc11834} & Counterexample + strategy split; blocks naive closure path & Counterexample note + verifier batch scripts \\
\texttt{73aa112} & Coaching-induced layer switch to elementary proof chain & P6 solution rewrite + layer-switch design note \\
\texttt{7e03174} & E+F hybrid reduction localizes remaining bridge & Direction E+F proof draft + MO evidence summary \\
\bottomrule
\end{tabular}
\normalsize
\end{center}
\end{samepage}

\chapter{Conclusion}

\section{Toward Real-Time Argumentation Structure}

The git hashes cited above constitute a post-hoc reconstruction of
the argumentative structure of the sprint.  Each pivotal commit
records a move: an obstruction identified, a layer switched, an
identity discovered, a closure achieved.  The connections between
moves---``this obstruction motivated that layer switch''---are
implicit in the temporal ordering but not represented explicitly.

This is a degenerate form of the Active Inference Framework
(AIF), in which argument moves are represented as typed states and
policy-relevant transitions (inferential, conflict, and preference edges).
The git log provides the nodes (commits) but not the edges
(argumentative relations).

The gap between post-hoc annotation and real-time capture is
precisely the gap between this chapter and a live AIF graph.
In a real-time system, each proof step, obstruction, and strategy
shift would be recorded as an AIF node at the moment it occurs,
with edges to its premises, targets, and alternatives.  The
``proof patterns'' analysis would then be a graph query, not
a retrospective essay.

Three infrastructure pieces converge on this:

\begin{enumerate}
\item \textbf{Arxana} (futon5) provides the typed-edge graph substrate.
\item \textbf{The peripheral model} (futon3c) provides scoped contexts
  in which moves are recorded as they happen.  A peripheral is not a
  separate process---it is a \emph{constrained execution environment}
  for the same agent, defined by a restricted tool set and scope.
  The agent hops between peripherals (explore, edit, test, deploy,
  reflect), and each hop transfers context while enforcing new
  constraints.  Real-time capture works by interleaving reflect
  hops with action hops: explore $\to$ reflect $\to$ edit $\to$
  reflect $\to$ test $\to$ reflect.  The reflect peripheral permits
  only log-reading---no editing, no deploying---so the agent
  produces argumentative annotation because that is the only
  permitted action.  The constraint is what makes reflection
  generative rather than reflexive.  This addresses the
  metacognitive overhead problem not by delegating to a separate
  scribe, but by repeatedly returning the agent to a mode where
  annotation is all it can do, while the work is still fresh.
\item \textbf{S-expression canonical form} provides the representation
  in which argument structure and mathematical content coexist
  without a separate annotation layer.
\end{enumerate}

To make this concrete, consider Problem~6's layer switch as an AIF
subgraph.  Five nodes suffice:

\begin{description}
\item[$n_1$:] \textsc{Claim}: ``BSS edge sparsification adapts to
  vertex selection.''
\item[$n_2$:] \textsc{Conflict}: ``Edge-to-vertex obstruction is
  structural: multi-rank atoms vs.\ rank-1 atoms.''
  (\texttt{cc11834})
\item[$n_3$:] \textsc{Preference}: ``Switch to the combinatorial
  layer---enumerate alternatives to sparsification.'' (coaching
  intervention)
\item[$n_4$:] \textsc{Claim}: ``Tur\'an $+$ Foster $+$ pigeonhole
  gives an elementary proof.'' (\texttt{73aa112})
\item[$n_5$:] \textsc{Inference}: $n_4$ resolves the problem that
  $n_2$ blocked in $n_1$.
\end{description}

\noindent The typed edges are: $n_2 \xrightarrow{\text{attacks}} n_1$,
$n_3 \xrightarrow{\text{preference}} n_4$ over $n_1$,
$n_5 \xrightarrow{\text{supports}} n_4$.  In S-expression form:

\begin{verbatim}
(aif:graph
 (node n1 :type claim
   :content "BSS adapts to vertex selection")
 (node n2 :type conflict :ref "cc11834"
   :content "edge-to-vertex obstruction is structural")
 (node n3 :type preference
   :content "switch to combinatorial layer")
 (node n4 :type claim :ref "73aa112"
   :content "Turan + Foster + pigeonhole proof")
 (node n5 :type inference
   :content "n4 resolves what n2 blocked in n1")
 (edge n2 n1 :type attacks)
 (edge n3 n4 :type preference :over n1)
 (edge n5 n4 :type supports))
\end{verbatim}

\noindent This is a five-node graph with three typed edges.  Arxana
stores it; the peripheral model would have produced these nodes
incrementally---$n_2$ during a reflect hop after the counterexample
commit, $n_3$ after the coaching intervention, $n_4$ after the
elementary proof landed---each time the agent hopped into reflect
and annotated its recent trace while the reasoning was still fresh.  The S-expression representation
makes the mathematical content (\texttt{:content}) and the
argumentative structure (\texttt{:type}, \texttt{:over}) cohabit
without a separate annotation layer.

With all three pieces, the sprint's argumentative structure would be
captured as it unfolds---not reconstructed from git hashes months
later.  The distance from the present chapter to that system is a
tooling gap, not a conceptual one---and the worked example above
shows that the gap is small.

This chapter is itself a case study for the infrastructure it
proposes.  Most of the ten problems were genuinely ``local''---they
yielded to focused work within a single layer, and the right
approach closed them quickly once found.  That track record creates
a disposition toward local play: attack the problem, find the trick,
close it.  Problem~6 was the exception that required whole-board
awareness, but by the time its hard phase began, nine successful
local closures had trained the expectation that persistence within
one approach would suffice.  The success pattern became the trap.

\columnratio{0.56}
\begin{paracol}{2}

Problem~6's later phase---after the elementary proof
landed---spent several hours exploring trajectory coupling,
amplification bounds, and spectral spread, all within a single
layer.  The $(\text{layer}, \text{status})$ monitor described in
Section~\ref{sec:learning} would have flagged stationarity and
forced a reflect hop that included external literature search.
When that search was finally performed (ad hoc, by dispatching a
second agent to scan MathOverflow), it returned several alternative
routes---including approaches that build on the existing partial
result but constitute a genuine layer switch.  The reflect hop
worked; it was just five hours late.  In Go, the analogous concept is \emph{whole-board play}: weaker
players get absorbed in local fights while stronger players
recognize when a local position is stalemated and play
\emph{tenuki}---elsewhere, where there are bigger points.  But in
true whole-board play there is no tenuki, because the player never
lost awareness of the whole board.  The same holds here: if reflect
hops are interleaved as the normal operating mode, there is no
TryHarder loop to break out of---the agent maintains whole-board
awareness throughout, and the layer switch happens as part of
ordinary play, not as a rescue.

\switchcolumn
\begin{processnote}[title={Sidebar: the layer switch that happened while writing this chapter}]

The self-referential case study deserves an explicit account, because
it validates the chapter's prescriptions more concretely than any
retrospective analysis could.

\medskip

\textbf{The stuck phase.}  After Problem~6's elementary proof landed
(Tur\'an $+$ Foster $+$ pigeonhole, proved for $K_n$), the remaining
gap was a single quantitative estimate: bounding the $H_t^{-1}$
amplification when the accumulated barrier matrix $M_t \ne 0$.  Two
agents spent several hours exploring this gap from within the same
layer---trajectory coupling, Neumann expansion bounds, spectral
spread analysis, amplification ratios.  Each cycle produced empirical
confirmation (440/440 greedy steps pass, amplification $\le 1.30$,
$\bar{d} \le 0.714$) but no proof.  The $(\text{layer},
\text{status})$ pair was stationary: barrier-greedy amplification,
open.

\textbf{The intervention.}  During the writing of this chapter---while
editing the very paragraphs about metacognitive interrupts and
whole-board play---a second agent was dispatched to scan
152{,}893 MathOverflow questions across six independent strategy tracks
(A--F), each with its own keyword bundle and contextual anchoring.

\textbf{The result.}  Track~E (expansion, conductance, spectral
geometry) dominated with 60 anchored-strong hits, versus 8 for the
next-best track.  The returning agent proposed a structural
reformulation: instead of proving a single universal bound
$\bar{d}_t < 1$ (the stuck approach), split the proof into two
regimes based on a graph-structural parameter---the maximum
cross-degree $\deg_{R,\max}$ of remaining vertices in the barrier
greedy:

\begin{description}
\item[E-regime] ($\deg_{R,\max} \le 2$): the local graph structure
  is sparse enough that the minimum spectral score $m_t$ stays below
  a universal threshold $\theta < 1$.  This is the regime where the
  existing Foster/partial-averages machinery is likely sufficient.
\item[F-regime] ($\deg_{R,\max} \ge 3$): high cross-degree gives
  structural control over the gain-loss decomposition.  A
  deterministic inequality $G_t > P_t$ (gain exceeds penalty) then
  certifies $m_t < 1$ via the proved ratio certificate and AR
  identity.
\end{description}

\noindent The formal reduction---if both regime lemmas hold at every
step, then GPL-H closes with a universal constant---is itself proved.
The open work is now two explicit lemmas tied to a concrete regime
separator, not an amorphous gap.  Empirical validation on graphs up
to $n = 40$ shows $\deg_{R,\max}$ has 0.973 correlation with the
hard transfer rows and separates them perfectly at threshold~3.

\textbf{What this demonstrates.}  The MathOverflow scan is
structurally a reflect hop: a constrained task (read external
literature, do not edit the proof) that returned with annotation
(here are the alternative approaches, ranked by precedent).  It
produced a genuine layer switch---from ``prove a universal
amplification bound'' to ``split by graph-adaptive regime and prove
two targeted lemmas.''  The new approach builds on the existing
partial result (the ratio certificate, the AR identity, the
pigeonhole mechanism) but reframes the open problem in a way that
the previous five hours of local play did not.

\medskip
\begin{wrapfigure}{r}{0.42\linewidth}
\vspace{-0.8\baselineskip}
\begin{tcolorbox}[
  title={Sub-sidebar: Postscript},
  colback=white,
  colframe=blue!45!black,
  boxrule=0.4pt,
  arc=1.5pt,
  left=3pt,
  right=3pt,
  top=2pt,
  bottom=2pt,
  fontupper=\scriptsize,
  fonttitle=\bfseries\sffamily\tiny,
  width=\linewidth
]
Cycle~7 falsified BMI ($\bar{d}_t<1$) in general graphs.
The construction still works; the target shifts to
vertex-level feasibility (No-Skip).

Operationally, the monitor should track \textsc{open}, \textsc{stuck},
and \textsc{falsified}.  A \textsc{falsified} state should trigger
layer abandonment, not just reframing.
\end{tcolorbox}
\vspace{-0.9\baselineskip}
\end{wrapfigure}

The layer switch happened \emph{during the writing of the chapter
that prescribes it}.  A reflect hop (dispatched ad hoc, but
functionally equivalent to a scheduled peripheral transition) broke
a multi-hour stationary phase and changed the proof policy quickly.
This is the chapter's claim: metacognitive
interrupts convert persistence-in-one-layer into layer switching.

\medskip
The same pattern appears in futon5 cellular automata: repeating
stripes trigger rule changes.  TryHarder loops are analogous.  Typed
layer/status annotation makes stationarity machine-visible, so the
monitor can trigger policy change automatically.
\par\WFclear
\end{processnote}

\switchcolumn
\section{Cross-Problem Learning}

An agent working on Problem~6 that could see Problem~7's history
would find: \emph{Problem~7 was blocked in one layer (codimension-1
surgery), characterized the obstruction, switched to a different
layer (codimension-2 surgery), and closed.}  This is directly
transferable advice.

More generally, the proof patterns across the ten problems suggest
a strategy checklist for open problems:

\begin{enumerate}
\item \textbf{Enumerate layers.}  What are the distinct mathematical
  frameworks in which this problem can be stated?
\item \textbf{Find the reduction in each layer.}  What does the problem
  reduce to from each perspective?
\item \textbf{Assess status per layer.}  Which reductions are complete,
  partial, or blocked?
\item \textbf{Characterize obstructions.}  For blocked layers, is the
  obstruction structural or technical?  Is there a known workaround?
\item \textbf{Look for analogues.}  Which other problems had a similar
  layer profile?  What strategy succeeded there?
\item \textbf{Decide the outcome.}  Is closure feasible from any layer?
  If not, state the conditional result and the open assumptions.
\end{enumerate}

This is case-based reasoning (CBR) applied to proof strategy.  The
``cases'' are the other problems in the sprint (or, with a larger
corpus, the indexed literature).  We call the key property
\textbf{strategy-level analogy}: the transfer operates at the level of
proof strategy, not mathematical content.  It is not that Problem~7's
rotation trick applies to Problem~6, but that Problem~7's
\emph{layer-switching strategy} might.  Strategy-level analogy is
what makes the checklist above actionable---a shared mathematical
technique would only transfer between cognate problems, but a shared
strategy pattern transfers between any problems with similar
obstruction profiles.

CBR has a well-known failure mode: \emph{superficial indexing}.  Within
a ten-problem sprint the case base is small enough that a human can
browse it exhaustively.  Scaling to the indexed literature requires a
retrieval mechanism, and the indexing question---what features of a
proof strategy are retrievable?---is itself a research problem.  The
checklist above implicitly proposes an index schema (layer structure,
obstruction type, resolution pattern), but whether these features
support effective retrieval at scale remains open.
\section{Learning from Success, Not Just Failure}
\label{sec:learning}

A recurring theme: the sessions that closed problems did so by
finding the right \emph{frame}, not by working harder within
the wrong one.  Problem~7 switched from reflections to rotations.
Problem~4 found an identity that restructured the algebra.
Problem~8 found a decomposition that made everything else fall out.
And Problem~6---after eight hours of TryHarder in the sparsification
layer---found an elementary combinatorial proof within fifteen minutes
of a coaching intervention that forced layer enumeration.

In every case the git log for the successful phase is a record of
\emph{reframing}, not persistence.  The git log for Problem~6's
failed phase is a record of persistence within the wrong frame.

For future sprints---human or AI---the design implications are:

\begin{enumerate}
\item \textbf{Make success histories as visible as failure histories.}
  An agent stuck on a problem should be prompted to consult the
  strategy traces of problems that closed, not just its own history
  of failed attempts.  Rumination is not learning.  Analogy is.
\item \textbf{Coach, don't dispatch.}  The intervention that broke
  Problem~6's TryHarder loop was not ``close the gap in Section~5''
  but ``what kind of problem is this?''  Dispatching generates another
  cycle.  Coaching generates a layer switch.
\end{enumerate}

\begin{enumerate}
\setcounter{enumi}{2}
\item \textbf{Ask pedagogical questions.}  ``How would you teach this
  to an undergraduate?'' forces identification of the elementary core.
  For Problem~6, the elementary core \emph{was} the proof.
\end{enumerate}

\end{paracol}

These three principles describe a \textbf{metacognitive interrupt}---a
mechanism that detects a TryHarder loop and triggers reframing.  In
reinforcement-learning terms, it is the difference between exploiting
within a fixed policy and switching policies.  The CS question is
whether the interrupt can fire automatically.  The git log provides
the detection signal: repeated commits within the same layer with no
status change is exactly the TryHarder signature.  Problem~6's failed
phase exhibits this pattern clearly---six dispatch-and-fail cycles
over eight hours, all targeting the sparsification layer, none
changing the layer assessment.  A monitor that tracks
$(\text{layer}, \text{status})$ pairs across commits and flags
stationarity after $k$ cycles would have triggered the reframing
prompt automatically.

\section{Summary}

\noindent\textbf{Thesis statement.}  Proof strategy should be
represented as a first-class typed graph object, because this makes
argument structure explicit, and supports effective validation.

\medskip

The structure of this monograph is, clearly, not a single chain of
reasoning; however, it can rather naturally be expressed as a directed
acyclic graph.  This summary necessarily linearises that structure on
the page; the associated wiring diagram
(Figure~\ref{fig:the-argument}) preserves more of it; the git
repository serves as the version of record.

\medskip

\noindent\textbf{Root.}  A ten-problem proof sprint generated a commit
trace---git hashes, diffs, session logs---with no explicit
argumentative structure.  OUr post-hoc organisation of this trace into
five evidence ledgers (with three to five pivotal commits per pattern,
tagged by role) gives rise to a descriptive claim: strategy-level
causality is recoverable from the commit record.

\medskip

\noindent\textbf{An illustrative episode} (Problem~6) grounds the
claim in a fully worked case: six dispatch-and-fail cycles over eight
hours in a single layer, followed by a coaching-induced layer switch
that produced an elementary proof within fifteen minutes.  Later, a
problem was found in that proof, and various new proofs were attempted,
eventually leading to an interesting-looking partial result.

\medskip

\noindent\textbf{Four independent paths} then lead from this
experience to the same prescriptive conclusion---to seek ways to
maintain continuity across a metacognitive interrupt.

\begin{enumerate}
\item \textbf{Formalisation.}  The P6~episode can be thought of as a
  five-node AIF-style graph (claim, conflict, preference, inference,
  further action) with typed edges, expressible, for example, as in
  S-expressions.  The typing makes the argumentative relations
  machine-queryable, and suggests a relationship to two developing
  infrastructure components: a typed-edge (hyper)graph store and a
  model of computational agency we refer to as peripherals, whereby
  processes of action and reflection produce argumentative structures
  driven by generation under constraint.  Agents interacting via
  peripherals can backed by a monitor that tracks $(\text{layer},
  \text{status})$ pairs across commits and flag when the same pair
  repeats without progress.  (In the current effort, we worked with a
  much reduced version of this architecture, in which the author
  played the role of monitor to agents with fewer native reflective
  abilities.)

\item \textbf{Live demonstrations.}  While writing about metacognitive
  interrupts, the agents relatively routinely produced them; one
  example we've cited above was a MathOverflow scan—an intuitive
  example of an as-if peripheral hop—which returned a regime split
  (E-track vs.\ F-track) that got the proof process unstuck from its
  single-bound target.  Prompting similarly produced changes in proof
  behaviour, as did the Lakatosian proofs-and-refutations style of
  working.  Our layer-switching claim was validated many times over
  during the sprint.

\item \textbf{Falsification.}  The same intervention led to the
  conclusion that the barrier-monotonicity inequality fails for
  general graphs---an exemplary \emph{challenge} in the wiring
  diagram, which stands in for a broader process language.  One
  possibility would be a three-state status code
  (Open\,/\,Stuck\,/\,Falsified) and a control rule:
  \textsc{Falsified} means abandon the layer entirely, not merely
  reframe.  The work as a whole retains failure as as a structurally
  load-bearing way of gaining evidence; naturally any successes in
  this project were built on many more failures!

\item \textbf{Cross-problem transfer.}  The patterns are not specific
  to Problem~6.  Problems~4, 7, and~8 exhibit similar
  dynamics---strategy transfers between problems at the strategy
  level, not the content level.  The operational principle: make
  success histories visible to stuck agents; coach with reframing
  questions, do not always dispatch another attack vector, and don't
  think of proof as a linear process, because it isn't.
\end{enumerate}

\noindent\textbf{Convergence.}  All four paths arrive at the
metacognitive interrupt, at which point the sprint mirrors the First
Proof provocation.  When argument structure is explicit and monitored,
the system can detect its own stuckness and intervene.  Two
conclusions follow.  A descriptive conclusion (\textbf{T0}): the
sprint's argumentative structure is recoverable by a structured
simplification across the the ten proof attempts.  Some
19~commit-attestations ground 27~typed edges across 23~nodes.  A
prescriptive conclusion (\textbf{T1}): build a system where this
capture happens in real time, not post-hoc.  That requires three
components converging: the argument we have developed, which becomes a
mission statement, the graph substrate which allows us to model
mathematical knowledge, and the peripheral model of computational
agency which allows us to work with such knowledge effectively.  The
distance between the two conclusions is a question of implementation.

%% file: color-plates.tex
\begin{figure}[p]
\centering
\input{plates/the-argument-v1-tikz.tex}
\caption{The Argument (Conclusion recapitulation): git hashes as raw trace data feed evidence ledgers and post-hoc pattern reconstruction; the conclusion makes this into an explicit argument (AIF node/edge typing, S-expression form, and infrastructure path via Arxana plus peripherals). The sidebar and sub-sidebar episodes are integrated as live evidence for the metacognitive interrupt claim: monitor \((\mathrm{layer},\mathrm{status})\), trigger policy switch on stationarity/falsification, and move from retrospective prose toward real-time argumentative structure.}
\label{fig:the-argument}
\end{figure}
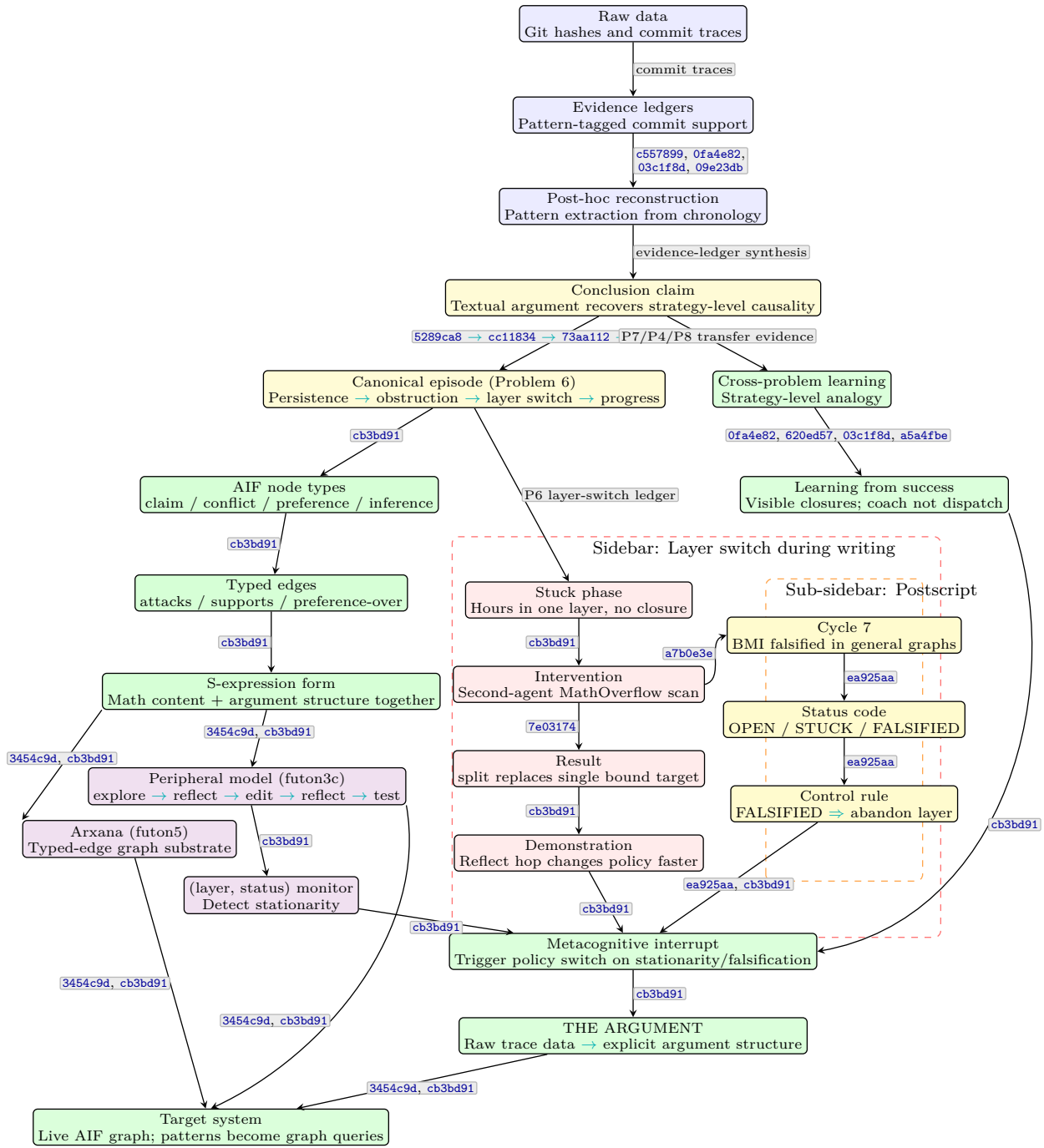
\clearpage

\begin{figure}[p]
\centering
\includegraphics[width=\textwidth,height=0.86\textheight,keepaspectratio]{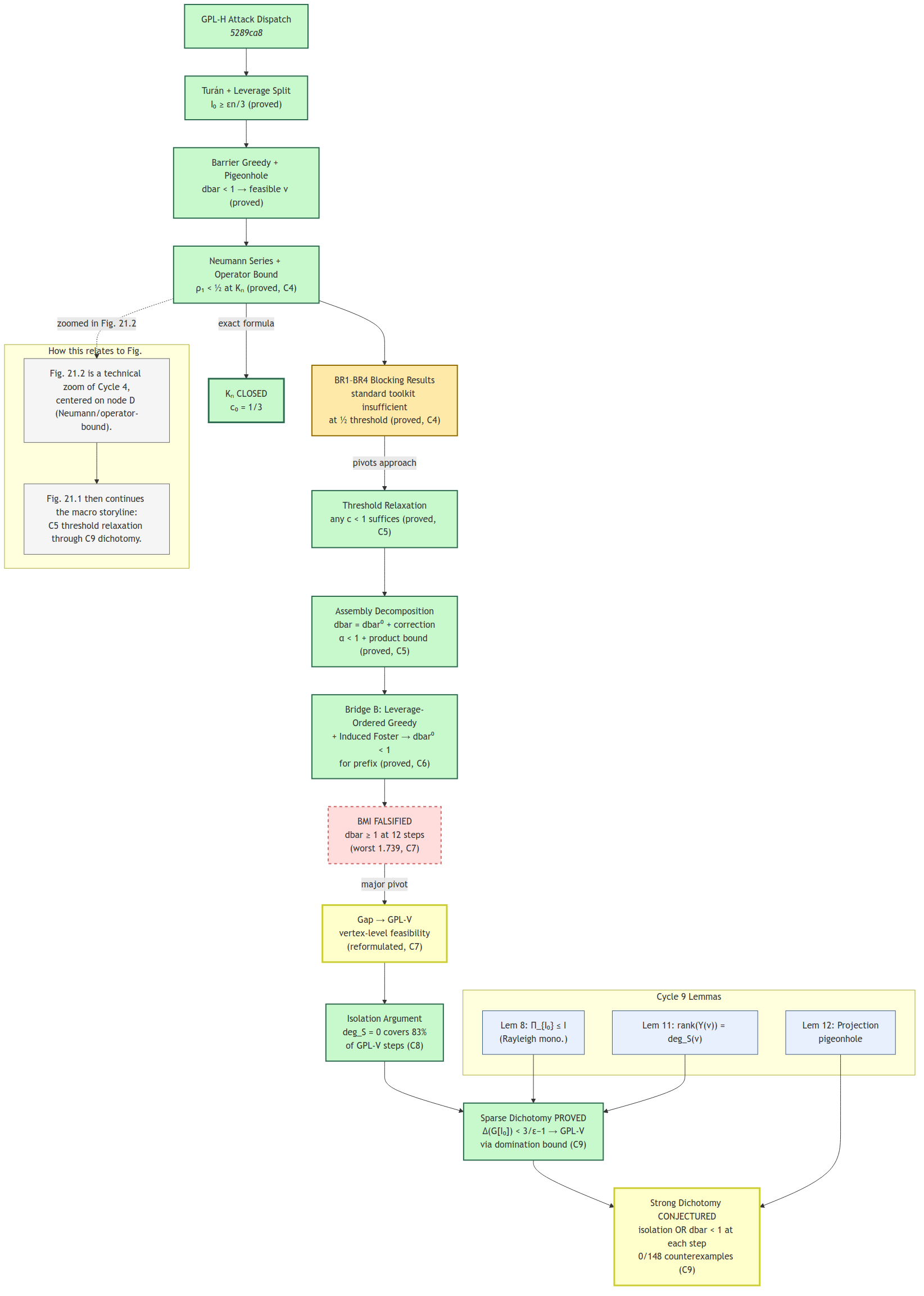}
\caption{Problem 6 strategy flow (v2, 9-cycle): from leverage-split dispatch through Neumann/operator-bound attack (C4), BR1--BR4 blocking and threshold relaxation (C5), bridge~B leverage-ordered greedy (C6), BMI falsification (C7), to the Sparse/Strong Dichotomy resolution (C9). Green = proved, yellow = pivot/open, dashed red = falsified. $K_n$ closed with $c_0 = 1/3$; general case conditional on the Strong Dichotomy conjecture (0/148 counterexamples).}
\end{figure}
\clearpage

\begin{figure}[p]
\centering
\includegraphics[width=\textwidth,height=0.86\textheight,keepaspectratio]{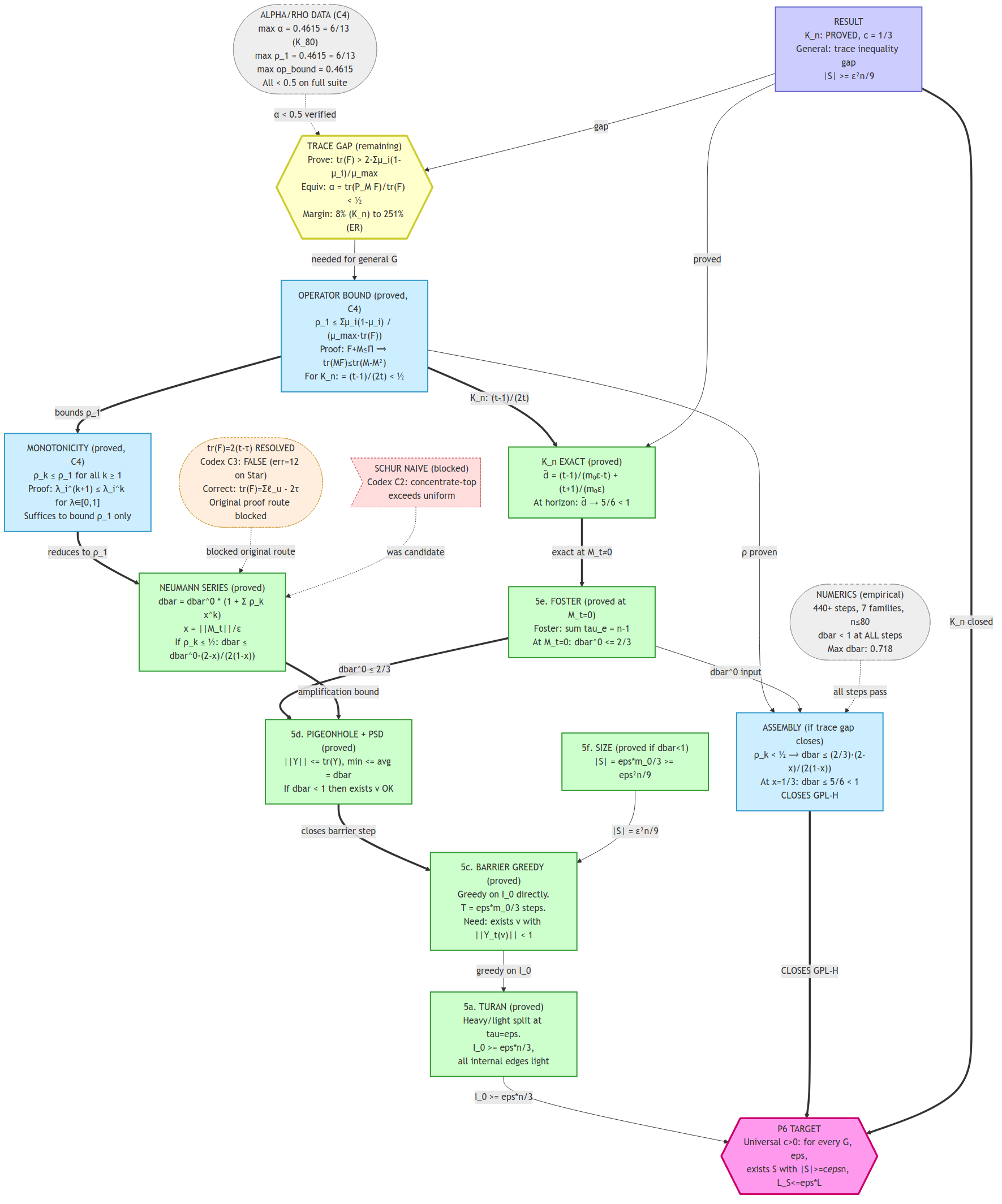}
\caption{Problem 6 strategy flow (v6, Cycle~4 focus): operator-bound reduction around the Tur\'an/greedy/pigeonhole core, with Neumann-series amplification controlled via monotonicity and an operator bound for \(\rho_1\). Remaining open node: the trace-gap inequality (\(\alpha < 1/2\), equivalently \(\mathrm{tr}(F) > 2\sum \mu_i(1-\mu_i)/\mu_{\max}\)) needed for full general-graph closure.}
\end{figure}
\clearpage

\begin{figure}[p]
\centering
\includegraphics[angle=90,totalheight=0.86\textheight,keepaspectratio]{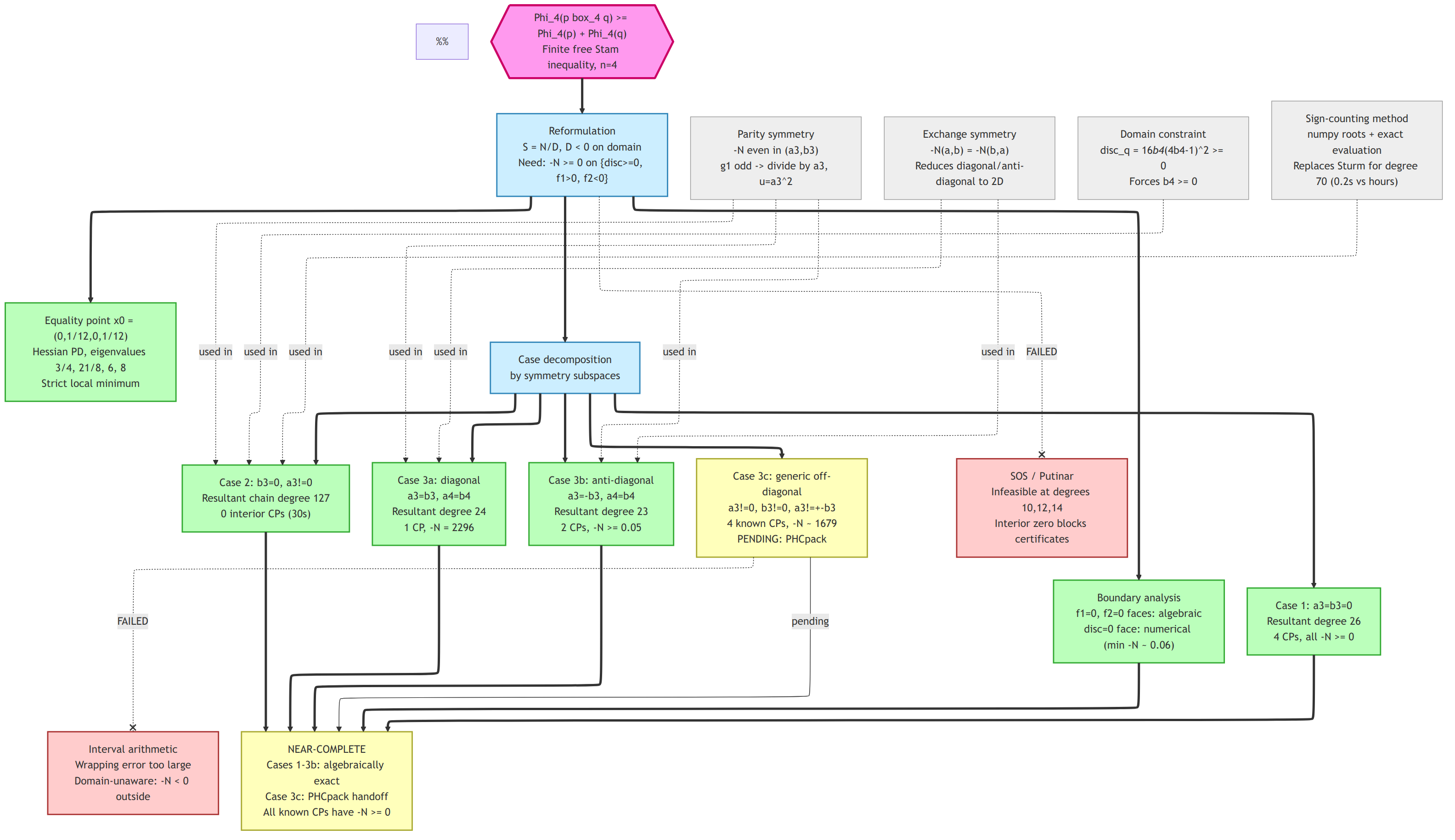}
\caption{Problem 4 (v3): finite-free Stam architecture with case decomposition and proof-status stratification.}
\end{figure}
\clearpage

\begin{figure}[p]
\centering
\includegraphics[width=\textwidth,height=0.86\textheight,keepaspectratio]{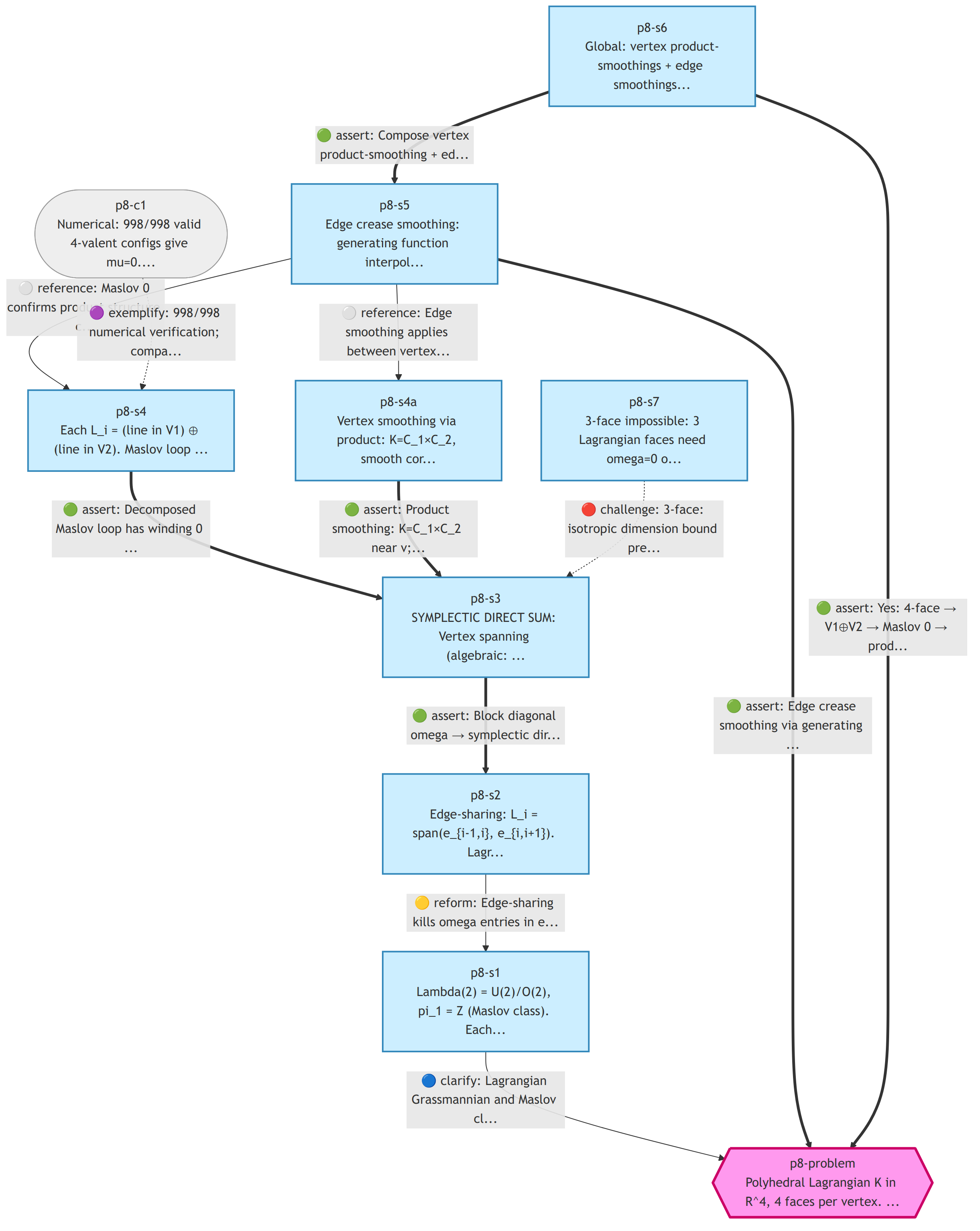}
\caption{Problem 8 (v3): Lagrangian polyhedral smoothing route with product-smoothing and edge generating-function steps.}
\end{figure}
\clearpage

\begin{figure}[p]
\centering
\includegraphics[width=\textwidth,height=0.86\textheight,keepaspectratio]{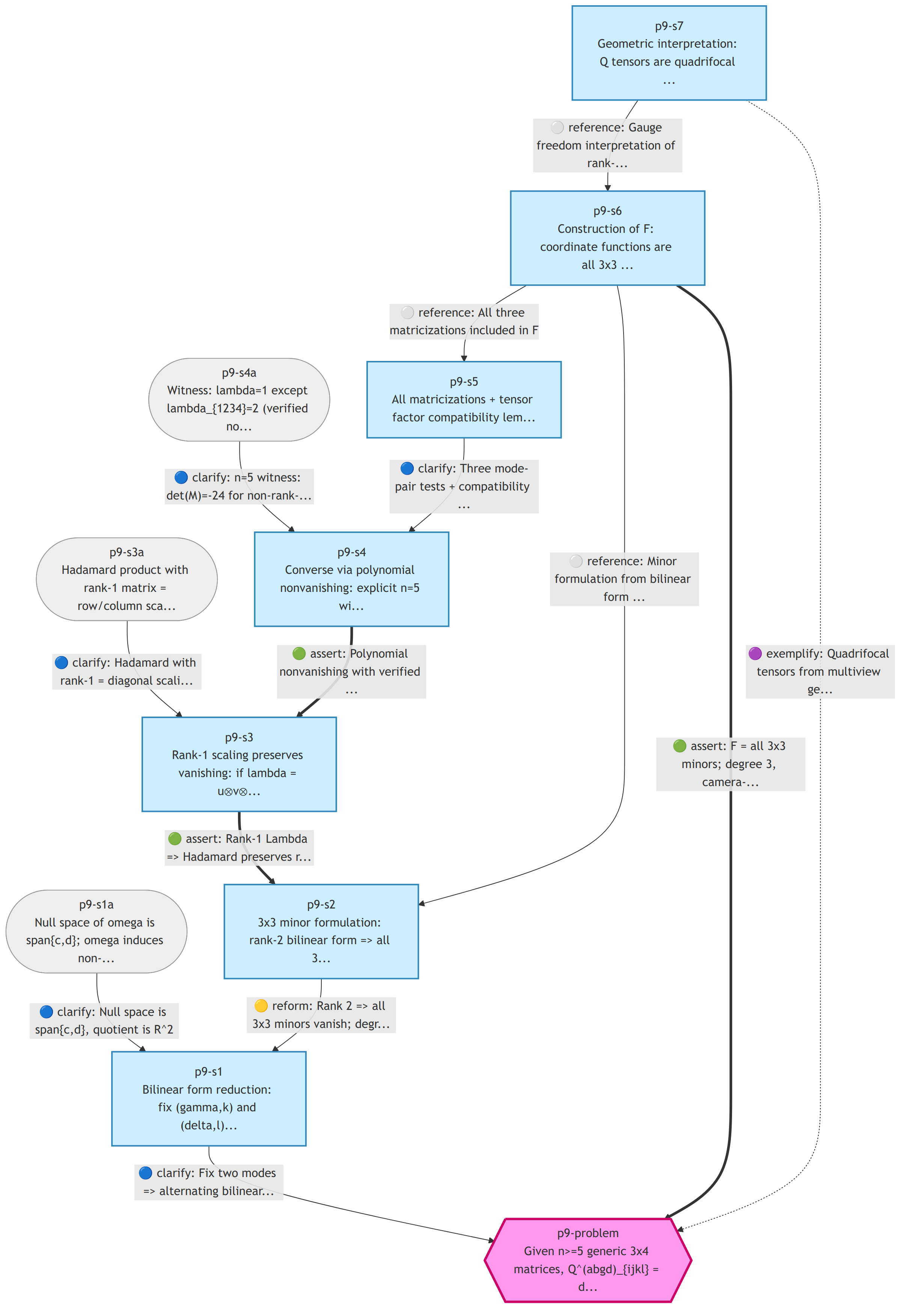}
\caption{Problem 9 (v2): rank-1 scaling characterization and polynomial nonvanishing witness flow.}
\end{figure}
\clearpage

\begin{figure}[p]
\centering
\includegraphics[width=\textwidth,height=0.86\textheight,keepaspectratio]{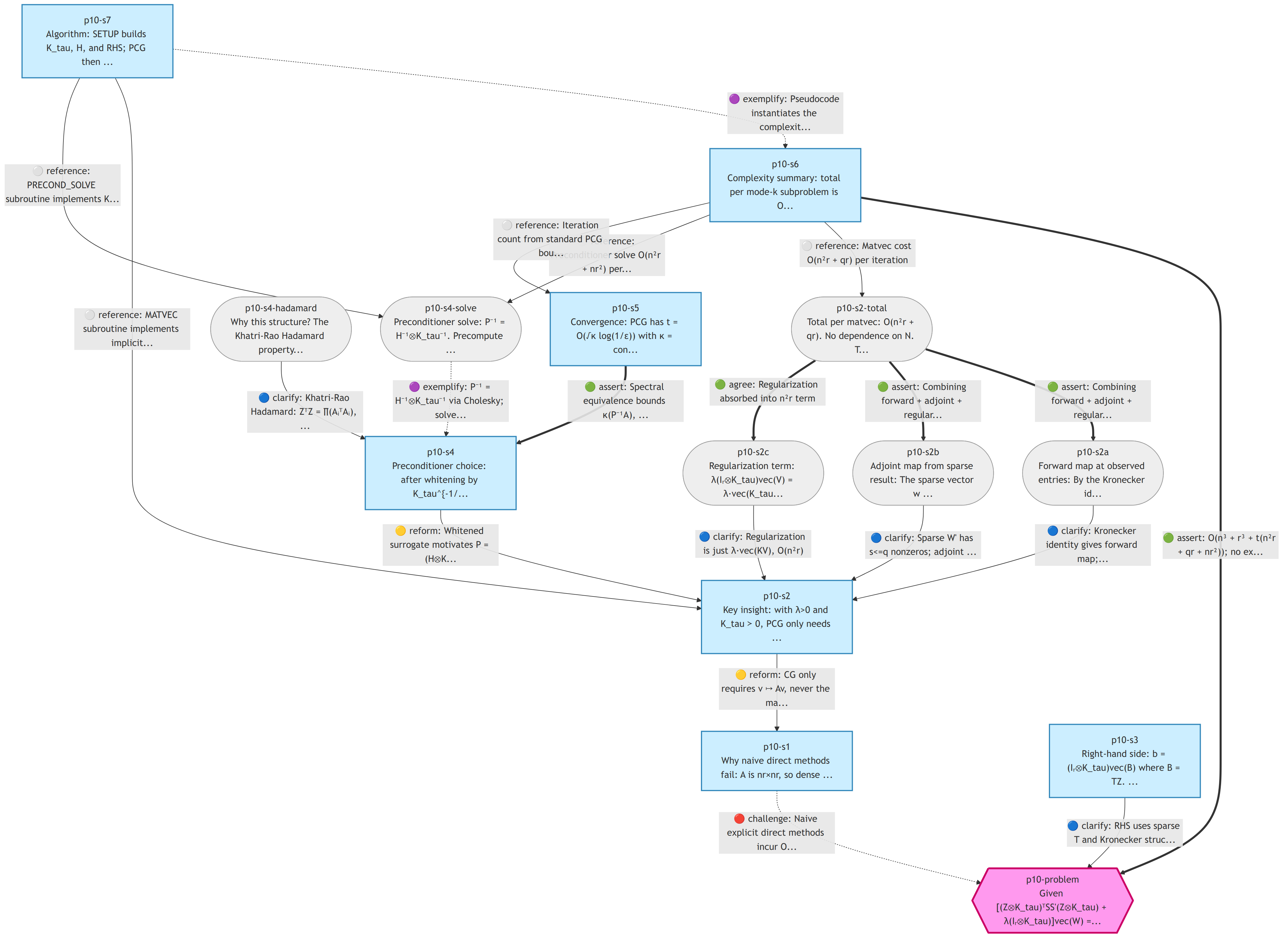}
\caption{Problem 10 (v2): preconditioned-Kronecker solve architecture and complexity path.}
\end{figure}
\clearpage

%% file: plates/the-argument-v1-tikz.tex
\begingroup
\footnotesize
\newcommand{\hash}[1]{\href{https://github.com/tothedarktowercame/futon6/commit/#1}{\texttt{#1}}}
\begin{tikzpicture}[
  x=0.96cm,
  y=1.12cm,
  >=stealth,
  line width=0.5pt,
  text depth=0pt,
  hashlbl/.style={fill=gray!15,draw=gray!55,rounded corners=0.7pt,inner sep=1.0pt,font=\fontsize{5.8}{6.3}\selectfont,align=center},
  box/.style={draw,rounded corners=2pt,align=center,inner sep=2.2pt,font=\fontsize{6.6}{7.4}\selectfont},
  data/.style={box,fill=blue!8},
  synth/.style={box,fill=yellow!20},
  formal/.style={box,fill=green!16},
  infra/.style={box,fill=violet!12},
  side/.style={box,fill=red!8},
  post/.style={box,fill=yellow!24},
  outc/.style={box,fill=green!14}
]

\node[data] (G0) at (0,14.2) {Raw data\\Git hashes and commit traces};
\node[data] (G1) at (0,12.9) {Evidence ledgers\\Pattern-tagged commit support};
\node[data] (G2) at (0,11.6) {Post-hoc reconstruction\\Pattern extraction from chronology};
\node[synth] (C0) at (0,10.3) {Conclusion claim\\Textual argument recovers strategy-level causality};

\node[synth] (C1) at (-2.8,9.0) {Canonical episode (Problem 6)\\Persistence $\to$ obstruction $\to$ layer switch $\to$ progress};
\node[outc]  (X0) at (2.8,9.0) {Cross-problem learning\\Strategy-level analogy};

\node[formal] (A0) at (-5.7,7.5) {AIF node types\\claim / conflict / preference / inference};
\node[formal] (A1) at (-6.0,6.1) {Typed edges\\attacks / supports / preference-over};
\node[formal] (A2) at (-6.0,4.7) {S-expression form\\Math content + argument structure together};
\node[infra]  (I0) at (-8.35,2.60) {Arxana (futon5)\\Typed-edge graph substrate};
\node[infra]  (I1) at (-6.4,3.35) {Peripheral model (futon3c)\\explore $\to$ reflect $\to$ edit $\to$ reflect $\to$ test};
\node[infra]  (I2) at (-6.0,1.8) {(layer, status) monitor\\Detect stationarity};

\node[outc] (X1) at (4.0,7.5) {Learning from success\\Visible closures; coach not dispatch};

\draw[red!55,dashed,rounded corners=3pt] (-3.0,1.2) rectangle (5.1,6.9);
\node[font=\scriptsize,anchor=west] at (-0.8,6.75) {Sidebar: Layer switch during writing};

\node[side] (S0) at (-0.9,6.0) {Stuck phase\\Hours in one layer, no closure};
\node[side] (S1) at (-0.9,4.8) {Intervention\\Second-agent MathOverflow scan};
\node[side] (S2) at (-0.9,3.6) {Result\\split replaces single bound target};
\node[side] (S3) at (-0.9,2.4) {Demonstration\\Reflect hop changes policy faster};

\draw[orange!70,dashed,rounded corners=3pt] (2.2,2.0) rectangle (4.8,6.3);
\node[font=\scriptsize,anchor=west] at (2.4,6.15) {Sub-sidebar: Postscript};
\node[post] (P0) at (3.5,5.5) {Cycle 7\\BMI falsified in general graphs};
\node[post] (P1) at (3.5,4.3) {Status code\\OPEN / STUCK / FALSIFIED};
\node[post] (P2) at (3.5,3.1) {Control rule\\FALSIFIED $\Rightarrow$ abandon layer};

\node[outc] (X2) at (0.0,1.0) {Metacognitive interrupt\\Trigger policy switch on stationarity/falsification};
\node[outc] (T0) at (0.0,-0.2) {THE ARGUMENT\\Raw trace data $\to$ explicit argument structure};
\node[outc] (T1) at (-7.0,-1.5) {Target system\\Live AIF graph; patterns become graph queries};

\draw[->] (G0) -- (G1) node[midway,right,hashlbl] {commit traces};
\draw[->] (G1) -- (G2) node[midway,right,hashlbl] {\hash{c557899}, \hash{0fa4e82},\\\hash{03c1f8d}, \hash{09e23db}};
\draw[->] (G2) -- (C0) node[midway,right,hashlbl] {evidence-ledger synthesis};
\draw[->] (C0) -- (C1) node[midway,above,hashlbl] {\hash{5289ca8} $\to$ \hash{cc11834} $\to$ \hash{73aa112} $\to$ \hash{7e03174}};
\draw[->] (C0) -- (X0) node[midway,above,hashlbl] {P7/P4/P8 transfer evidence};

\draw[->] (C1) -- (A0) node[midway,above,hashlbl] {\hash{cb3bd91}};
\draw[->] (A0) -- (A1) node[midway,left,hashlbl] {\hash{cb3bd91}};
\draw[->] (A1) -- (A2) node[midway,left,hashlbl] {\hash{cb3bd91}};
\draw[->] (A2.south west) -- (I0.north west) node[midway,above,hashlbl] {\hash{3454c9d}, \hash{cb3bd91}};
\draw[->] (A2) -- (I1) node[midway,above,hashlbl] {\hash{3454c9d}, \hash{cb3bd91}};
\draw[->] (I1) -- (I2) node[midway,right,hashlbl] {\hash{cb3bd91}};

\draw[->] (C1) -- (S0) node[midway,right,hashlbl] {P6 layer-switch ledger};
\draw[->] (S0) -- (S1) node[midway,left,hashlbl] {\hash{cb3bd91}};
\draw[->] (S1) -- (S2) node[midway,left,hashlbl] {\hash{7e03174}};
\draw[->] (S1.east) to[out=14,in=198] node[pos=0.40,left,yshift=6pt,hashlbl] {\hash{a7b0e3e}} (P0.west);
\draw[->] (S2) -- (S3) node[midway,left,hashlbl] {\hash{cb3bd91}};
\draw[->] (S3) -- (X2) node[midway,below,hashlbl] {\hash{cb3bd91}};

\draw[->] (P0) -- (P1) node[midway,right,hashlbl] {\hash{ea925aa}};
\draw[->] (P1) -- (P2) node[midway,right,hashlbl] {\hash{ea925aa}};
\draw[->] (P2) -- (X2) node[midway,below,hashlbl] {\hash{ea925aa}, \hash{cb3bd91}};

\draw[->] (X0) -- (X1) node[midway,above,hashlbl] {\hash{0fa4e82}, \hash{620ed57}, \hash{03c1f8d}, \hash{a5a4fbe}};
\draw[->] (X1.south east) to[out=-70,in=10] node[pos=0.58,right,hashlbl] {\hash{cb3bd91}} (X2.east);
\draw[->] (I2) -- (X2) node[midway,below,hashlbl] {\hash{cb3bd91}};
\draw[->] (X2) -- (T0) node[midway,right,hashlbl] {\hash{cb3bd91}};
\draw[->] (T0) -- (T1) node[midway,below,hashlbl] {\hash{3454c9d}, \hash{cb3bd91}};
\draw[->] (I0) -- (T1) node[midway,left,hashlbl] {\hash{3454c9d}, \hash{cb3bd91}};
\draw[->] (I1.south east) to[out=-80,in=25,looseness=0.90] node[pos=0.62,left,hashlbl] {\hash{3454c9d}, \hash{cb3bd91}} (T1.north);

\end{tikzpicture}
\endgroup